\documentclass[letterpaper,journal,twoside]{IEEEtran}

\usepackage{xr}

\newcommand{\omitted}[1]{}

\title{
 \fontsize{22}{22} \selectfont 
Simultaneous System Identification and Model \\
Predictive Control with No Dynamic Regret
}

\author{
\thanks{Manuscript received November 25, 2024; Revised March 13, 2025; Accepted April 21, 2025. This work was partially supported by NSF CAREER No. 2337412. This paper was recommended for publication by Editor-in-Chief Wolfram Burgard and Editor Jeannette Bohg upon evaluation of reviewers’ comments. (Corresponding author: Hongyu Zhou.)}
Hongyu Zhou \;\; Vasileios Tzoumas
\thanks{Department of Aerospace Engineering, University of Michigan, Ann Arbor, MI 48109 USA;  {\tt\footnotesize \{zhouhy, vtzoumas\}@umich.edu}}
\thanks{Digital Object Identifier (DOI): see top of this page.}
}

\usepackage{cite}

\usepackage{comment}
\usepackage{siunitx}
\usepackage{relsize}
\usepackage{ifthen}
\usepackage[colorinlistoftodos]{todonotes}
\usepackage[vlined,ruled,linesnumbered]{algorithm2e}
\usepackage{graphics} 
\usepackage{rotating}
\usepackage{color}
\usepackage{enumerate}
\usepackage[T1]{fontenc}
\usepackage{psfrag}
\usepackage{epsfig} 
\usepackage{booktabs}
\usepackage{graphicx}
\usepackage{url}
\usepackage{multirow}
\usepackage{array}
\usepackage{latexsym}
\usepackage{amsfonts}
\usepackage{amsmath}
\usepackage{amssymb}
\usepackage{xstring}
\usepackage{algorithmic}
\usepackage{multirow}
\usepackage{xcolor}
\usepackage{prettyref}
\usepackage{flexisym}
\usepackage{bigdelim}
\usepackage{breqn} 
\usepackage{listings}

\usepackage{xspace}
\usepackage{bm}
\usepackage{soul}
\graphicspath{{./figures/}}
\usepackage{tikz}
\usetikzlibrary{matrix,calc}
\usepackage{lipsum}
\usepackage{mdwlist}

\let\itemize\undefined
\makecompactlist{itemize}{stditemize}
\usepackage{enumitem}
\usepackage{caption}
\usepackage{epstopdf}
\usepackage{subfigure}

\usepackage{amsthm}
\usepackage{mathtools}

\makeatletter
\let\NAT@parse\undefined
\makeatother
\usepackage{hyperref}
\usepackage{cleveref}

\hypersetup{%
  colorlinks=true,
  linkcolor=blue,
  filecolor=magenta,      
  urlcolor=blue,
  citecolor=red,
  linkbordercolor={0 0 1}
}

\newtheorem{theorem}{Theorem}
\newtheorem{problem}{Problem}

\newtheorem{corollary}{Corollary}

\newtheorem{lemma}{Lemma}
\newtheorem{assumption}{Assumption}
\newtheorem{definition}{Definition}
\newtheorem{proposition}{Proposition}

\newtheorem{remark}{Remark}

\newcommand{\bdmath}{\begin{dmath}}
\newcommand{\edmath}{\end{dmath}}
\newcommand{\beq}{\begin{equation}}
\newcommand{\eeq}{\end{equation}}
\newcommand{\bdm}{\begin{displaymath}}
\newcommand{\edm}{\end{displaymath}}
\newcommand{\bea}{\begin{eqnarray}}
\newcommand{\eea}{\end{eqnarray}}
\newcommand{\beal}{\beq \begin{array}{lll}}
\newcommand{\eeal}{\end{array} \eeq}
\newcommand{\beas}{\begin{eqnarray*}}
\newcommand{\eeas}{\end{eqnarray*}}
\newcommand{\ba}{\begin{array}}
\newcommand{\ea}{\end{array}}
\newcommand{\bit}{\begin{itemize}}
\newcommand{\eit}{\end{itemize}}
\newcommand{\ben}{\begin{enumerate}}
\newcommand{\een}{\end{enumerate}}

\newcommand{\calD}{{\cal D}}

\newcommand{\calF}{{\cal F}}

\newcommand{\calH}{{\cal H}}

\newcommand{\calO}{{\cal O}}

\newcommand{\calU}{{\cal U}}

\newcommand{\calZ}{{\cal Z}}

\definecolor{myblue}{RGB}{65 105 225}

\newcommand{\hide}[1]{}

\newcommand{\hiddenText}{{\color{gray} hidden text.}}
\newcommand{\hideWithText}[1]{\hiddenText}

\newcommand{\diag}[1]{\mathrm{diag}\left(#1\right)}

\newcommand{\scenario}[1]{{\fontsize{9}{8.7}\selectfont\sf#1}\xspace}

\newcommand{\scenariof}[1]{{\fontsize{7}{7}\selectfont\sf#1}\xspace}

\newcommand{\myx}{\mathbf{x}}

\newcommand{\ie}{\emph{i.e.},\xspace}
\newcommand{\eg}{\emph{e.g.},\xspace}

\newcommand{\myParagraph}[1]{{\bf #1.}\xspace}

\newcommand{\OCO}{\scenario{{OCO}}}
\newcommand{\RKHS}{\scenario{{RKHS}}}

\newcommand{\OGD}{\scenario{{OGD}}}
\newcommand{\GP}{\scenario{{GP}}}
\newcommand{\INDI}{\scenario{{INDI}}}
\newcommand{\MPC}{\scenario{{MPC}}}
\newcommand{\MPCf}{\scenariof{{MPC}}}
\newcommand{\NMPC}{\scenario{{Nominal MPC}}}
\newcommand{\GPMPC}{\scenario{{GP-MPC}}}
\newcommand{\NSMPC}{\scenario{{NS-MPC}}}
\newcommand{\LMPC}{\scenario{{L1-MPC}}}
\newcommand{\SReg}{\operatorname{Regret}_T^S}
\newcommand{\DReg}{\operatorname{Regret}_T^D}
\newcommand{\OLMPC}{\scenario{{SSI-MPC}}}

\PassOptionsToPackage{end}{algorithmic}

\renewcommand{\baselinestretch}{0.96}

\begin{document}

\makeatletter

\g@addto@macro\@maketitle{
\setcounter{figure}{0}
\centering
\includegraphics[width=0.32\textwidth]{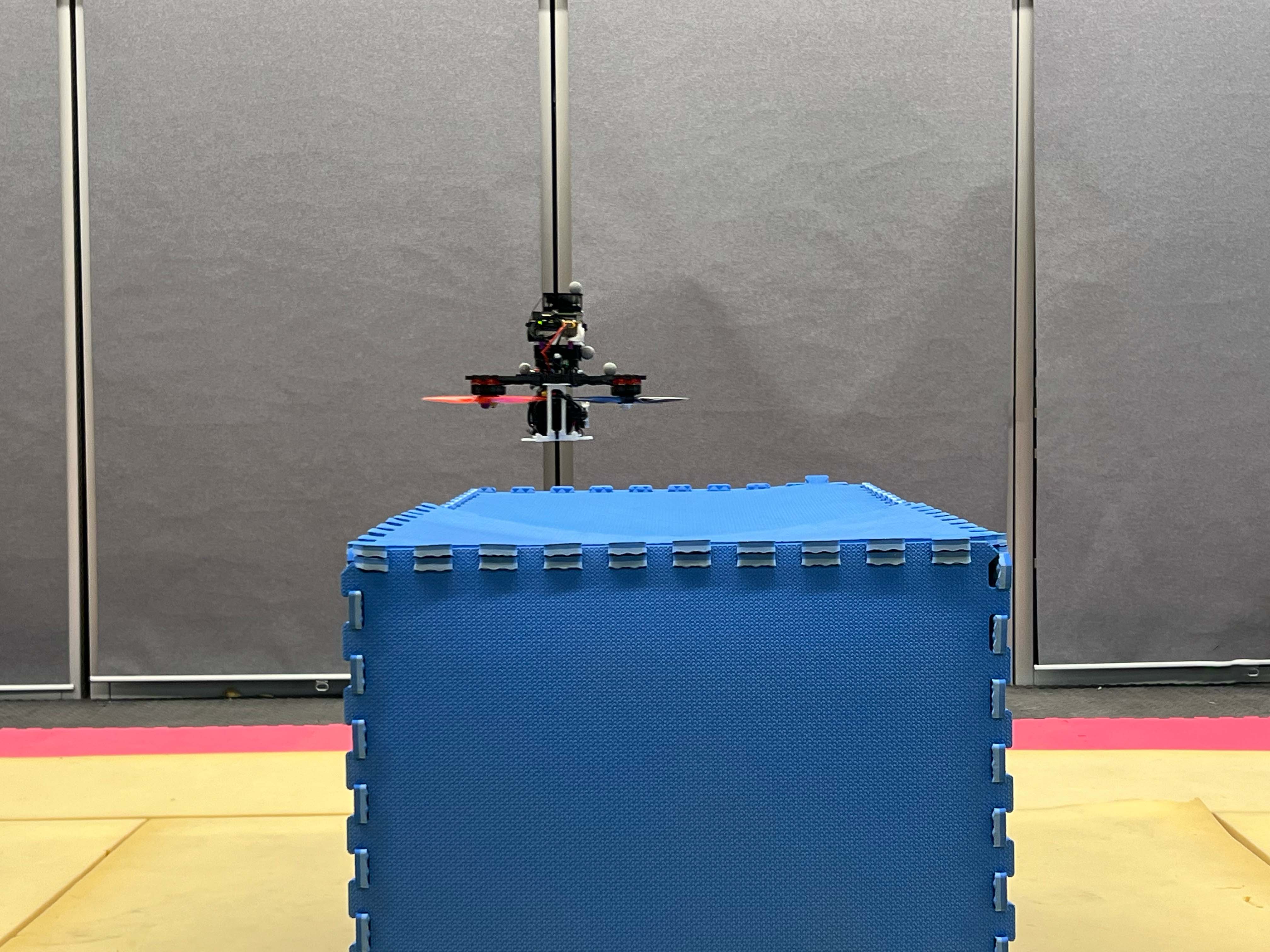}~~
\includegraphics[width=0.32\textwidth]{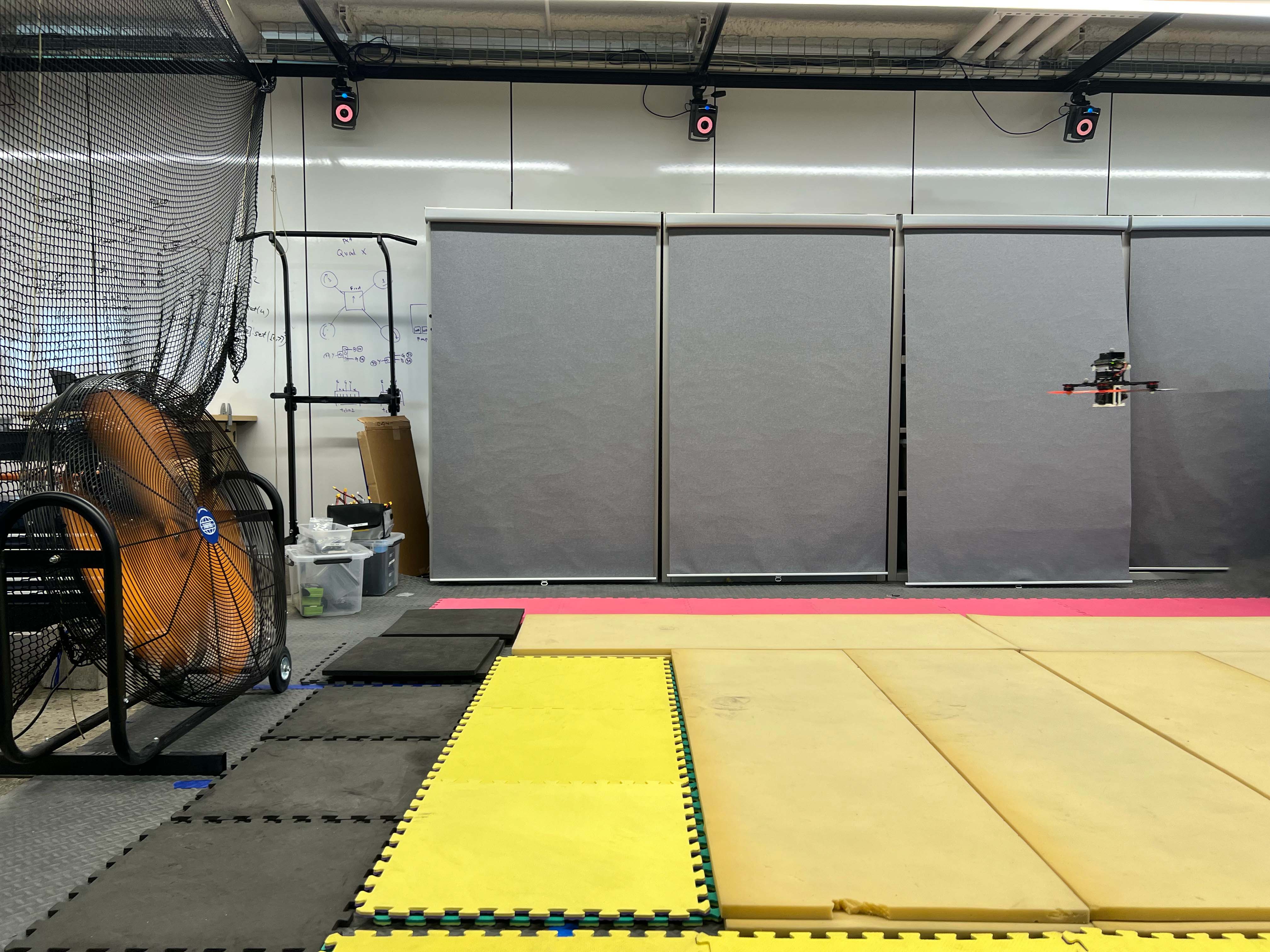}~~
\includegraphics[width=0.32\textwidth]{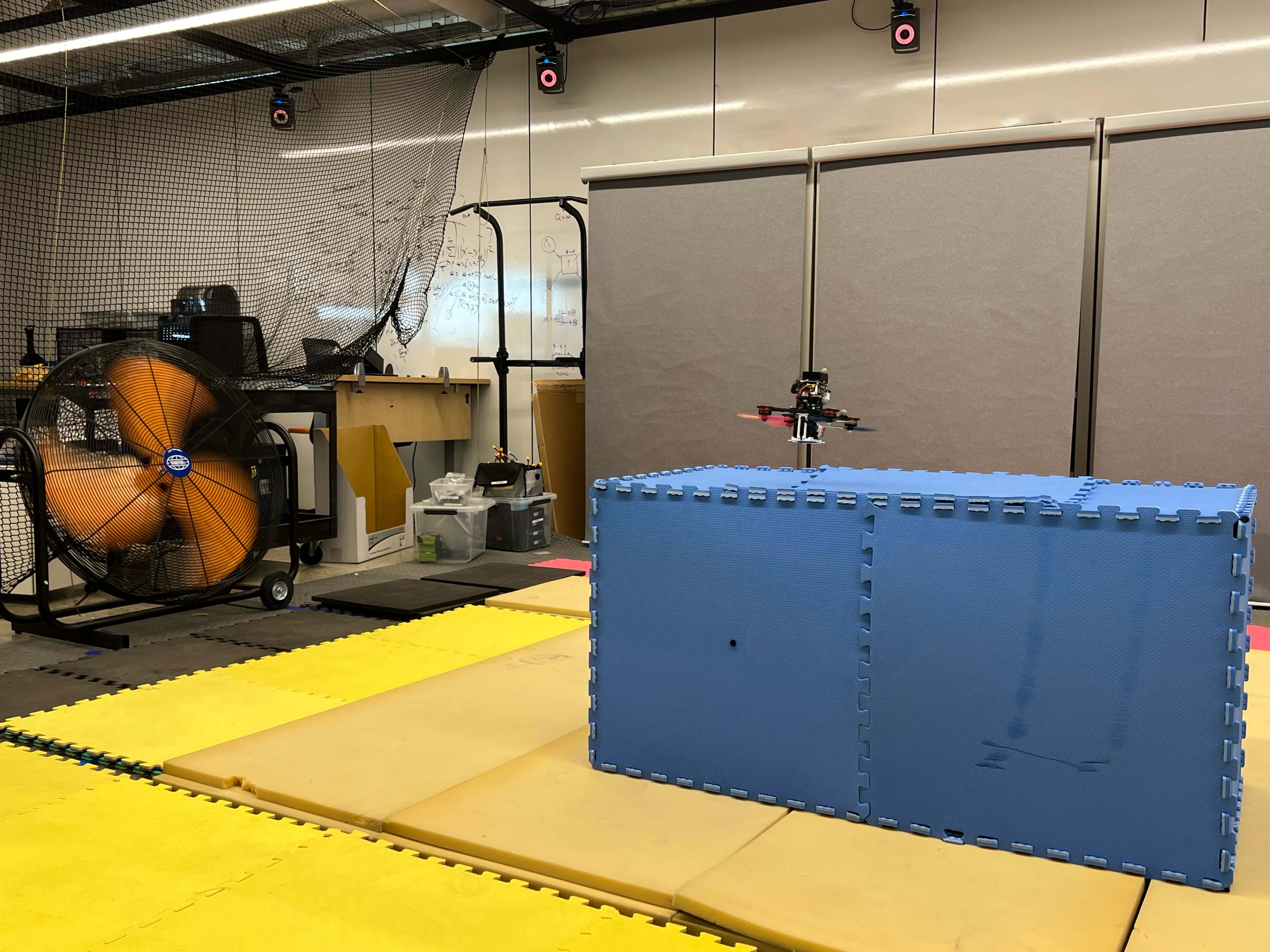}
\captionsetup{font=small}
\captionof{figure}{
\textbf{Simultaneous System Identification and Model Predictive Control:   Tested Scenarios of Reference Trajectory Tracking subject to Various Unmodeled Aerodynamic Disturbances.} In this paper, we focus on simultaneous system identification and model predictive control where the robots' capacity to achieve accurate and efficient tracking control is challenged by unknown both dynamics and (possibly state-dependent) disturbances. For example, the ability of a quadrotor to track a reference trajectory can be challenged by a variety of unknown aerodynamic disturbances: \textit{ground effects}~(left \& right), \textit{wind disturbances}~(center \& right), and \textit{drag} (all). We provide a control algorithm that demonstrates an ability to simultaneously learn such unknown dynamics/disturbances in a self-supervised manner, based on the data collected on-the-go, and uses the learned models for predictive control. We prove that the algorithm guarantees bounded suboptimality against the optimal controller in hindsight, under technical assumptions that we describe later in the paper.
\label{fig_hardware}}
\vspace{-.6cm}
}

\makeatother
\maketitle

\begin{abstract}
We provide an algorithm for the simultaneous sys- tem identification and model predictive control of nonlinear systems. The algorithm has finite-time near-optimality guarantees and asymptotically converges to the optimal (non-causal) controller.  Particularly, the algorithm enjoys sublinear \textit{dynamic regret}, defined herein as the suboptimality against an optimal clairvoyant controller that knows how the unknown disturbances and system dynamics will adapt to its actions.
The algorithm is self-supervised and applies to control-affine systems with unknown dynamics and disturbances that can be expressed in reproducing kernel Hilbert spaces.  
Such spaces can model external disturbances and modeling errors that can even be adaptive to the system's state and control input.  For example, they can model wind and wave disturbances to aerial and marine vehicles, or inaccurate model parameters such as inertia of mechanical systems.
We are motivated by the future of autonomy where robots will autonomously perform complex tasks despite real-world unknown disturbances such as wind gusts.
The algorithm first generates random Fourier features that are used to approximate the unknown dynamics or disturbances. Then, it employs model predictive control based on the current learned model of the unknown dynamics (or disturbances).
The model of the unknown dynamics is updated online using least squares based on the data collected while controlling the system.
We validate our algorithm in both hardware experiments and physics-based simulations.  The simulations include (i) a cart-pole aiming to maintain the pole upright despite inaccurate model parameters, and (ii) a quadrotor aiming to track reference trajectories despite unmodeled aerodynamic drag effects.
The hardware experiments include a quadrotor aiming to track a circular trajectory despite unmodeled aerodynamic drag effects, ground effects, and wind disturbances.
The code is open-sourced at \href{https://github.com/UM-iRaL/SSI-MPC}{https://github.com/UM-iRaL/SSI-MPC}.
\end{abstract}



\section{Introduction}\label{sec:Intro}
\IEEEPARstart{I}{n} the future, mobile robots will automate fundamental tasks such as package delivery~\cite{ackerman2013amazon},  target tracking~\cite{chen2016tracking}, and inspection and maintenance~\cite{seneviratne2018smart}.
Such tasks require accurate and efficient tracking control.
But achieving accuracy and efficiency is challenging since such tasks often require the robots to operate under highly uncertain conditions, particularly, under unknown dynamics and (possibly state-dependent) disturbances.  
For example, they require the quadrotors to (i) pick up and carry packages of unknown weight, (ii) chase mobile targets at high speeds where the induced aerodynamic drag is hard to model, and (iii) inspect and maintain outdoor infrastructure exposed to turbulence and wind gusts.

State-of-the-art methods for control under unknown dynamics and disturbances typically rely on offline or online methods, or a mix of the two, including:   
robust control~\cite{mayne2005robust,goel2020regret,sabag2021regret,martin2022safe,didier2022system,zhou2023safe,martin2024guarantees}; 
adaptive control and disturbance compensation~\cite{slotine1991applied,krstic1995nonlinear,ioannou1996robust,boffi2021regret,boffi2022nonparametric,tal2020accurate,wu2023mathcal,das2024robust,jia2023evolver,rahman2016tutorial}; 
offline learning for control~\cite{sanchez2018real,carron2019data,torrente2021data,hewing2019cautious,tobin2017domain,ramos2019bayessim,lee2020learning,koopman1931hamiltonian,brunton2016sparse,abraham2017model,bruder2020data}; 
offline learning for control with online adaption~\cite{finn2017model,williams2017information,nagabandi2018learning,belkhale2021model,shi2019neural,o2022neural,saviolo2023active};
and online learning~\cite{hazan2022introduction,agarwal2019online,zhao2022non,gradu2023adaptive,zhou2023efficient,zhou2023safecdc,zhou2023saferal,krause2011contextual,berkenkamp2016safe,chowdhury2017kernelized,djeumou2022fly,vinod2022fly,valko2013finite}. 
Among these, the offline learning methods require data collection for offline training, a process that can be expensive/time-consuming~\cite{pierson2017deep}, instead of collecting data online and learning on the spot.  For this reason, these methods may not also generalize to different scenarios than those used for offline learning~\cite{mohri2018foundations,abu2012learning}.  The robust control methods, given a known upper bound on the magnitude of the noise, can be conservative due to assuming worse-case disturbance realization~\cite{zhou1998essentials,lavretsky2012robust}, instead of planning based on an accurate predictive model of the disturbance. Similarly, the adaptive control and the online learning control methods may exhibit sub-optimal performance due to only reacting to the history of observed disturbances, instead of planning based on an accurate predictive model of the disturbance~\cite{slotine1991applied,krstic1995nonlinear,ioannou1996robust,agarwal2019online,djeumou2022fly}.
We discuss the related work in extension in \Cref{sec:lit_review}.

\begin{figure}[t]
    \centering
    \includegraphics[width=0.49\textwidth]{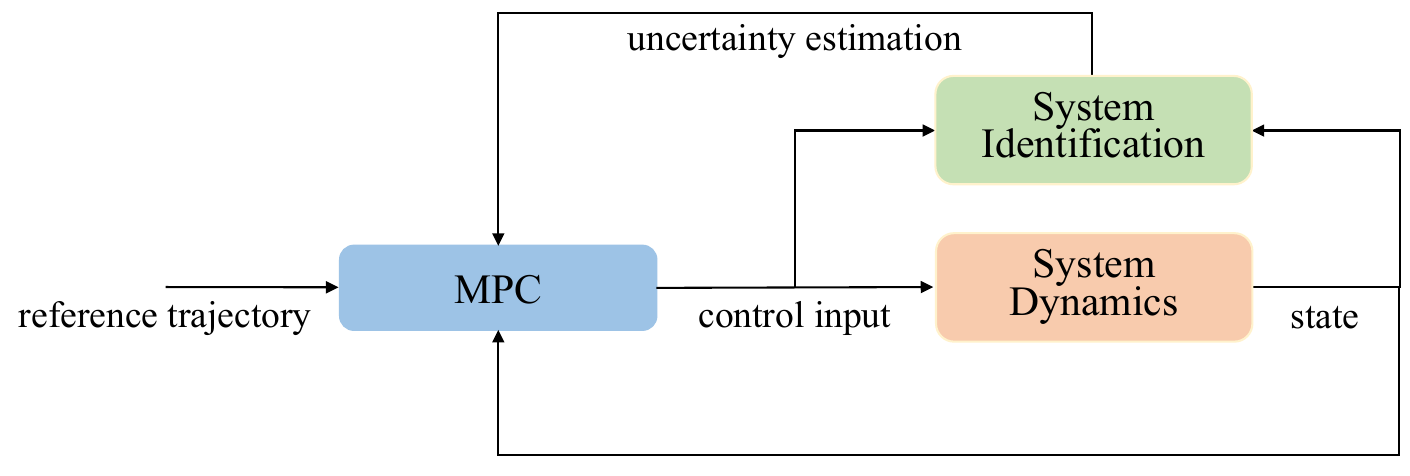}
    \caption{\textbf{Overview of Simultaneous System Identification and Model Predictive Control Pipeline.} The pipeline is composed of two interacting modules: (i) a model predictive control (\MPC) module, and (ii) an online system identification module. The \MPC module uses the estimated unknown disturbances/dynamics from the system identification module to calculate the next control input. Given the control input and the observed new state, the online system identification module then updates the estimate of the unknown disturbances/dynamics. }
    \label{fig_framework}
\end{figure}

In this paper, instead, we leverage the success of model predictive control methods for accurate tracking control~\cite{sun2022comparative}.  To this end, we propose a self-supervised method to learn online a predictive model of the unknown uncertainties (Fig.~\ref{fig_framework}). Therefore, the proposed method promises to enable: one-shot online learning, instead of offline (or episodic learning); online control that adapts to the actual disturbance realization, instead of the worst-case; and control planned over a look-ahead horizon of predicted system dynamics and disturbances, instead of their past.  We elaborate on our contributions next.

\myParagraph{Contributions} 
We provide a real-time and asymptotically-optimal algorithm for the simultaneous system identification and control of nonlinear systems. 
Specifically, the algorithm enjoys sublinear \textit{dynamic regret}, \textit{defined herein as the suboptimality against an optimal clairvoyant controller that knows how the unknown disturbances and system dynamics will adapt to its actions}.  {The said definition of dynamic regret differs from the standard definition of dynamic regret in the literature, \eg \cite{agarwal2019online,zhao2022non,gradu2023adaptive,zhou2023efficient,zhou2023safecdc,zhou2023saferal}  where the suboptimality of an Algorithm is measured against the controller that is optimal for the same realization of disturbances as the Algorithm experienced, instead of the realization that the optimal controller would experience given that the disturbances can be adaptive (\Cref{rem:adaptivity}).}

The algorithm is self-supervised and applies to control-affine systems with unknown dynamics and disturbances that can be expressed in reproducing kernel Hilbert spaces~\cite{cucker2002mathematical}.  
Such spaces can model external disturbances and modeling errors that are even adaptive to the system's state and control input.  For example, they can model wind and wave disturbances to aerial and marine vehicles, or inaccurate model parameters such as inertia of mechanical systems.

\paragraph*{Description of the algorithm} The algorithm is composed of two interacting modules (Fig.~\ref{fig_framework}): (i) a model predictive control (\MPC) module, and (ii) an online system identification module.  At each time step, the \MPC module uses the estimated unknown disturbances and system dynamics from the system identification module to calculate the next control input. Given the control input and the observed new state, the online system identification module then updates the estimate of the unknown disturbances and system dynamics.  

\paragraph*{Approach of predictive online system identification} We enable the predictive online system identification module by employing online least-squares estimation via online gradient descent (\OGD)~\cite{hazan2016introduction}, where the unknown disturbances/dynamics are parameterized as a linear combination of random Fourier features~\cite{rahimi2007random,rahimi2008uniform,boffi2022nonparametric}.  
This allows us to maintain the computational efficiency of classical finite-dimensional parametric approximations that can be used in model predictive control while retaining the expressiveness of the \RKHS with high probability.
Using the finite-dimensional parametric approximation, the parameters are updated in a self-supervised manner using \OGD at each time step.
Such approximation then is used in \MPC to select control input.

\paragraph*{Performance guarantee}  
The algorithm asymptotically has \textit{no dynamic regret}, that is,  it asymptotically matches the performance of the optimal controller that knows a priori how the unknown disturbances and system dynamics will adapt to its actions. Particularly, we provide the following finite-time performance guarantee for \Cref{alg:MPC} (\Cref{theorem:regret_OLMPC}):
    $$\text{dynamic regret} \leq \calO\left(T^{\frac{3}{4}}\right).$$

\paragraph*{Technical assumptions} 
The no-dynamic-regret guarantee holds true under the assumptions of bounds on \MPC's cost function and value function, and Lipschitzness of \MPC's cost function and unknown dynamics/disturbances, along with the previously stated assumption that the unknown disturbances and dynamics can be expressed in reproducing kernel Hilbert spaces. 
These assumptions are milder than the exponential stability or exponentially incrementally stability assumption required in \cite{lale2021model,boffi2021regret,boffi2022nonparametric,karapetyan2024regret,nonhoff2024online} and persistent excitation required in \cite{lale2021model,muthirayan2022online}.

\myParagraph{Numerical Evaluations}
We validate the algorithm in physics-based simulations and conduct an ablation study (\Cref{sec:exp-sim}). 
Specifically, we validate our algorithm in simulated scenarios of (i) a cart-pole system that aims to stabilize at a setpoint despite inaccurate model parameters~(\Cref{subsec:sim-1} and \Cref{subsec:sim-2}), 
and (ii) a quadrotor that aims to track reference trajectories subject to unmodeled aerodynamic forces~(\Cref{subsec:sim-3}). 
In the cart-pole experiments~(\Cref{subsec:sim-1}), We compare our algorithm with a nominal \MPC (\NMPC) that ignores the unknown dynamics or disturbances, the non-stochastic \MPC (\NSMPC)~\cite{zhou2023saferal}, and the Gaussian process \MPC (\GPMPC)~\cite{hewing2019cautious}.
In the quadrotor experiment (\Cref{subsec:sim-3}), we compare our algorithm with a \NMPC controller, and the Gaussian process \MPC (\GPMPC)~\cite{torrente2021data}. 
In the simulations, the algorithm achieves real-time control and superior tracking performance over the compared methods, demonstrating its ability to perform simultaneous online model learning and control.
We also show that the performance of \Cref{alg:MPC} can be further improved by combining it with an incremental nonlinear dynamic inversion (\INDI) inner-loop controller~\cite{tal2020accurate} in~\Cref{subsec:sim-3}.
In addition, we conduct sensitivity analysis in~\Cref{subsec:sim-2} on the tuning parameters of \Cref{alg:MPC}: how the number of random Fourier features and learning rate of \OGD affect the performance of \Cref{alg:MPC}.

\myParagraph{Hardware Experiments} 
We implement the provided algorithm in a quadrotor (\Cref{sec:exp-hardware} and \Cref{app:hardware}).
In \Cref{sec:exp-hardware}, we consider that the quadrotor is tasked to track a circular trajectory with diameter $1m$ at maximal speed $v_{m} = 0.8~m/s$ under (i) ground effects~(\Cref{fig_hardware}, left), (ii) wind disturbances~(\Cref{fig_hardware}, center), (iii) both ground effects and wind disturbances~(\Cref{fig_hardware}, right).
In \Cref{app:hardware}, we consider the quadrotor is tasked to track the same circular trajectory with maximal speed either $0.8~m/s$ or $1.3~m/s$, without ground effects and wind disturbances. In such cases, the quadrotor experiences unknown aerodynamic effects including body drag, rotor drag, and turbulent effects caused by the propellers.
Across all hardware experiments, we compare \Cref{alg:MPC} with a nominal \MPC (\NMPC), and the L1-adaptive \MPC  (\LMPC)~\cite{wu2023mathcal}. 
The algorithm achieves superior tracking performance over the compared methods under all tested scenarios, demonstrating its ability to perform simultaneous online system identification and predictive control.

\section{Simultaneous System Identification \\ and Model Predictive Control}\label{sec:problem}

We formulate the problem of \textit{Simultaneous System Identification and Model Predictive Control} (\Cref{prob:control}).
To this end, we use the following framework and assumptions.

\myParagraph{Control-Affine Dynamics}
We consider control-affine system dynamics of the form
\begin{equation}
	x_{t+1} = f\left(x_{t}\right) + g\left(x_{t}\right) u_{t} + h\left(z_{t}\right), \quad t \geq 1, 
    \label{eq:affine_sys}
\end{equation}
where $x_t \in\mathbb{R}^{d_x}$ is the state, $u_t \in\mathbb{R}^{d_u}$ is the control input, $f: \mathbb{R}^{d_x} {\rightarrow} \mathbb{R}^{d_x}$ and $g: \mathbb{R}^{d_x} {\rightarrow} \mathbb{R}^{d_x} \times \mathbb{R}^{d_u}$ are known locally Lipschitz functions, $h: \mathbb{R}^{d_z} \rightarrow \mathbb{R}^{d_x}$ is an unknown locally Lipschitz function, and $z_t \in\mathbb{R}^{d_z}$ is a vector of features chosen as a subset of $[x_t^\top \ u_t^\top]^\top$.\footnote{The feature vector $z_t$ can also include, for example, $t$~\cite[Remark~3.1]{boffi2022nonparametric} and $h\left(z_{t-1}\right)$. We can also model time delay by using $x_{t+1} = f\left(x_{t}\right) + g\left(x_{t}\right) u_{t-\tau} + \tilde{h}\left(\tilde{z}_{t}\right) = f\left(x_{t}\right) + g\left(x_{t}\right) u_{t} + \left( g\left(x_t\right)u_{t-\tau} - g\left(x_t\right)u_{t} + \tilde{h}\left(\tilde{z}_{t}\right)\right)$, where $\tau$ represents the latency. We can define
$h\left(z_t\right) \triangleq g\left(x_t\right)u_{t-\tau} - g\left(x_t\right)u_{t} + \tilde{h}\left(\tilde{z}_{t}\right)$ with $z_t$ including $x_{t}, u_{t}, x_{t-\tau}, u_{t-\tau}, \tilde{z}_t$.}  Particularly, $h\left(\cdot\right)$ represents unknown disturbances or system dynamics. 

We refer to the undisturbed system dynamics $x_{t+1} = f\left(x_{t}\right) + g\left(x_{t}\right) u_{t}$ as the \textit{nominal dynamics}.

\begin{remark}[Extension to Systems with Unmodeled Noise]
In the Appendix, we extend the results in this paper to systems corrupted with unknown (potentially non-stochastic)
noise that is additive to the system dynamics, \ie
\begin{equation}
	x_{t+1} = f\left(x_{t}\right) + g\left(x_{t}\right) u_{t} + h\left(z_{t}\right) + w_{t}, 
\end{equation}
where $w_{t}$ represents, \eg \space process noise or estimation error of $h\left(z_{t}\right)-\hat{h}\left(z_{t}\right)$.


\end{remark}



The unknown disturbance (or system dynamics) $h\left(\cdot\right)$ may be adapting to the state and the control input.  Assumptions on $h\left(\cdot\right)$ will be introduced in \Cref{subsec:RFF}.




\myParagraph{Model Predictive Control (\MPC)} 
\MPC selects a control input $u_t$ by simulating the nominal system dynamics over a look-ahead horizon $N$, \ie~\MPC selects $u_t$ by solving the optimization problem~\cite{rawlings2017model,borrelli2017predictive}: 
\begin{subequations}
    \label{eq:mpc_def}
    \begin{align}
        & \underset{{u}_{t}, \ \ldots, \ {u}_{t+N-1}}{\textit{min}} \sum_{k=t}^{t+N-1} c_{k}\left(x_{k},u_{k}\right) \label{eq:mpc_def_obj} \\
        & \ \ \operatorname{\textit{subject~to}} \;\quad x_{k+1} = f\left(x_{k}\right) + g\left(x_{k}\right) u_{k}, \label{eq:mpc_def_dyn} \\
        & \qquad \qquad \qquad \; u_{k}\in \calU, \ \ k\in\{t,\ldots, t+N-1\},
    \end{align}
\end{subequations}
where $c_{t}\left(\cdot,\cdot\right): \mathbb{R}^{d_{x}} \times \mathbb{R}^{d_{u}} {\rightarrow} \mathbb{R}$ is the cost function, and $\calU$ is a compact set that represents constraints on the control input due to, \eg controller saturation. 
We discuss how to incorporate state constraints in \Cref{app:safety}.

The optimization problem in \cref{eq:mpc_def} ignores the unknown disturbances $h\left(\cdot\right)$. To improve performance in the presence of $h\left(\cdot\right)$, in this paper we propose a method to estimate $h\left(\cdot\right)$ online so \cref{eq:mpc_def} can be adapted to the optimization problem:
\begin{subequations}
    \label{eq:mpc_ada_def}
    \begin{align}
       &\underset{{u}_{t}, \ \ldots, \ {u}_{t+N-1}}{\textit{min}}  \sum_{k=t}^{t+N-1} c_{k}\left(x_{k},u_{k}\right) \label{eq:mpc_ada_obj} \\
        & \ \ \operatorname{\textit{subject~to}} \;\quad x_{k+1} = f\left(x_{k}\right) + g\left(x_{k}\right) u_{k} + \hat{h}\left(z_{k}\right), \label{eq:mpc_ada_dyn}\\
        & \qquad \qquad \qquad \ u_{k}\in \calU,  \ \ k\in\{t,\ldots, t+N-1\}, 
    \end{align}
\end{subequations}
where $\hat{h}\left(\cdot\right)$ is the estimate of $h\left(\cdot\right)$.  Specifically, $\hat{h}\left(\cdot\right)  \triangleq \hat{h}\left(\cdot~; \hat{\alpha}\right)$ where $\hat{\alpha}$ is a parameter that is updated online by our proposed method to improve the control performance.


We define the notion of \MPC's \textit{value function} and state the assumption on the cost function and value function.


\begin{definition}[Value Function~\cite{grimm2005model}]\label{def:opt_value}
Given a state $x$ and parameter $\hat{\alpha}$, the \emph{value function} $V_t\left(x ; \hat{\alpha}\right)$  is defined as the optimal value of \cref{eq:ada_mpc_value_obj_mother}:
\begin{subequations}
\label{eq:ada_mpc_value_obj_mother}
    \begin{align}
        &\underset{{u}_{t}, \ \ldots, \ {u}_{t+N-1}}{\textit{min}}  \sum_{k=t}^{t+N-1} c_{k}\left(x_{k},u_{k}\right) \label{eq:ada_mpc_value_obj} \\
        & \ \ \operatorname{\textit{subject~to}} \;\quad x_{k+1} = f\left(x_{k}\right) + g\left(x_{k}\right) u_{k} + \hat{h}\left(z_{k}\right), \label{eq:ada_mpc_value_dyn}\\
        & \qquad \qquad \qquad \ x_{t} = x, \; u_{k}\in \calU,  \;  k\in\{t,\ldots, t+N-1\}.
    \end{align}
    \label{eq:ada_mpc_value}
\end{subequations}
\end{definition}

\begin{assumption}[Bounds on Cost Function and Value Function~\cite{grimm2005model}]\label{assumption:stability}
    For all $\hat{\alpha}$ in a compact set $\mathcal{D}$, there exist positive scalars $\underline{\lambda}$, $\overline{\lambda}$, and a continuous function $\sigma: \mathbb{R}^{d_x} \rightarrow \mathbb{R}_{+}$, such that
    (i) $c_{t}\left(x,u\right) \geq \underline{\lambda} \sigma\left(x\right)$, $\forall x,\; u,\; t$; 
    (ii) $V_t\left(x; \hat{\alpha}\right) \leq \overline{\lambda}\sigma\left(x\right)$, $\forall x,\; t$,
    and (iii) $\lim_{\|x\| \rightarrow \infty} \sigma\left(x\right) \rightarrow \infty$.
\end{assumption}

Under \Cref{assumption:stability}, the \MPC policy in \cref{eq:mpc_ada_def} can be proved to ensure that the system in \cref{eq:mpc_ada_dyn} is globally asymptotically stable \cite{grimm2005model}. 
\Cref{assumption:stability} requires the asymptotically stability of the system in \cref{eq:mpc_ada_dyn} for all $\hat{\alpha}$. We discuss how to obtain stability of the designed \MPC algorithm under only the stability of the nominal system~(\cref{eq:mpc_def_dyn}) in \Cref{app:stability}.


\begin{remark}[Discussion on \Cref{assumption:stability}]
    Consider the quadratic cost $c_{t}\left(x_{t}, u_{t}\right) = x_{t}^\top Q x_{t} + u_{t}^\top R u_{t}$. For linear systems, \Cref{assumption:stability} is satisfied~\cite[Lemma~1]{grimm2005model}. For nonlinear systems, \Cref{assumption:stability} is satisfied when the quadratic cost is (exponentially/asymptotically) controllable to zero with respect to $\sigma: \mathbb{R}^{d_x} \rightarrow \mathbb{R}_{+}$~\cite[Section. III]{grimm2005model}.
    Particulary, for the nonlinear case: a) when $R=0$, the (exponentially/asymptotically) controllability holds if \MPC can drive $x\rightarrow0$; b) for $R\neq0$, it is non-trivial to verify \Cref{assumption:stability} and a necessary condition is $\left(x,u\right)=\left(0,0\right)$ being an equilibrium point or $R\triangleq R\left(x\right)=0$ when $x=0$~\cite{grimm2005model}.
\end{remark}


\begin{assumption}[Lipschitzness]\label{assumption:lipschitz}
    We Assume that ${c}_t\left(x, u\right)$ is locally Lipschitz in $x$ and $u$, $\hat{h}\left(\cdot\right)$ is locally Lipschitz in $\hat{\alpha}$.
\end{assumption} 
\Cref{assumption:lipschitz} will be used to establish the Lipschitzness of the value function $V_t\left(x ; \hat{\alpha}\right)$ with respect to the initial state $x$ and parameter $\hat{\alpha}$.

\myParagraph{Control Performance Metric} We design $u_t$ to ensure a control performance that is comparable to an optimal clairvoyant (non-causal) policy that knows the disturbance function $h$ a priori. Particularly, we consider the metric below.

\begin{definition}[Dynamic Regret]\label{def:DyReg_control}
Assume a total time horizon of operation $T$, and loss functions $c_t$, $t=1,\ldots, T$. Then, \emph{dynamic regret} is defined as
\begin{equation}
	\DReg = \sum_{t=1}^{T} c_{t}\left(x_{t}, u_{t}, h(z_t)\right)-\sum_{t=1}^{T} c_{t}\left(x_{t}^{\star}, u_{t}^{\star}, h(z_t^\star)\right),
	\label{eq:DyReg_control}
\end{equation}
where we made the dependence of the cost $c_t$ to the unknown disturbance $h$ explicit, $u_{t}^{\star}$ is the optimal control input in hindsight, \ie the optimal (non-causal) input given a priori knowledge of the unknown function $h$ and $x_{t+1}^{\star}$ is the state reached by applying the optimal control inputs $\left(u_{1}^{\star}, \; \dots, \; u_{t}^{\star}\right)$.
\end{definition}

\begin{remark}[Adaptivity of $h$]\label{rem:adaptivity}
In the definition of regret in \cref{eq:DyReg_control}, $h$ adapts (possibly differently) to the state and control sequences $(x_{1},u_1), \; \dots, \;(x_{T}, u_T)$ and $(x_{1}^{\star}, u_1^\star), \; \dots, \; (x_{T}^{\star},u^\star_T)$ since $h$ is a function of the state and the control input.
In particular, the proposed control algorithm selects control input using \cref{eq:mpc_ada_def} with estimated $\hat{h}$, and the optimal controller with the knowledge of $h$ and (potentially) with a look-ahead horizon different than N. Both controllers then evolve to the next state experiencing the real disturbances, defined by $h(\cdot)$. Since $h(\cdot)$ is a function of the state and control input, the proposed controller witnesses $h(z_t)$, while the optimal controller witnesses $h(z_t^{\star})$.
This is in contrast to previous definitions of dynamic regret, \eg \cite{agarwal2019online,zhao2022non,gradu2023adaptive,zhou2023efficient,zhou2023safecdc,zhou2023saferal} and references therein, where the optimal state $x_{t+1}^{\star}$ is reached given the same realization of $h$ as of $x_{t+1}$, \ie $x_{t+1}^{\star} = f\left(x_{t}^{\star}\right) + g\left(x_{t}^{\star}\right) u_{t}^{\star} + h\left(z_{t}\right)$.
\end{remark}

\begin{problem}[Simultaneous System Identification and Model Predictive Control (\OLMPC)]\label{prob:control}
Suppose $x_1$ is known and $x_t$ can be measured.
At each $t=1,\ldots, T$, estimate the unknown disturbance $\hat{h}\left(\cdot\right)$, and identify a control input $u_t$ by solving \cref{eq:mpc_ada_def}, such that $\DReg$ is sublinear.
\end{problem}

A sublinear dynamics regret means $\lim_{T\rightarrow\infty} \DReg/T \rightarrow 0$, which implies the algorithm asymptotically converges to the optimal (non-causal) controller.


\section{Algorithm}\label{sec:alg}

We present the algorithm for \Cref{prob:control} (\Cref{alg:MPC}). The algorithm is sketched in \Cref{fig_framework}.  The algorithm is composed of two interacting modules: (i) an \MPC module, and (ii) an online system identification module.  At each $t=1,2,\ldots,$ the \MPC module uses the estimated $\hat{h}(\cdot)$ from the system identification module to calculate the control input $u_t$. Given the current control input $u_t$ and the observed new state $x_{t+1}$, the online system identification module updates the estimate $\hat{h}(\cdot)$.  To this end, it employs online least-squares estimation via online gradient descent, where $h(\cdot)$ is parameterized as a linear combination of random Fourier features. 

To rigorously present the algorithm, we thus first provide background information on random Fourier features for approximating an $h\left(\cdot\right)$~(\Cref{subsec:RFF}), and on online gradient descent for estimation~(\Cref{subsec:OLS}). 

\subsection{Function Approximation via Random Fourier Features}\label{subsec:RFF}

We overview the randomized approximation algorithm  introduced in~\cite{boffi2022nonparametric} for approximating an $h\left(\cdot\right)$ under \Cref{assumption:func_space} and \Cref{assumption:feature_map}. 
The algorithm is based on random Fourier features~\cite{rahimi2007random,rahimi2008uniform} and their extension to vector-valued functions~\cite{brault2016random,minh2016operator}.
By being randomized, the algorithm is computationally efficient 
while retaining the expressiveness of the \RKHS with high probability.

We assume the following for the unknown disturbances or system dynamics $h\left(\cdot\right)$.

\begin{assumption}[Function Space of $h$~\cite{bach2017breaking}]\label{assumption:func_space}
$h: \mathbb{R}^{d_z} \rightarrow \mathbb{R}^{d_x}$ lies in a subspace of a Reproducing Kernel Hilbert Space 
\emph{(\RKHS)} 
 $\calH$
where the kernel $K$ is considered known~\cite{carmeli2010vector} and
 can be written via a feature map $\Phi: \mathbb{R}^{d_z} \times \Theta \rightarrow \mathbb{R}^{d_x \times d_1}$ as
    \begin{equation}
        K\left(z_1, z_2\right) = \int_\Theta \Phi\left( z_1, \theta \right) \Phi\left( z_2, \theta \right)^\top \mathrm{d} \nu(\theta),
    \end{equation}
    where $d_1 \leq d_x$, $\nu$ is a known probability measure on a measurable space $\Theta$.
\end{assumption}

The subspace defined in \Cref{assumption:func_space} will be rigorously defined later in \cref{eq:subspace_f2}.  


\begin{assumption}[Operator-Valued Bochner's Theorem~\cite{brault2016random}]\label{assumption:feature_map}
The measurable space $\Theta$ is a subset of $\mathbb{R}^{d_z+1}$ such that $\theta \in \Theta$ can be written as $\theta=\left(w, b\right)$, where $w \in \mathbb{R}^{d_z}$ and $b \in \mathbb{R}$. Also, the feature map can be factorized as $\Phi \left(z, \theta\right)= B(w) \phi\left(w^{\top} z+b\right)$, where $B: \mathbb{R}^{d_z} \rightarrow \mathbb{R}^{d_x \times d_{1}}$ and $\phi: \mathbb{R} \rightarrow[-1,1]$ is a $1$-Lipschitz function.
\end{assumption}

Examples of kernels that satisfy \Cref{assumption:func_space} and \Cref{assumption:feature_map} include the Gaussian and Laplace kernels. Additional examples of kernels can be found in~\cite[Section~5.2]{nonhoff2023relation}.

Under \Cref{assumption:func_space}, a function $h$ can be written as~\cite{bach2017breaking} 
\begin{equation}
    h\left(\cdot\right) = \int_\Theta \Phi\left(\cdot, \theta\right) \alpha(\theta) \mathrm{d}\nu (\theta),
    \label{eq:kernal_int}
\end{equation}
with $\left\| h \right\|_\calH^2 = \left\| \alpha \right\|_{L_2\left(\Theta, \nu\right)}^2 \triangleq \int_\Theta \left\| \alpha(\theta) \right\|^2 \mathrm{d} \nu(\theta)$, where $\alpha: \Theta \rightarrow \mathbb{R}^{d_1}$ is a square-integrable signed density. The corresponding Hilbert space where $h$ satisfies \cref{eq:kernal_int} is referred to as $\calF_2$~\cite{bach2017breaking,bengio2005convex}.

\Cref{eq:kernal_int} implies that $h\left(\cdot\right)$ is an integral of $\Phi\left(\cdot, \theta\right) \alpha(\theta)$ over the base measure $\nu$, thus,  we can obtain a finite-dimensional approximation of $h\left(\cdot\right)$ by 
\begin{equation}
    h\left(\cdot\right) \approx \hat{h}(\cdot;\alpha)\triangleq\frac{1}{M} \sum_{i=1}^{M} \Phi\left(\cdot, \theta_i \right) \alpha_i,
    \label{eq:kernal_approx}
\end{equation}
where $\theta_i \sim \nu$ are drawn i.i.d. from the base measure $\nu$, $\alpha_i \triangleq \alpha\left(\theta_i\right)$ are parameters to be learned, and $M$ is the number of sampling points that decides the approximation accuracy.

To rigorously establish the relationship between the number of random features $M$ and the accuracy of uniformly approximation of $h\left(\cdot\right)$, we first define $B_{\Phi}(\delta)$ as any function on a fixed compact set $\calZ \subset \mathbb{R}^{d_z}$ such that, for any $\delta \in(0,1)$,
\begin{equation}
    \mathbb{P}_{\theta \sim \nu}\left(\sup _{z \in \calZ}\|\Phi(z, \theta)\|_{\mathrm{op}}>B_{\Phi}(\delta)\right) \leqslant \delta.
\end{equation}
Then, we define a truncated $\Phi$ for any $\mu \in(0,1)$ as
\begin{equation}
    \Phi_{\mu}(z, \theta) \triangleq \Phi(z, \theta) \mathbf{1}\left\{\|\Phi(z, \theta)\|_{\mathrm{op}} \leqslant B_{\Phi}(\mu)\right\}.
\end{equation}
We aim to approximate $h\left(\cdot\right)$ over a subset of $\calF_2$, particularly,
\begin{equation}
    \calF_2 \left(B_h\right) \triangleq  \Bigg\{ h\left(\cdot\right) = \left. \int_\Theta \Phi\left(\cdot, \theta\right) \alpha(\theta) \mathrm{d}\nu (\theta)  \right\vert \alpha \in \calD\left(B_h\right) \Bigg\},
    \label{eq:subspace_f2}
\end{equation}
where $ \calD\left(B_h\right) \triangleq \{ \alpha\left(\theta\right) \mid \operatorname{{ess} \ sup}_{\theta \in \Theta} \;\| \alpha\left(\theta\right)\|\; \leq B_h \}$.  For simplicity, we will drop the dependence of $\calD$ on $B_h$.


Under \Cref{assumption:func_space} and \Cref{assumption:feature_map}, \cite{boffi2022nonparametric} extends the approximation theory of \cite{rahimi2008uniform} to vector-valued functions:
\begin{proposition}[Uniformly Approximation Error~\cite{boffi2022nonparametric}]\label{prop:approx_error}
     Assume $h \in$ $\mathcal{F}_{2}\left(B_{h}\right)$. Let $\delta \in(0,1)$ and $\mu=\frac{\delta}{2 M}$.  With probability at least $1-\delta$, there exist $\left\{\alpha_{i}\right\}_{i=1}^{M} \in \calD$ such that
\begin{equation} 
    \begin{aligned}
        & \left\| h\left(\cdot\right) - \frac{1}{M} \sum_{i=1}^{M} \Phi\left(\cdot, \theta_{i}\right) \alpha_{i}\right\|_{\infty} \\
        \leq & \frac{B_{h}}{\sqrt{M}} \Bigg[ 2 B_{\Phi}\left(\frac{\delta}{2 M}\right)\Bigg( 2 B_{\calZ} \sqrt{\mathbb{E}\left\|w_{1}\right\|^{2}}  \\ 
        & \qquad \qquad  + 2 \sqrt{d_{1}}+\sqrt{\log \frac{2}{\delta}} \Bigg) + \sqrt{\frac{\delta}{2} \mathbb{E}\|B(w)\|_{\mathrm{op}}^{2}} \Bigg] 
    \end{aligned}
    \end{equation}
where $B_{\calZ} \triangleq \sup_{z\in\calZ} \|z\|$.
\end{proposition}
    
\Cref{prop:approx_error}, therefore, indicates that the uniformly approximation error scales $\calO\left(\frac{1}{\sqrt{M}}\right)$. 

\begin{assumption}[Small Uniformly Approximation Error]\label{assump:small_approx_error}
    For a chosen $M$ value, we assume that the uniformly approximation error in \Cref{prop:approx_error} is negligible, \ie $h\left(\cdot\right)$ can be expressed as $\frac{1}{M} \sum_{i=1}^{M} \Phi\left(\cdot, \theta_{i}\right) \alpha_{i}$, where $\alpha_i \in \calD$.
\end{assumption}

\begin{remark}[Beyond Random Fourier Features]
    Beyond random Fourier features, $h$ can be learned by any models given that (i) $\hat{h}(\cdot;\alpha)$ is convex in $\alpha$, (ii) \Cref{assumption:lipschitz} and \Cref{assump:small_approx_error} hold, and (iii) $\hat{h}(\cdot;\alpha)$ and $\nabla_{\alpha}\hat{h}$ are uniformly bounded for all $\alpha \in \calD$.\footnote{The finite-dimensional approximation of $h\left(\cdot\right)$ using random Fourier features satisfies the third condition per \Cref{assumption:feature_map}.} 
    This includes, but not limited to:
    \begin{itemize}
        \item trained neural network model as feature map~\cite{shi2019neural,o2022neural}, where $h\left(z_t\right) \triangleq  \Phi\left(z_t, \theta\right) \alpha$, with $\Phi\left(\cdot, \theta\right)$ the uniformly bounded neural network model (\eg using $\tanh$ function or sigmoid function as the last layer activation function), $\theta$ the parameter of neural network learned offline, and $\alpha$ the parameter to be updated online.
        \item Koopman model~\cite{zhou2025no,jia2023evolver}, where $\Phi\left(h\left(z_t\right)\right) \triangleq  F \Phi\left(h\left(z_{t-1}\right) \right)  + H \Psi\left(h\left(z_{t-1}\right), z_t \right)$ and $h\left(z_t\right) \triangleq C\Phi\left(h\left(z_t\right)\right)$ with $\Phi\left(\cdot\right)$ locally Lipschitz nonlinear function, $\Psi\left(\cdot, \cdot \right)$ uniformly bounded nonlinear function, $C$ a constant matrix that maps $\Phi\left(h\left(\cdot\right)\right)$ back to $h\left(\cdot\right)$, $h\left(\cdot\right)$ uniformly bounded unknown disturbances, and $F$ and $H$ the parameters to be learned.
    \end{itemize}
    The resulting performance of \MPC depends on the specific models, \eg their expressiveness and generalization to different disturbances, and the computational complexity.
\end{remark}

\subsection{Online Least-Squares Estimation}\label{subsec:OLS}
Given a data point $\left( z_t, \; h\left(z_t\right) \right)$ observed at time $t$, we employ an online least-squares algorithm that updates the parameters $\hat{\alpha}_t \triangleq \left[ \alpha_{i,t}^\top, \; \dots, \; \alpha_{M,t}^\top\right]^\top$ to minimize the approximation error $l_t = \| h\left(z_t\right) - \hat{h}\left(z_t\right) \|^2$, where $ \hat{h}(\cdot) \triangleq  \frac{1}{M} \sum_{i=1}^{M} \Phi\left(\cdot, \theta_i \right) \hat{\alpha}_{i,t} $ and $\Phi\left(\cdot,\theta_i\right) $ is the random Fourier feature as in \Cref{subsec:RFF}. 
Specifically, the algorithm used the online gradient descent algorithm~(\OGD)~\cite{hazan2016introduction}. At each $t = 1, \dots, T$, it makes the steps:
\begin{itemize}
    \item Given $\left( z_t, \; h\left(z_t\right) \right)$, formulate the estimation loss function (approximation error):
            \begin{equation*}
                l_t\left(\hat{\alpha}_t\right) \triangleq \left\| h\left(z_t\right) -  \frac{1}{M} \sum_{i=1}^{M} \Phi\left(z_t, \theta_i \right) \hat{\alpha}_{i,t} \right\|^2.
            \end{equation*}
    \item Calculate the gradient of $l_t\left(\hat{\alpha}_t\right)$ with respect to $\hat{\alpha}_t$: 
            \begin{equation*}
                \nabla_t \triangleq \nabla_{\hat{\alpha}_t} l_t\left(\hat{\alpha}_t\right).
            \end{equation*}
    \item Update using gradient descent with learning rate $\eta$:
            \begin{equation*}
                \hat{\alpha}_{t+1}^\prime= \hat{\alpha}_t- \eta \nabla_t.
            \end{equation*}
    \item Project each $\hat{\alpha}_{i,t+1}^\prime$ onto $\calD$:
            \begin{equation*}
                \hat{\alpha}_{i,t+1} = \Pi_{\calD}(\hat{\alpha}_{i,t+1}^\prime) \triangleq \underset{\alpha \in \calD}{\operatorname{\textit{argmin}}}\; \| \alpha - \hat{\alpha}_{i,t+1}^\prime \|^2.
            \end{equation*}
\end{itemize}

The above online least-squares estimation algorithm enjoys an $\calO\left(\sqrt{T}\right)$ regret bound, per the regret bound of \OGD~\cite{hazan2016introduction}.

\begin{proposition}[Regret Bound of Online Least-Squares Estimation~\cite{hazan2016introduction}]\label{theorem:OGD}
    Assume $\eta=\calO\left({1}/{\sqrt{T}}\right)$.  Then,
    \begin{equation}
       \SReg\triangleq \sum_{t=1}^{T} l_t \left(\alpha_t\right) - \sum_{t=1}^{T} l_t \left(\alpha^{\star}\right)  \leq \calO\left(\sqrt{T}\right),
    \end{equation}
    where $\alpha^{\star} \triangleq \underset{\alpha \in \calD}{\operatorname{\textit{argmin}}}\;\sum_{t=1}^{T} l_t \left(\alpha\right)$ is the optimal parameter that achieves lowest cumulative loss in hindsight.
\end{proposition}

The online least-squares estimation algorithm thus asymptotically achieves the same estimation error 
as the optimal parameter $\alpha^{\star}$ since $\lim_{T\rightarrow\infty} \;\SReg/T = 0$.

\begin{remark}[Time-Varying Optimal Parameters]
    While we focus on the case of time-invariant optimal parameter $\alpha^\star$, our analysis naturally generalizes to the case of time-variant optimal parameter $\left(\alpha_1^\star, \dots, \alpha_T^\star\right)$, by using dynamic regret analysis of online least-square estimation~\cite{zinkevich2003online}.
\end{remark}

\subsection{Algorithm for \Cref{prob:control}}\label{subsec:MPC}

\setlength{\textfloatsep}{-0.1mm}
\begin{algorithm}[t]
\small
\caption{Simultaneous System Identification and Model Predictive Control.}
\begin{algorithmic}[1]
    \REQUIRE Number of random Fourier features $M$; base measure $\nu$; domain set $\calD$;  gradient descent learning rate $\eta$.
    \ENSURE At each time step $t=1,\ldots,T$, control input $u_{t}$.
    \medskip
        \STATE Initialize $x_1$, $\hat{\alpha}_{i,1} \in \calD$; 
        \STATE Randomly sample $\theta_i \sim \nu$ and formulate $\Phi\left(\cdot, \theta_i\right)$, where $i \in \{1, \dots, M\}$;
    \FOR {each time step $t = 1, \dots, T$}
    \STATE Apply control input $u_t$ by solving \cref{eq:mpc_ada_def} with $\hat{h}(\cdot) \triangleq  \frac{1}{M} \sum_{i=1}^{M} \Phi\left(\cdot, \theta_i \right) \hat{\alpha}_{i,t} $;
        \STATE Observe state $x_{t+1}$, and calculate disturbance via $h\left(z_t\right) = x_{t+1} - f(x_{t}) - g(x_{t}) u_{t}$;
        \STATE Formulate estimation loss $l_t\left(\hat{\alpha}_t\right) \triangleq \| h\left(z_t\right) - \frac{1}{M} \sum_{i=1}^{M} \Phi\left(z_t, \theta_i \right) \hat{\alpha}_{i,t} \|^2$;
        \STATE Calculate gradient $\nabla_t \triangleq \nabla_{\hat{\alpha}_t} l_t\left(\hat{\alpha}_t\right)$;
        \STATE Update $\hat{\alpha}_{t+1}^\prime= \hat{\alpha}_t- \eta \nabla_t$;
        \STATE Project  $\hat{\alpha}_{i,t+1}^\prime$ onto $\calD$, \ie $\hat{\alpha}_{i,t+1} = \Pi_{\calD}(\hat{\alpha}_{i,t+1}^\prime)$, for $i \in \{1, \; \dots, \; M\}$;
        \ENDFOR
\end{algorithmic}\label{alg:MPC}
\end{algorithm}

We describe the algorithm for \OLMPC. The pseudo-code is in \Cref{alg:MPC}.  The algorithm is composed of three steps, initialization, control, and estimation, where the control and estimation steps influence each other at each time steps (Fig.~\ref{fig_framework}):

\begin{itemize}
    \item \textit{Initialization steps:} \Cref{alg:MPC} first initializes the system state $x_1$ and parameter $\hat{\alpha}_1 \in \calD$~(line 1). Then given the number of random Fourier features, \Cref{alg:MPC} randomly samples $\theta_i$ and formulates $\Phi\left(\cdot, \theta_i\right)$, where $i \in \{1, \dots, M\}$~(line 2).

    \item \textit{Control steps:} Then, at each~$t$, given the current estimate $\hat{h}(\cdot) \triangleq  \frac{1}{M} \sum_{i=1}^{M} \Phi\left(\cdot, \theta_i \right) \hat{\alpha}_{i,t}$, \Cref{alg:MPC} applies the control inputs $u_t$ obtained by solving \cref{eq:mpc_ada_def}~(line 4).

    \item \textit{Estimation steps:}
    The system then evolves to state $x_{t+1}$, and, $h\left(z_t\right)$ is calculated upon observing $x_{t+1}$~(line 5).
    Afterwards, the algorithm formulates the loss $l_t\left(\hat{\alpha}_t\right) \triangleq \| h\left(z_t\right) - \sum_{i=1}^{M} \Phi\left(z_t, \theta_i \right) \hat{\alpha}_{i,t} \|^2$~(line 6), and calculates the gradient $\nabla_t \triangleq \nabla_{\hat{\alpha}_t} l_t\left(\hat{\alpha}_t\right)$~(line 7). 
    \Cref{alg:MPC} then updates the parameter $\hat{\alpha}_t$ to $\hat{\alpha}_{t+1}^\prime$~(line 8) and, finally, projects each $\hat{\alpha}_{i,t+1}^\prime$ back to the domain set $\calD$~(line 9).
\end{itemize}

\section{No-Regret Guarantee}\label{sec:Reg}

We present the sublinear regret bound of \Cref{alg:MPC}.




\begin{theorem}[No-Regret]\label{theorem:regret_OLMPC}
Assume \Cref{alg:MPC}'s learning rate is $\eta=\calO\left({1}/{\sqrt{T}}\right)$.  Then, \Cref{alg:MPC} achieves  
\begin{equation}
    \DReg \leq \calO\left(T^{\frac{3}{4}}\right).
    \label{eq:theorem_regret_OLMPC}
\end{equation}


\end{theorem}

\Cref{theorem:regret_OLMPC} serves as a finite-time performance guarantee as well as implies that \Cref{alg:MPC} converges to the optimal (non-causal) control policy since $\lim_{T\rightarrow\infty}\DReg / T \rightarrow 0$. 

We have no proof that the bound in \cref{eq:theorem_regret_OLMPC} is tight.  We will consider this analysis in our future work. 
Notably, the bound matches that of the adaptive control method in \cite[Corollary~5.7]{boffi2021regret}. 

\begin{remark}[Extension to Systems with Unmodeled Noise]
In \Cref{corollary:regret_OLMPC_extension}~(\Cref{app:extension}), we extend the result of \Cref{theorem:regret_OLMPC} to systems in the form of $x_{t+1} = f\left(x_{t}\right) + g\left(x_{t}\right) u_{t} + h\left(z_{t}\right) + w_{t}$, where $w_{t}$ represents, \eg \space process noise or estimation error of $h\left(z_{t}\right)-\hat{h}\left(z_{t}\right)$. 
Specifically, the regret bound becomes
\begin{equation}
    \DReg \leq \calO\left(T^{\frac{3}{4}}\right) + L \sqrt{ T\sum_{t=1}^{T}\|w_{t}\|^2 },
\end{equation}
where the regret bound has an additional term that depends on the energy of the noise sequence $(w_1, \dots, w_T)$, \ie $\sum_{t=1}^{T}\|w_{t}\|^2$. Specifically, when the energy is less than $\calO(T)$, we achieve sublinear regret.
\end{remark}

\begin{remark}[Stability Analysis]
In \Cref{app:stability}, we analyze the stability of \Cref{alg:MPC} under the assumption of the stability of the nominal system, \ie when \Cref{assumption:stability} only holds for the nominal system~(\cref{eq:mpc_def_dyn}, \ie $\alpha = 0$). We show that the state $x_t$ remain bounded under $\Cref{alg:MPC}$, where the bound scales with the size of domain set $\calD$ and the bound of $\|h(\cdot)\|$.
\end{remark}

\section{Numerical Evaluations}\label{sec:exp-sim}

We evaluate \Cref{alg:MPC} in extensive simulated scenarios of control under uncertainty, where the controller aims to track a reference setpoint/trajectory despite unknown disturbance. 
We first detail how \Cref{alg:MPC} is implemented empirically in \Cref{subsec:implementation}.
For the experiments, we first consider a cart-pole aiming to stabilize around a setpoint despite inaccurate model parameters, \ie inaccurate cart mass, pole mass, and pole length~(\Cref{subsec:sim-1}).
Then, we consider the same setup of cart-pole and conduct the parameter sensitivity analysis, \ie how the number of random Fourier features and the learning rate affect the performance of \Cref{alg:MPC}~(\Cref{subsec:sim-2}).
We consider a quadrotor aiming to track given reference trajectories subject to unknown aerodynamic forces~(\Cref{subsec:sim-3}).

\subsection{Empirical Implementation of \Cref{alg:MPC}}\label{subsec:implementation}
We employ \Cref{alg:MPC} as follows:
\begin{itemize}
    \item We use the quadratic cost function in \MPC since it is widely applied and intuitive to tune, despite the difficulty of verifying if \Cref{assumption:stability} holds.
    \item We assume $h\left(\cdot\right)$ can be fitted via $\frac{1}{M} \sum_{i=1}^{M} \Phi\left(\cdot, \theta_{i}\right) \alpha_{i}$, per \Cref{assump:small_approx_error}.
    \item We choose the kernel $K$ to be the Gaussian kernel. For its randomized approximation, we obtain $\{\theta_i\}_{i=1}^{M}$ i.i.d.~by sampling $w_i$ from a Gaussian distribution, and $b_i$ from an uniform distribution from $[0, \; 2\pi]$~\cite{rahimi2007random}. Unless specified, we sample $w_i$ from a standard Gaussian distribution.
    \item We use different sets of parameters for each entry of $h\left(\cdot\right)$: Suppose $h\left(\cdot\right) = [ h_1\left(\cdot\right), \; h_2\left(\cdot\right)]^\top \in \mathbb{R}^2$, then we approximate $h_1\left(\cdot\right)$ and $h_2\left(\cdot\right)$ by $\frac{1}{M} \sum_{i=1}^{M} \Phi\left(\cdot, \theta_{i}\right) \alpha_{i}^{(1)}$ and $\frac{1}{M} \sum_{i=1}^{M} \Phi\left(\cdot, \theta_{i}\right) \alpha_{i}^{(2)}$, respectively; \ie $h_1\left(\cdot\right)$ and $h_2\left(\cdot\right)$ share the same set of random Fourier features but have different parameters $\alpha$. We make this design choice to reduce the feature map to $\Phi \left(z, \theta\right)= \phi\left(w^{\top} z+b\right)$ and avoid the tuning of $B\left(w\right)$.

\end{itemize}

\subsection{Cart-Pole Scenario}\label{subsec:sim-1}

\begin{figure*}[t]
    \centering
    \subfigure[Average stabilization error.]{\includegraphics[width=0.32\textwidth]{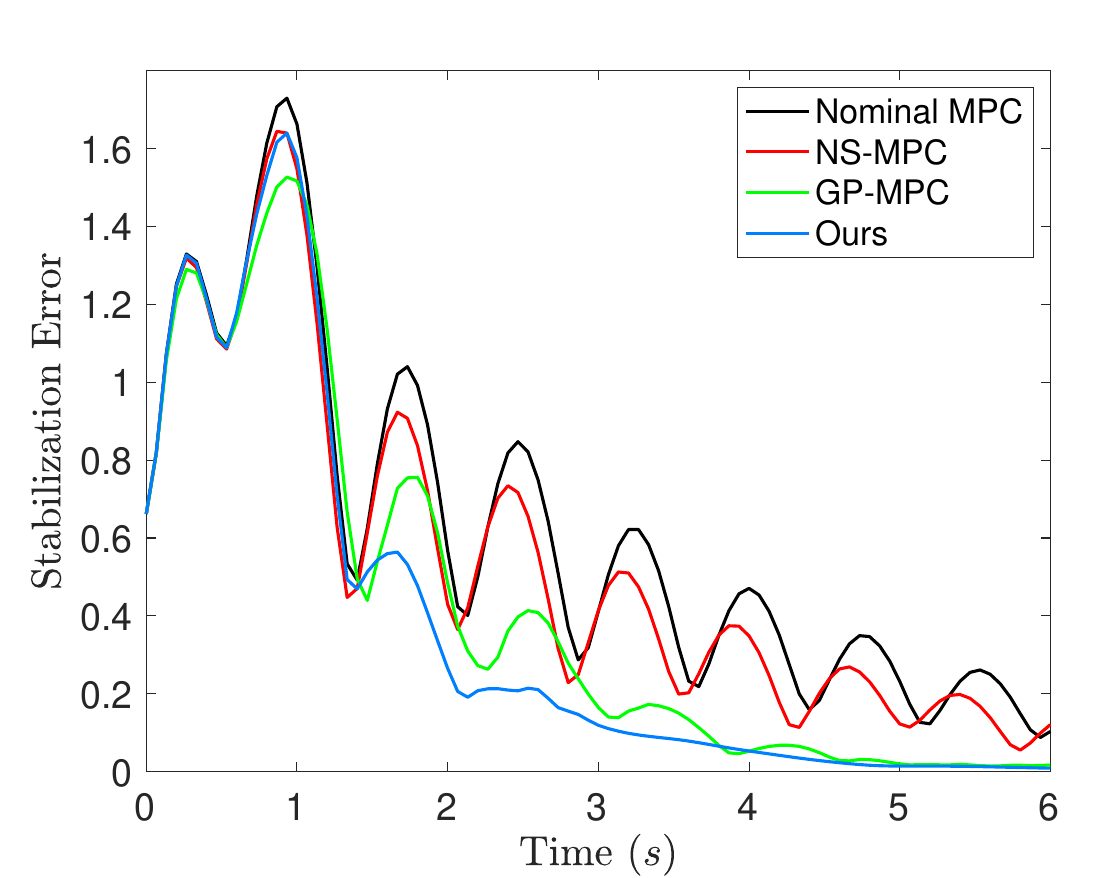}\label{fig_cartpole_stab}}	
    \subfigure[Sample trajectory.]{\includegraphics[width=0.32\textwidth]{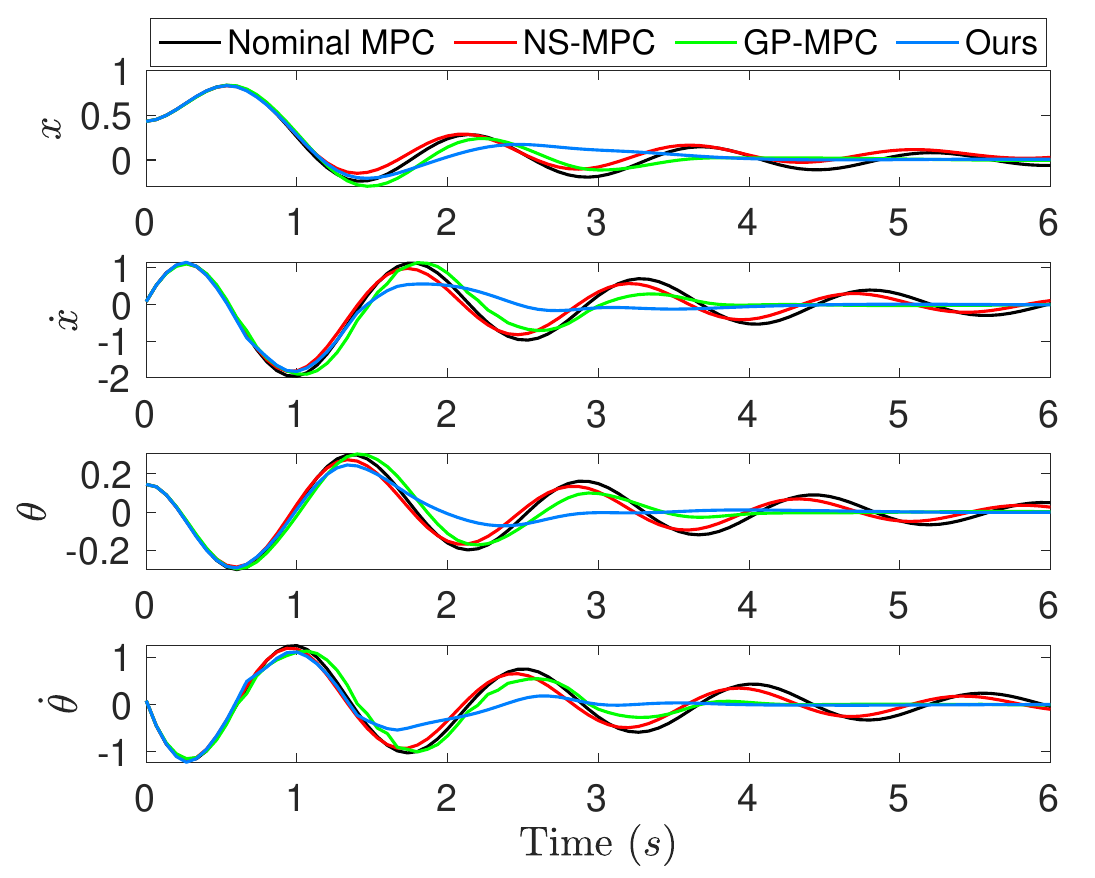}\label{fig_cartpole_traj}}
    \subfigure[Prediction error.]{\includegraphics[width=0.32\textwidth]{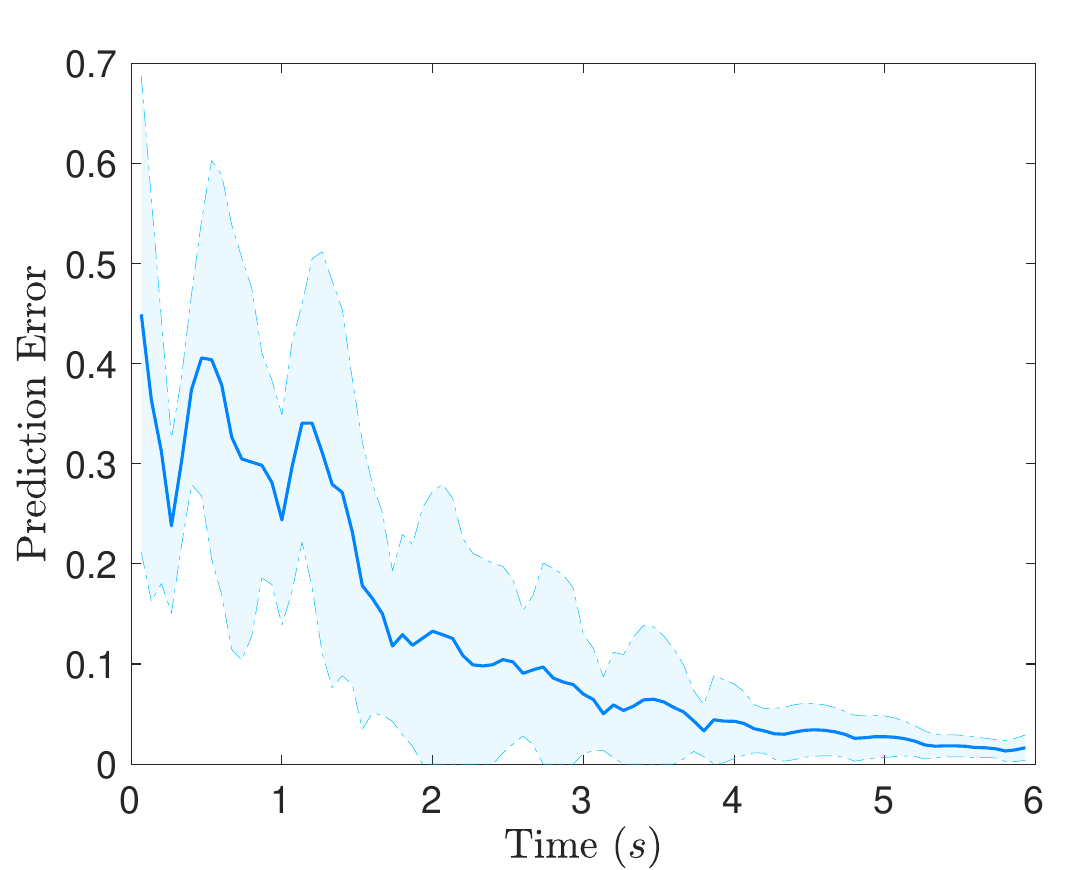}\label{fig_cartpole_pred}}	
    \caption{\textbf{Simulation Results of the Cart-Pole Stabilization Experiment in \Cref{subsec:sim-1}.} (a) and (b) demonstrate that \Cref{alg:MPC} achieves stabilization in the least time among all tested algorithms. \GPMPC comes second but it incurs a larger deviation from the stabilization goal $(0,0,0,0)$ than \Cref{alg:MPC}.
    \NSMPC and \NMPC have similar performance, showing that the state-of-the-art non-stochastic control methods are insufficient when the unknown disturbance is adaptive. (c) shows that as \Cref{alg:MPC} collects more data, the prediction error decreases.}
    \label{fig_cartpole_stabilization}
\end{figure*}

\begin{figure*}[t]
    \centering
    \subfigure[Average cumulative stabilization error.]{\includegraphics[width=0.32\textwidth]{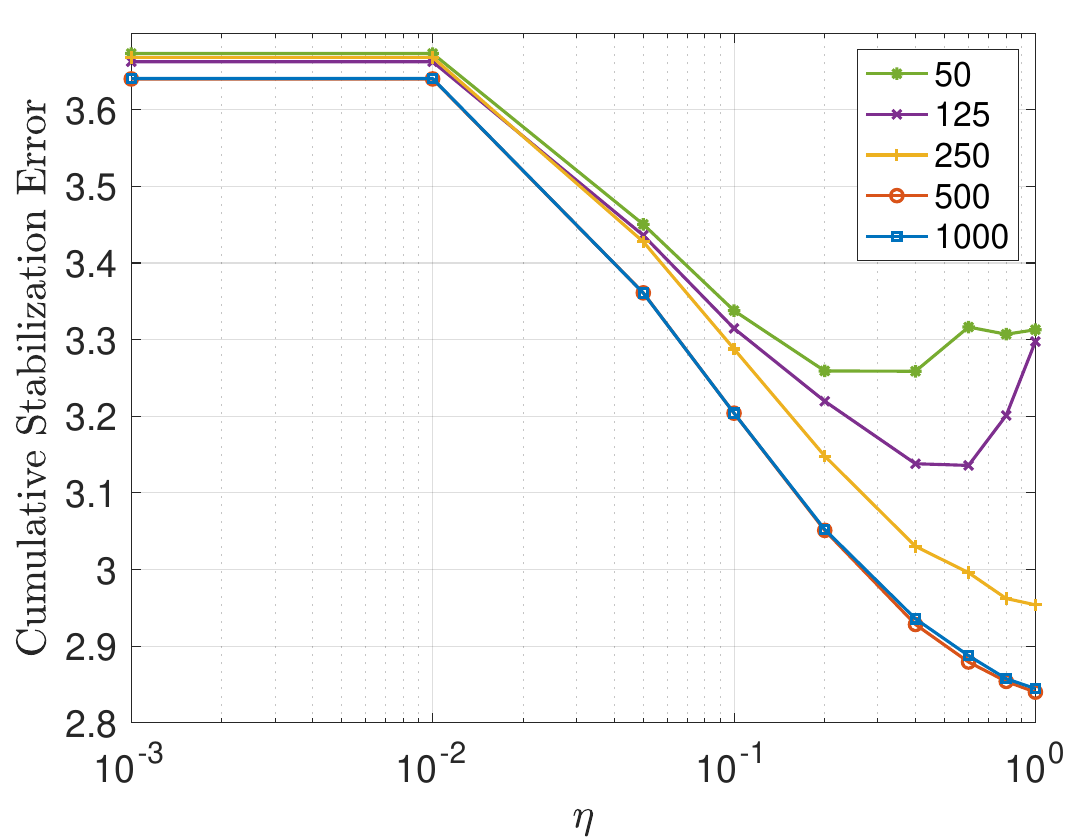}\label{fig_sensitivity_all}}	
    \subfigure[Average stabilization error with various number of random features using $\eta=0.4$.]{\includegraphics[width=0.32\textwidth]{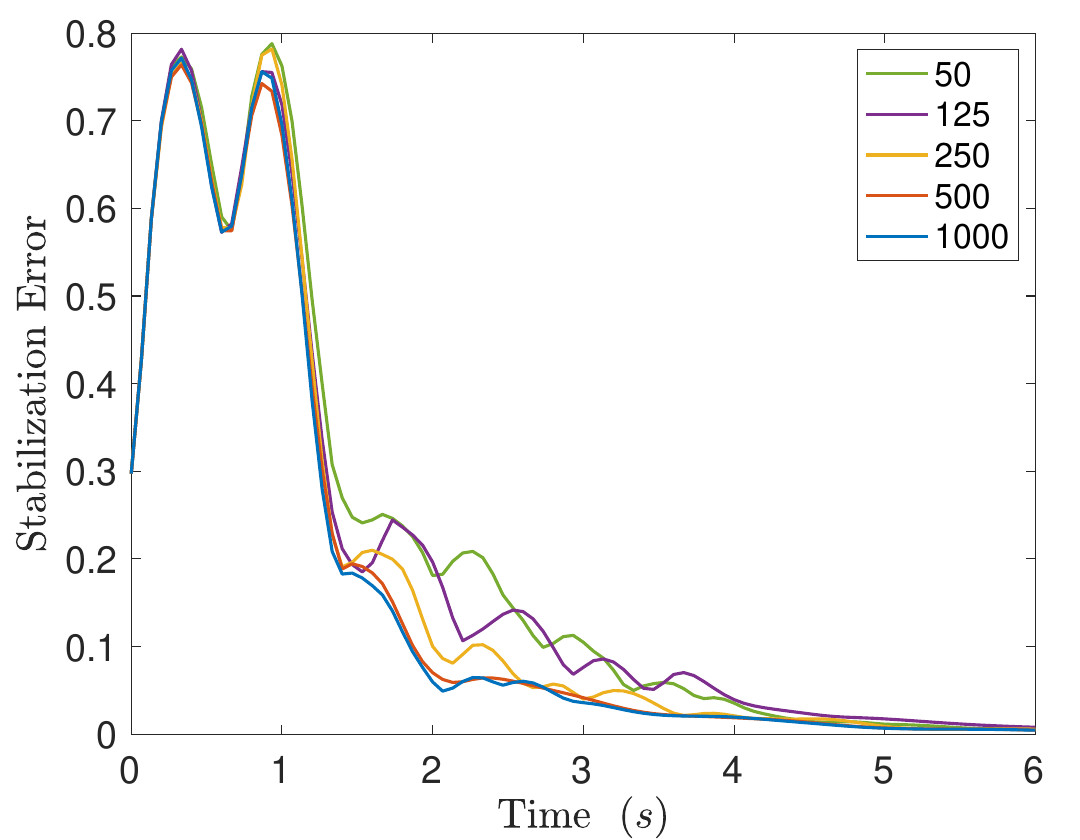}\label{fig_sensitivity_rf}}
    \subfigure[Average stabilization error with various learning rate $\eta$s using $M=250$ random Fourier features.]{\includegraphics[width=0.32\textwidth]{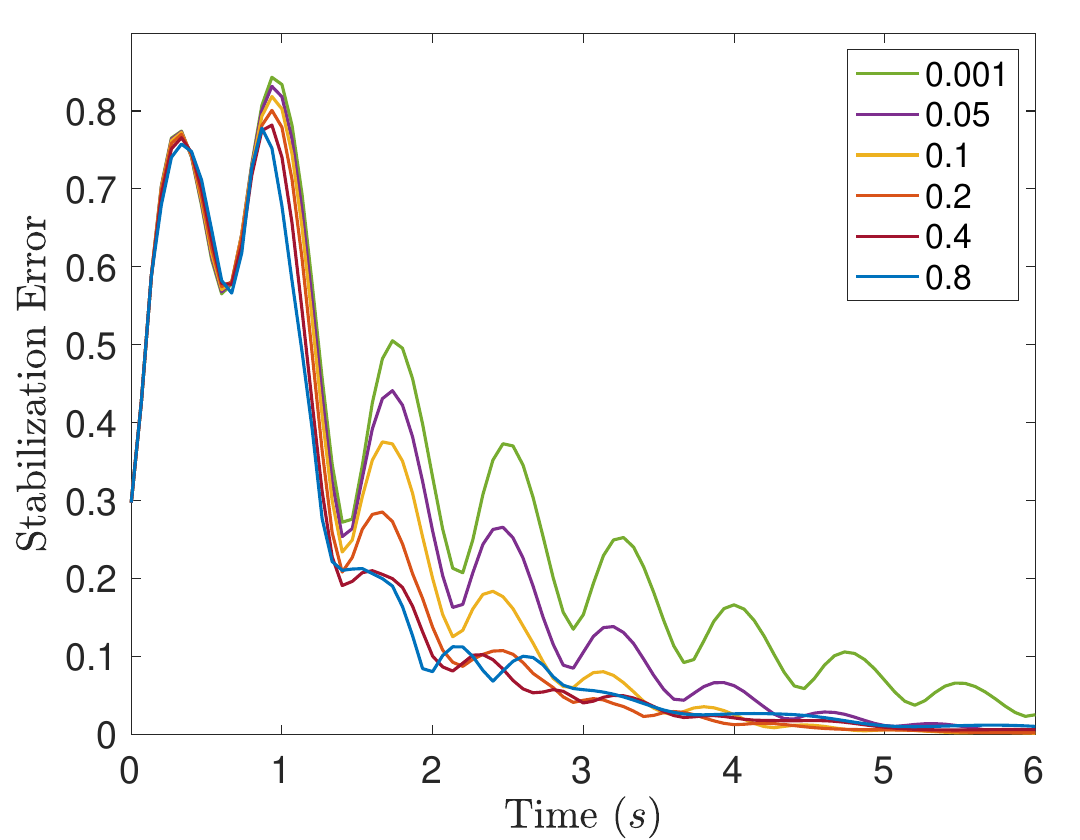}\label{fig_sensitivity_lr}}	
    \caption{\textbf{Simulation Results for the Sensitivity Analysis in \Cref{subsec:sim-2} over the Cart-Pole System.} The results suggest that large $M$ and $\eta$ achieves better performance. However, $M$ cannot be arbitrarily large as it increases the computational complexity of solving \cref{eq:mpc_ada_def}, shown in \Cref{table:cartpole-sensitivity}.}
    \label{fig_sensitivity}
\end{figure*}

\myParagraph{Simulation Setup} We consider a cart-pole system, where a cart of mass $m_c$ connects via a prismatic joint to a $1D$
track, while a pole of mass $m_p$ and length $2l$ is hinged to the cart.
The state vector $\myx$ includes the horizontal position of the cart $x$, the velocity of the cart $\dot{x}$,  the angle of the pole with respect to vertical $\theta$, and the angular velocity of the pole $\dot{\theta}$.
The control input is the force $F$ applied to the center of mass of the cart. 
The goal of the cart-pole is to stabilize at $(x, \dot{x}, \theta, \dot{\theta}) = (0,0,0,0)$. The dynamics of the cart-pole are~\cite{florian2007correct}:
\begin{equation}
    \begin{aligned}
        \ddot{x} &= \frac{m_p l \left(\dot{\theta}^2 \sin\theta - \ddot{\theta} \cos\theta \right) + F}{m_c + m_p}, \\
        \ddot{\theta} &= \frac{g \sin\theta + \cos\theta \left( \frac{- m_p l \dot{\theta}^2 \sin\theta - F}{m_c + m_p} \right) }{l \left( \frac{4}{3} - \frac{m_p \cos^2\theta}{m_c + m_p} \right)} ,
    \end{aligned}
\end{equation}
where $g$ is the acceleration of gravity.

To control the system, we will employ \MPC at $15Hz$ with a look-ahead horizon $N=20$. We use quadratic cost functions with $Q = \diag{[5.0,\; 0.1,\; 5.0,\; 0.1]}$ and $R=0.1$. We use the fourth-order Runge-Kutta (RK4) method~\cite{kloeden1992stochastic} for discretizing the above continuous-time dynamics. 
The true system parameters are $m_c=1.0$, $m_p=0.1$, and $l=0.5$ but the parameters for the nominal system dynamics are scaled to $75\%$ of the said true values.
We use $z_t = [x_t^\top \ u_t^\top]^\top$.
We use $M=75$ random Fourier features and $\eta=0.25$, and initialize $\hat{\alpha}$ as a zero vector ---an ablation study follows in the next section. 
We simulate the setting for $6s$ and perform the simulation $10$ times with random initialization sampled uniformly from $x \in [-1,\; 1]$, $\dot{x} \in [-0.1,\; 0.1]$, $\theta \in [-0.2,\; 0.2]$, $\dot{\theta} \in [-0.1,\; 0.1]$. 
The simulation environment is based on~\cite{yuan2022safe} in Pybullet~\cite{coumans2016pybullet}.
We use CasADi~\cite{andersson2019casadi} to solve \cref{eq:mpc_ada_def}.



\begin{table}[t]
    \captionsetup{font=small}
    \centering
    \caption{\textbf{Computational Performance across Different Methods for the Cart-Pole System in \Cref{subsec:sim-1}.} The table reports the average and standard deviation of computational time in millisecond. }
     \label{table:cartpole}
     \resizebox{.85\columnwidth}{!}{
     {
     \begin{tabular}{ccccc}
     \toprule
        & \MPC & \NSMPC & \GPMPC  & Ours \cr
    \midrule
        Time ($ms$) & $9.63 \pm 3.34$  &  $9.94 \pm 3.39$ & $24.28 \pm 38.24$  & $16.28 \pm 21.77$  \cr
    \bottomrule
    \end{tabular}}
     }
     \vspace{4mm}
\end{table}


\myParagraph{Compared Algorithms} 
We compare \Cref{alg:MPC} with an \MPC  that uses the (wrong) nominal system parameters (\NMPC), the non-stochastic \MPC (\NSMPC)~\cite{zhou2023saferal}, and the Gaussian process \MPC (\GPMPC)~\cite{hewing2019cautious}.
In more detail, the \NMPC uses the nominal dynamics to select control input by solving \cref{eq:mpc_def}.
The \NSMPC augments the \NMPC with an additional control $v_t$ input that is updated by running the online gradient descent algorithm to minimize the state cost $\myx_{t+1} Q \myx_{t+1}^\top$. 
The \GPMPC learns $\hat{h}\left(\cdot\right)$ with a sparse Gaussian process~(\GP)~\cite{quinonero2005unifying,snelson2005sparse} whose data points are collected online, \ie~\GP fixes its hyperparameters and collects data points $\left(z_t, h\left(z_t\right)\right)$ online.

\myParagraph{Performance Metric} 
We evaluate the performance of \NMPC, \NSMPC, \GPMPC, and \Cref{alg:MPC} in terms of their stabilization error $\|\myx_t\|^2$ and computational time. Also, we evaluate the prediction accuracy of \Cref{alg:MPC} as we collect more data online via estimation error $e_t$.

\begin{table}[t]
    \captionsetup{font=small}
    \centering
    \caption{\textbf{Computational Performance across  Different $M$ for our Method over the Cart-Pole System using $\eta=0.4$ in \Cref{subsec:sim-2}.} The table reports the average value and standard deviation of computational time in milliseconds. }
     \label{table:cartpole-sensitivity}
     \resizebox{1\columnwidth}{!}{
     {
     \begin{tabular}{cccccc}
     \toprule
        $M$ & $50$ & $125$ & $250$  & $500$ & $1000$ \cr
    \midrule
        Time ($ms$) & $14.32 \pm 18.80$  &  $23.84 \pm 40.36$ & $31.32 \pm 71.47$  & $53.16 \pm 148.94$  &  $96.39 \pm 299.99$ \cr
    \bottomrule
    \end{tabular}}
     }
     \vspace{5mm}
\end{table}

\myParagraph{Results} 
The results are given in \Cref{fig_cartpole_stabilization} and \Cref{table:cartpole}.
\Cref{fig_cartpole_stab} and \Cref{fig_cartpole_traj} demonstrate that \Cref{alg:MPC} achieves stabilization the fastest. 
\GPMPC comes second but it incurs a larger deviation from the stabilization goal $(0,0,0,0)$ than \Cref{alg:MPC}.
\NSMPC and \NMPC have similar performance, showing that the state-of-the-art non-stochastic control methods are insufficient when the unknown disturbance is adaptive. 
\Cref{fig_cartpole_pred} shows a fast convergence of the estimation error: the error decreases to around $0.1$ within $2s$ (\ie less than $30$ iterations), which benefits our stabilization goal as shown in \Cref{fig_cartpole_traj}.
In \Cref{table:cartpole}, \NMPC is the most computationally efficient, followed by \NSMPC since the online gradient descent update of $v_t$ only needs one projection step. Our method maintains the expressiveness of random Fourier features such that it learns the unknown dynamics due to incorrect model parameters and is computationally more efficient than \GPMPC.

\subsection{Sensitivity Analysis over Cart-Pole Scenario}\label{subsec:sim-2}
\myParagraph{Simulation Setup} 
We consider the same simulation setup as in \cref{subsec:sim-1}. In addition, we run the simulation with various number of random Fourier features and learning rate: we use $M \in \{50,\;125,\;250,\;500,\;1000\}$, and $\eta \in \{0.001,\;0.01,\;0.05,\;0.1,\;0.2,\;0.4,\;0.6,\;0.8,\;1\}$.

\myParagraph{Performance Metric} 
We evaluate the performance of \Cref{alg:MPC} in terms of the stabilization error $\|\myx_t\|^2$ and cumulative stabilization error $\sum_{t=1}^{T}\|\myx_t\|^2$. Also, we consider the computational time for calculating control inputs with different $M$ while fixing $\eta$.

\myParagraph{Results} 
The results are given in \Cref{fig_sensitivity} and \Cref{table:cartpole-sensitivity}.
First, we notice that \Cref{alg:MPC} has similar performance as \NMPC for small $\eta$, \ie $\eta\leq0.01$, regardless of $M$. The reason is that very small $\eta$ will keep $\hat{\alpha}$ to stay around zero.
\Cref{fig_sensitivity} suggests that large $M$ and $\eta$ achieve better performance. 
For $M=50$ and $M=125$, increasing $\eta$ to $1$ causes worse performance, due to overshoot behaviors in estimating $\hat{\alpha}$ and, therefore, steady-state error. 
This can be avoided by increasing $M$.
\Cref{fig_sensitivity_all} shows that the cumulative stabilization error cannot be improved after $M\geq500$. 
Further, \Cref{table:cartpole-sensitivity} shows that increasing $M$ makes the optimization problem in \cref{eq:mpc_ada_def} computationally expensive and prevents the method from real-time implementations, as the frequency of \MPC is less than $3~Hz$ with $M\geq250$ in Python. 

Another tuning parameter is the standard deviation of the Gaussian distribution from which $w_i$ are sampled. This depends on the magnitude of feature vector $z_t$, since they multiply together per the definition of feature map $\Phi \left(z, \theta\right)= \phi\left(w^{\top} z+b\right)$. Hence, if $z_t$ has a large magnitude, the standard deviation might need to be small. Alternatively, we can normalize $z_t$ based on its magnitude and start tuning with a standard Gaussian distribution.

\subsection{Quadrotor Scenario}\label{subsec:sim-3}

\begin{figure*}[t]
    \centering
    \subfigure[Circle trajectory.]{\includegraphics[width=0.245\textwidth]{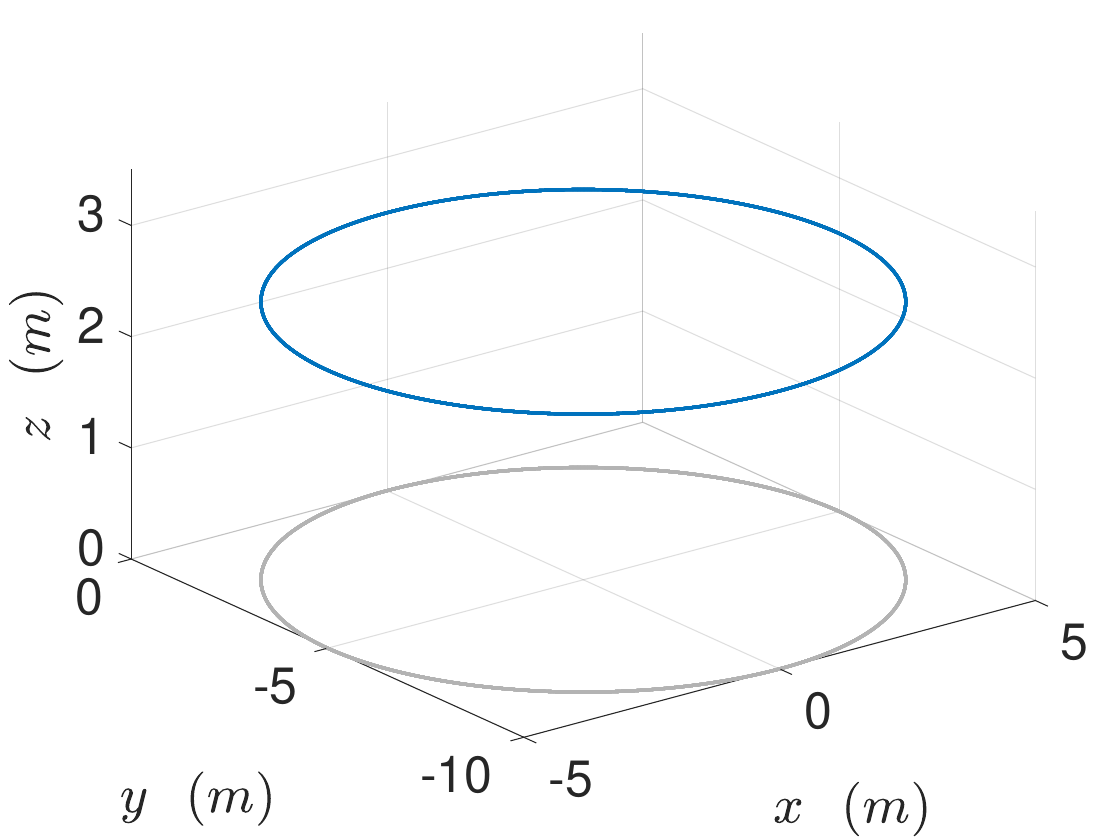}\label{fig_quad_ref_circle}}	
    \subfigure[Wrapped Circle trajectory.]{\includegraphics[width=0.245\textwidth]{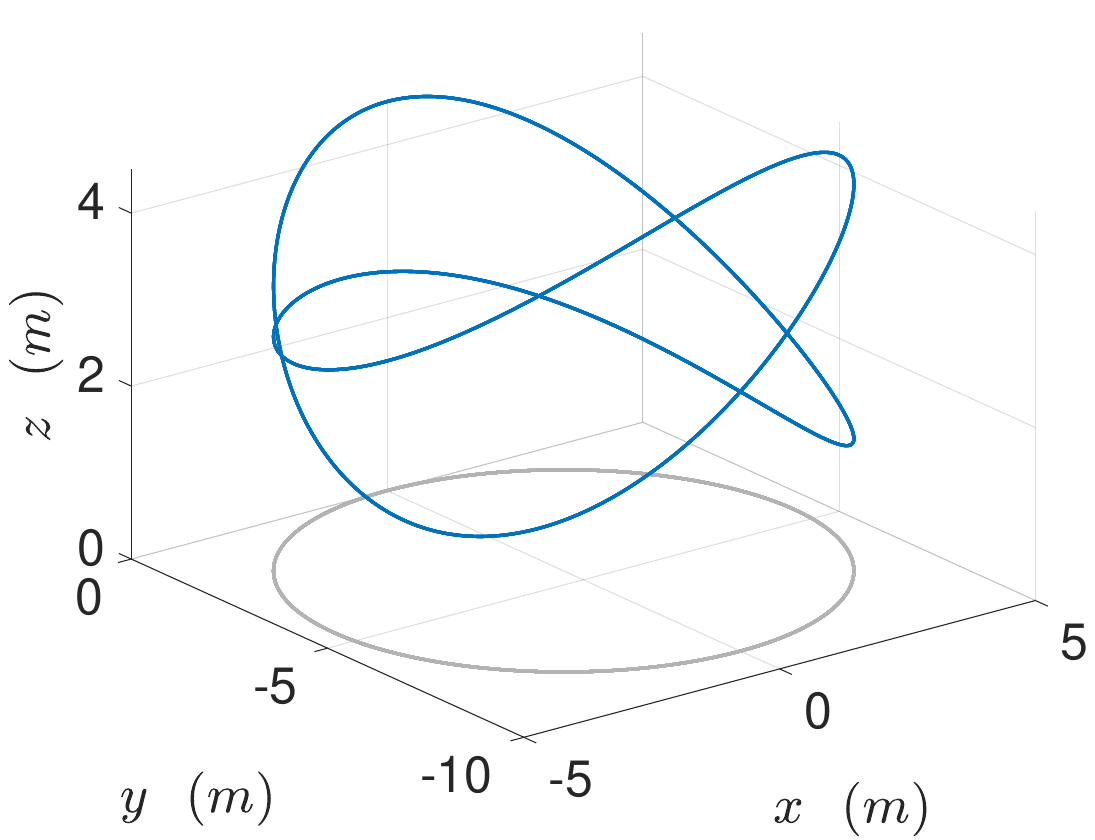}\label{fig_quad_ref_wrapped_circle}}
    \subfigure[Lemniscate trajectory.]{\includegraphics[width=0.245\textwidth]{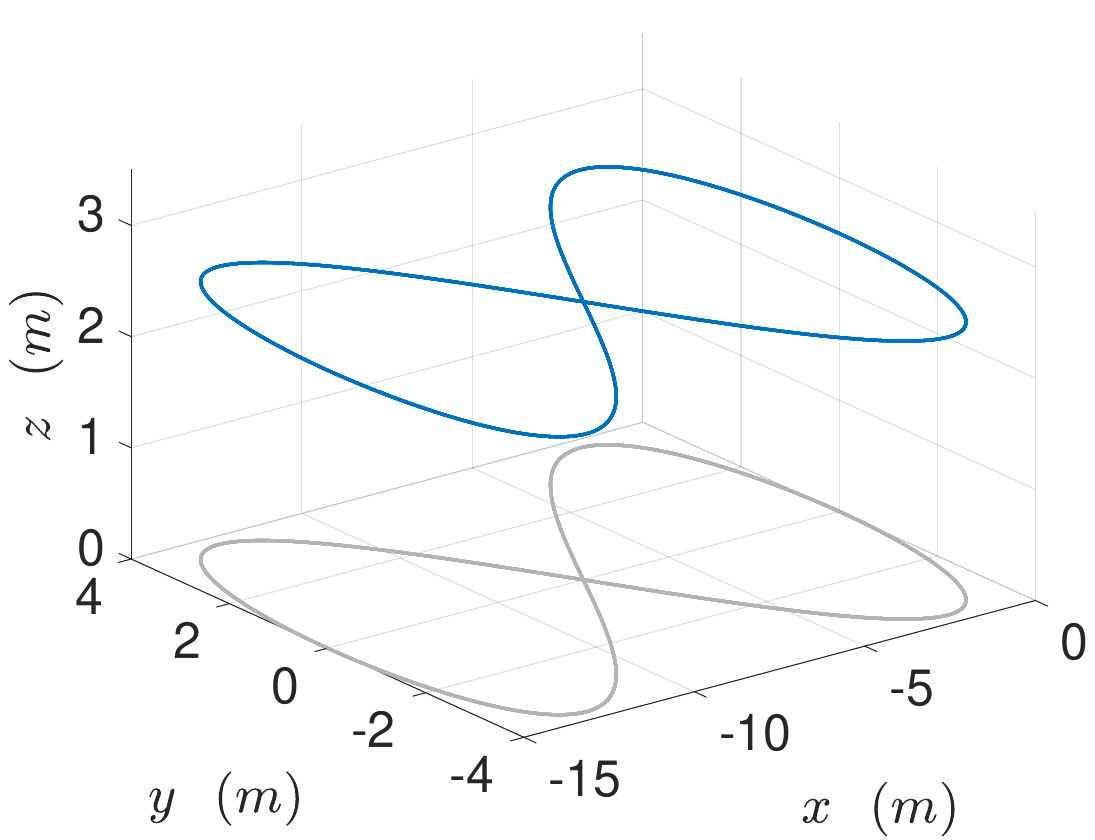}\label{fig_quad_ref_lemniscate}}	
    \subfigure[Wrapped Lemniscate trajectory.]{\includegraphics[width=0.245\textwidth]{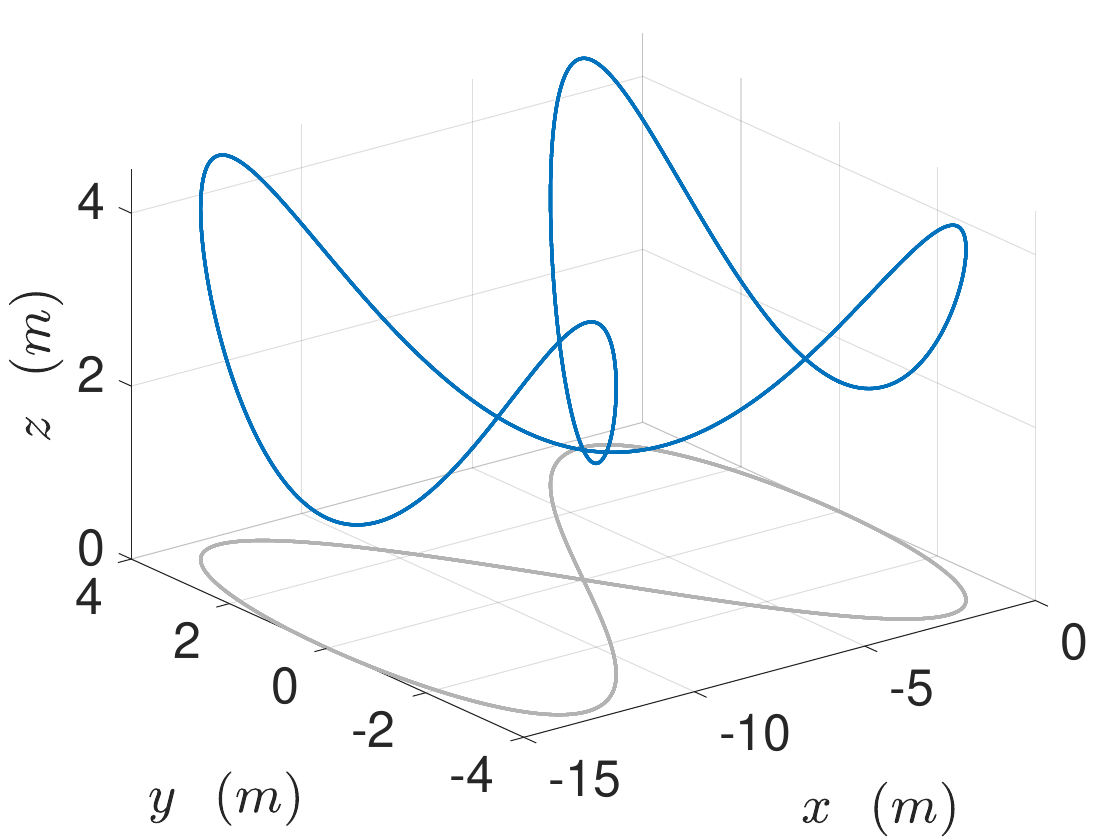}\label{fig_quad_ref_wrapped_lemniscate}}	
    \caption{\textbf{Reference Trajectory for the Quadrotor Experiments in \Cref{subsec:sim-3}.} The blue lines represent the reference trajectories in $3D$. The gray lines are the projection of reference trajectories onto the ground.}
    \label{fig_quad_ref_traj}
\end{figure*}

\begin{figure*}[t]
    \centering
    \subfigure[Circle trajectory.]{\includegraphics[width=0.245\textwidth]{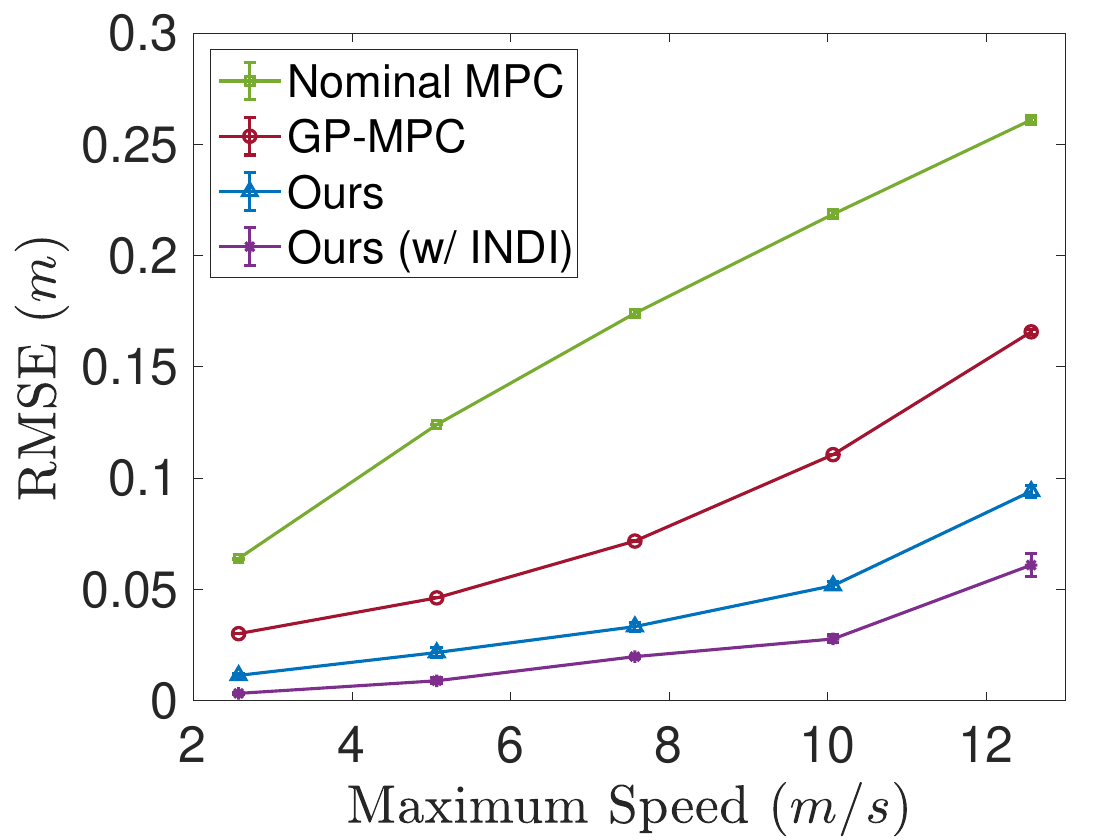}\label{fig_quad_track_circle}}	
    \subfigure[Wrapped Circle trajectory.]{\includegraphics[width=0.245\textwidth]{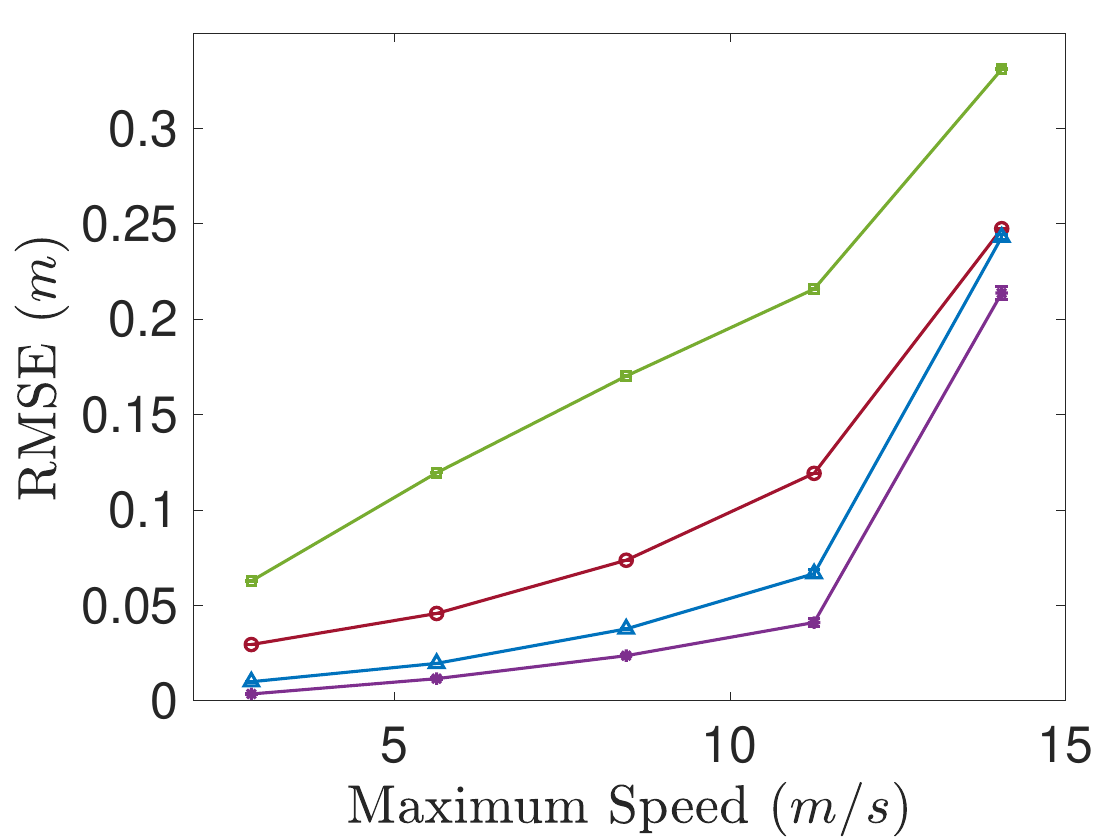}\label{fig_quad_track_wrapped_circle}}
    \subfigure[Lemniscate trajectory.]{\includegraphics[width=0.245\textwidth]{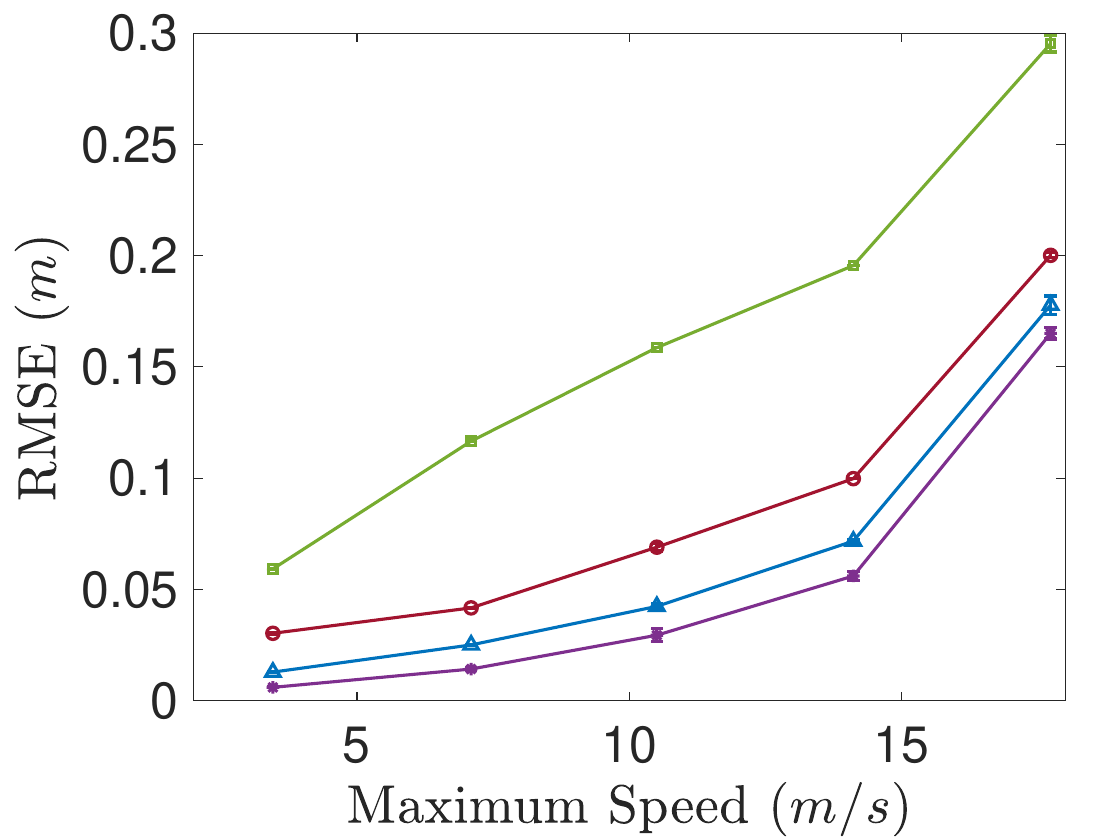}\label{fig_quad_track_lemniscate}}	
    \subfigure[Wrapped Lemniscate trajectory.]{\includegraphics[width=0.245\textwidth]{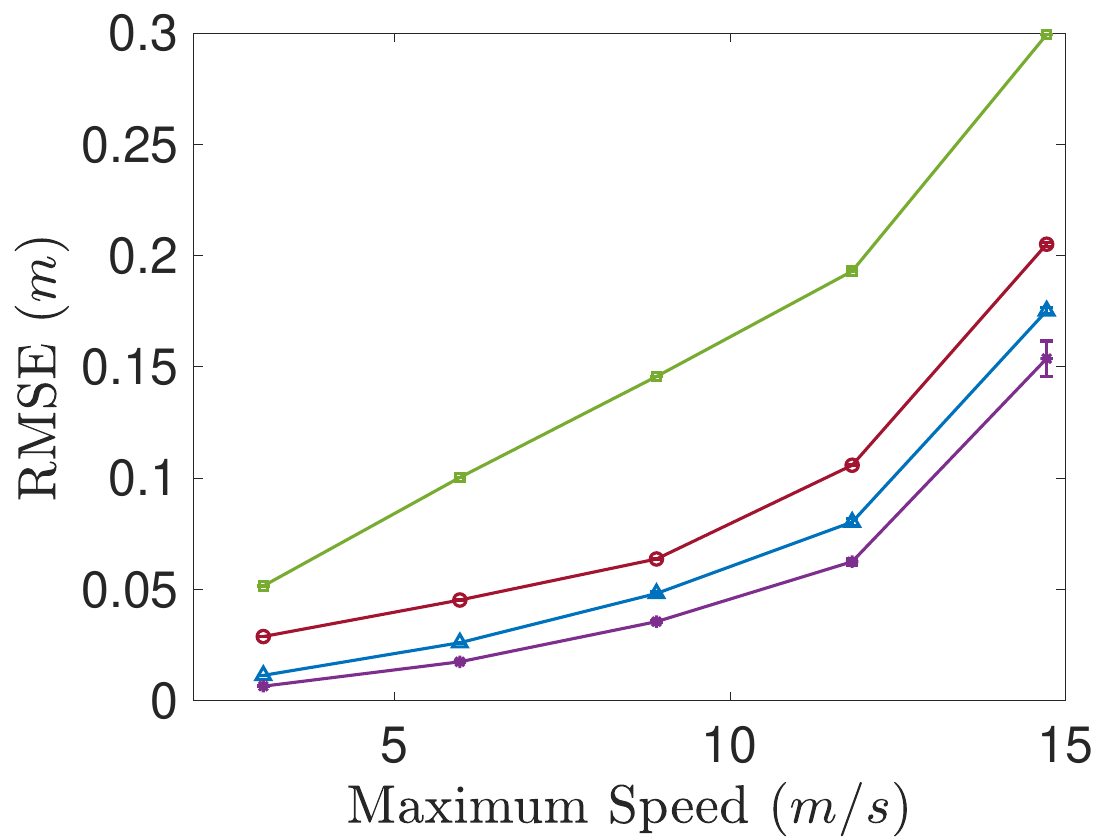}\label{fig_quad_track_wrapped_lemniscate}}	
    \caption{\textbf{Tracking Performance Comparison for the Quadrotor Experiments in \Cref{subsec:sim-3}.} \Cref{alg:MPC} demonstrates improved performance compared to \NMPC and \GPMPC in terms of tracking error over all tested reference trajectories and maximal speeds. \Cref{alg:MPC} with \INDI achieves the best performance as \INDI provides better tracking in attitude dynamics.}
    \label{fig_quad_track}
\end{figure*}

\begin{figure}[t]
    \centering
    \subfigure[Sample trajectories in tracking a Circle trajectory with $v_m=10.07 \ m/s$.]{\includegraphics[width=0.4\textwidth]{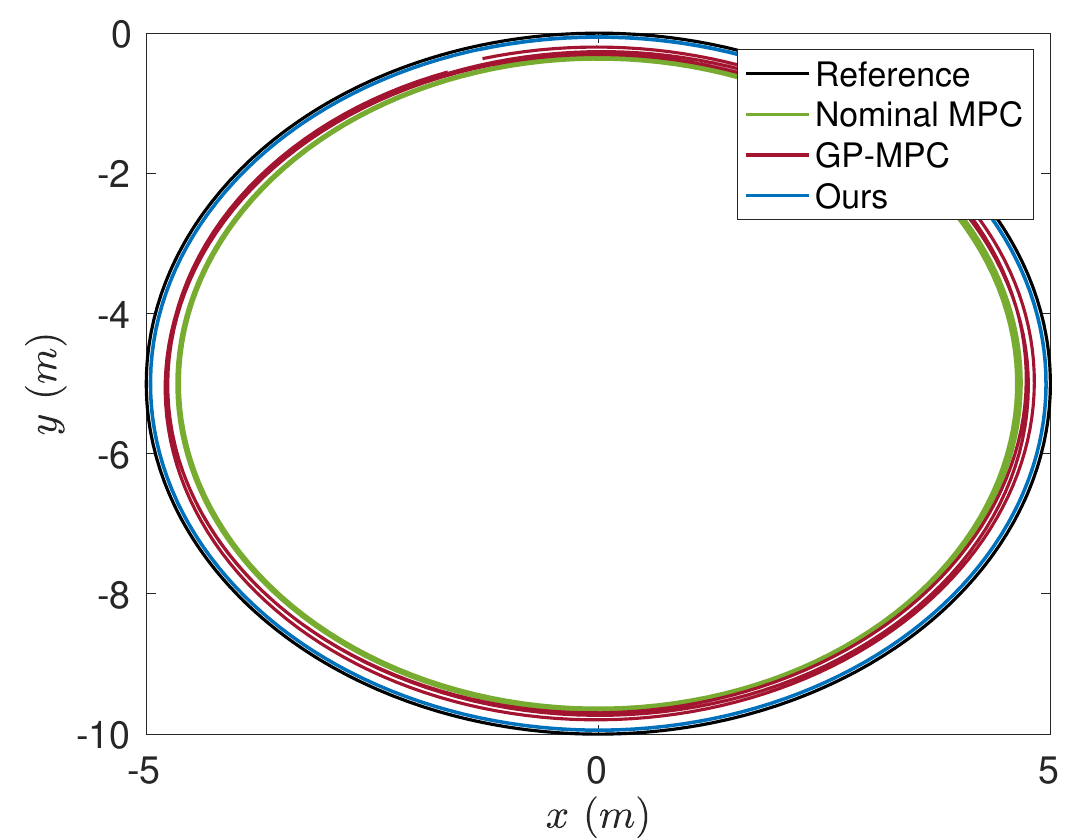}\label{fig_quad_traj_circle}}\quad\quad\quad\quad
    \subfigure[Sample trajectories in tracking a Lemniscate trajectory with $v_m=14.11 \ m/s$.]{\includegraphics[width=0.4\textwidth]{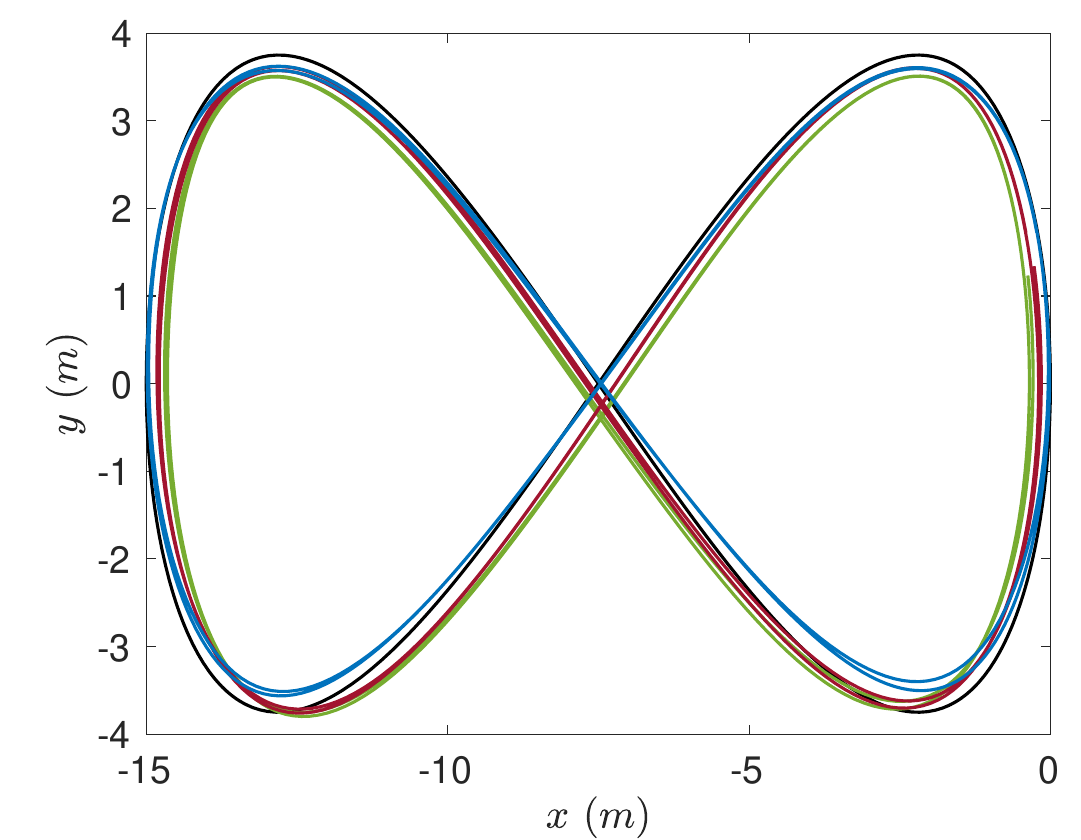}\label{fig_quad_traj_lemniscate}}	
    \caption{\textbf{Tracking Performance Comparison for the Quadrotor Experiments of Circle and Lemniscate Trajectories in \Cref{subsec:sim-3}.} The plots present the trajectories of \NMPC, \GPMPC, and \Cref{alg:MPC} tracking a Circle trajectory with $v_m=10.07 \ m/s$, and a Lemniscate trajectory with $v_m=14.11 \ m/s$, from $t=40s$ to $t=50s$ when the quadrotor reaches the maximal speeds. The trajectory of \Cref{alg:MPC} with \INDI is omitted as it overlaps with \Cref{alg:MPC} w/o \INDI.}
    \label{fig_quad_traj}
    \vspace{3mm}
\end{figure}

\myParagraph{Simulation Setup} The quadrotor dynamics, the simulation environment, and the controller setup are as follows:
\subsubsection{Quadrotor Dynamics} 
\begin{align}
    \dot{\boldsymbol{p}} &= \boldsymbol{v}, \quad & m \dot{\boldsymbol{v}} &= m \boldsymbol{g} + \boldsymbol{f} + \boldsymbol{f}_{a},  \label{eq:uav_trans} \\
    \dot{\boldsymbol{q}} &=\frac{1}{2}\boldsymbol{q}\otimes\left[\begin{array}{c}
        0 \\
        \boldsymbol{\omega}
    \end{array}\right], \quad & \mathcal{J} \dot{\boldsymbol{\omega}} &= -\boldsymbol{\omega} \times \mathcal{J} \boldsymbol{\omega} + \boldsymbol{\tau}, \label{eq:uav_attitude}
\end{align}
where $\boldsymbol{p} \in \mathbb{R}^{3}$ and $\boldsymbol{v} \in \mathbb{R}^{3}$ are position and velocity in the inertial frame, $\boldsymbol{q}$ is the quaternion, $\otimes$ is the quaternion multiplication operator, $\boldsymbol{\omega} \in \mathbb{R}^{3}$ is the body angular velocity, $m$ is the quadrotor mass, $\mathcal{J}$ is the inertia matrix of the quadrotor, $\boldsymbol{g}$ is the gravity vector, $\boldsymbol{f} =  \boldsymbol{R}\left[0\ 0\ T\right]^\top\in \mathbb{R}^3$ and $\boldsymbol{\tau} \in \mathbb{R}^3$ are the total thrust and body torques from the four rotors, $T$ is the thrust from the four rotors along the $z-$axis of the body frame, and $\boldsymbol{f}_{a} \in \mathbb{R}^3$ is the aerodynamic force.

\subsubsection{Gazebo Environment and Control Architecture} 
We employ the AscTec Hummingbird quadrotor model using the RotorS simulator in Gazebo~\cite{RotorS2016}. 
{
The RotorS simulator implements the rotor drag as aerodynamic effects, which is a linear mapping with respect to the body frame velocity~\cite{martin2010true}.
}
We use the following control architecture: the \MPC is running at $50Hz$, takes as input the reference trajectory, and outputs the desired total thrust and desired body angular velocity to the low-level body rate controller~\cite{faessler2016thrust,faessler2017differential};
the body rate controller is running at $200Hz$, then converts them into motor speed commands to control the quadrotor in the simulator;
the state of the quadrotor is available at $100Hz$.

\subsubsection{Control Design} 
The \MPC uses look-ahead horizon $N=10$ simulating the quadrotor dynamics for $1s$. We use quadratic cost functions with $Q = \diag{[0.5\textbf{I}_{3},\; 0.1\textbf{I}_{4},0.05\textbf{I}_{3},\; 0.01\textbf{I}_{3}]}$ and $R=\textbf{I}_{4}$. We use the RK4 method~\cite{kloeden1992stochastic} for discretization. 
We use as the feature $z_t$ the velocity, quaternion, body angular velocity, and normalized thrust from four rotors.
We sample $w_i$ from a Gaussian distribution with standard deviation $0.01$.
We use $M=50$ random Fourier features and $\eta=0.25$, and initialize $\hat{\alpha}$ as a zero vector. 
We use CasADi~\cite{andersson2019casadi} and acados~\cite{verschueren2022acados} to solve \cref{eq:mpc_ada_def}.

\subsubsection{Benchmark Experiment Setup}
We consider that the quadrotor is tasked to track a prescribed trajectory at different maximal speeds $v_{m}$, affecting the aerodynamic forces; we consider four types of reference trajectories: Circle, Wrapped Circle, Lemniscate, and Wrapped Lemniscate, showed in \Cref{fig_quad_ref_traj}.
We simulate each reference trajectory and each maximal speed 5 times.
We use as the performance metric the root mean squared error (RMSE) in position.

\myParagraph{Compared Algorithms} 
We compare \Cref{alg:MPC} with: a nominal \MPC that assumes no aerodynamic forces (\NMPC), and the Gaussian process \MPC (\GPMPC)~\cite{torrente2021data}. 
The \GP model in~\cite{torrente2021data} is pre-trained. We adopt the default training procedure per~\cite{torrente2021data}'s open-sourced code: a \NMPC is given $10$ random trajectories with maximal speeds $v_m \in [6.95,\; 15.62]$; then the \NMPC commands the quadrotor to track these trajectories and collects the training dataset with $3556$ data points; finally, the \GP model is trained such that $\boldsymbol{f}_{a}$ is predicted based on body velocities.
The prediction of the \GP model is used only for the first step over the look-ahead horizon.
We also combine \Cref{alg:MPC} with incremental nonlinear dynamic inversion (\INDI)~\cite{tal2020accurate} to account for unknown \mbox{disturbance in the attitude dynamics in \cref{eq:uav_attitude}.}

\myParagraph{Results} 
The results are given in \Cref{fig_quad_track} and \Cref{fig_quad_traj}. 
In \Cref{fig_quad_track}, \Cref{alg:MPC} demonstrates improved performance over \NMPC and \GPMPC in terms of the tracking error over all tested reference trajectories and maximal speeds. 
The limitation of \GPMPC appears to be: (i) due to computational complexity, the prediction of \GP is only used for one step in \MPC; by contrast, \Cref{alg:MPC} incorporates $\hat{h}\left(\cdot\right)$ over the entire look-ahead horizon $N$; (ii) \GPMPC may not perform well if the trajectories in the training dataset are different from the executed ones; in contrast, \Cref{alg:MPC} aims to collect data and learn $\hat{h}\left(\cdot\right)$ online, thus alleviates such generalization errors.
Additionally, the performance of \Cref{alg:MPC} is further improved with \INDI. However, due to the low frequency in \INDI control loop ($200Hz$) and state estimate ($100Hz$), the improvement is marginal especially in high-speed scenarios.\footnote{To achieve accurate tracking performance, \scenariof{{INDI}} requires high-frequency control update~(\eg $500Hz$ in \cite{tal2020accurate}, $300Hz$ in \cite{sun2022comparative}) and measurements (\eg $500Hz$ motor speed and IMU measurements in \cite{sun2022comparative}) to accurately estimate the external disturbances.}
\Cref{fig_quad_traj} presents the trajectories of \NMPC, \GPMPC, and \Cref{alg:MPC} tracking a Circle trajectory with $v_m=10.07 \ m/s$, and a Lemniscate trajectory with $v_m=14.11 \ m/s$, from $t=40s$ to $t=50s$ when the quadrotor reaches the maximal speeds.
In Circle trajectory, \Cref{alg:MPC} clearly achieves the best tracking performance. In Lemniscate trajectory, \Cref{alg:MPC} has good tracking performance except in the corners where the quadrotor needs to turn; while \NMPC and \GPMPC have tracking errors along the whole trajectory.



\section{Hardware Experiments}\label{sec:exp-hardware}
\begin{figure}[t]
    \centering
    \includegraphics[width=0.32\textwidth]{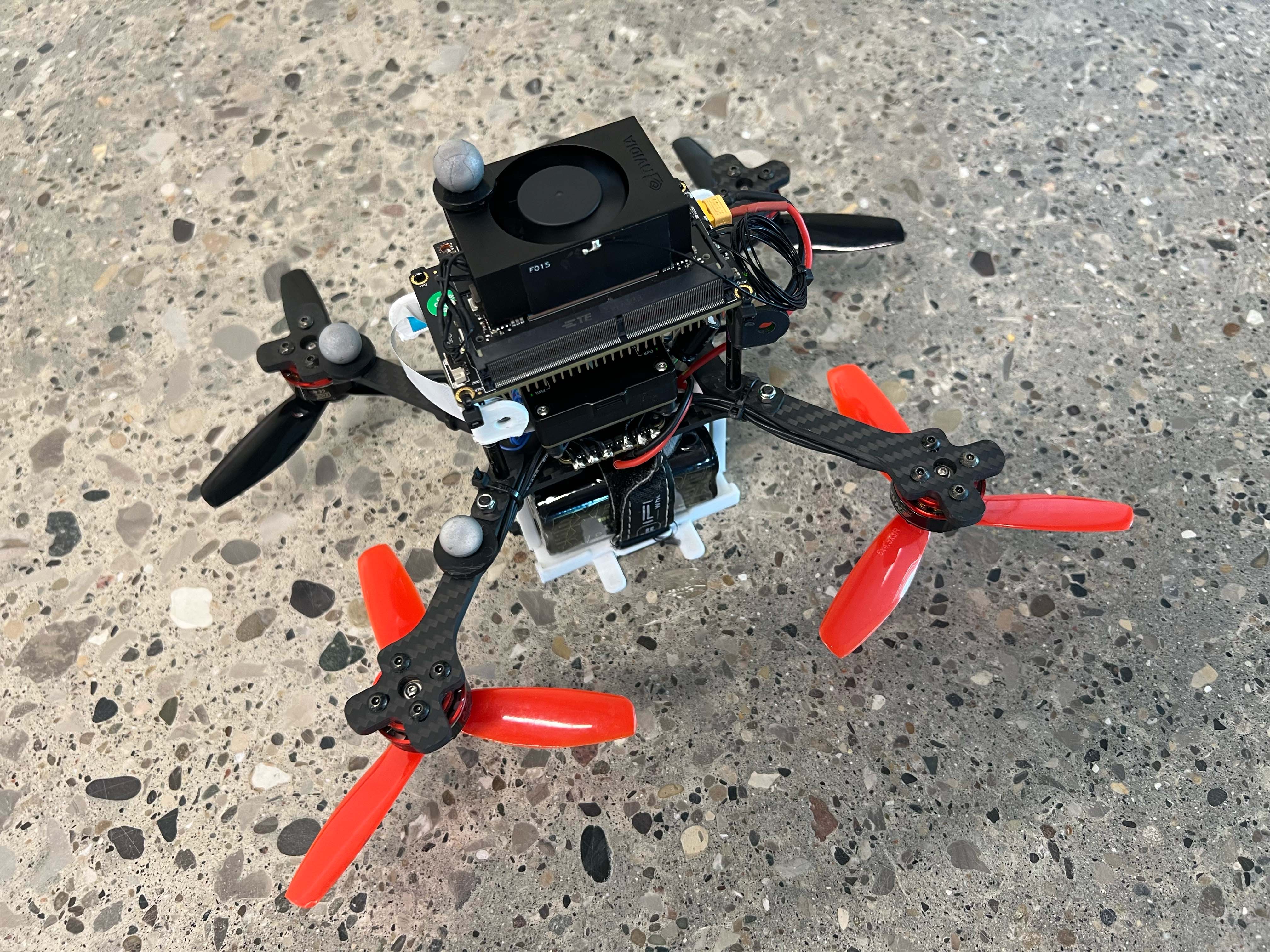}
    \caption{The quadrotor for the hardware experiments.}
    \vspace{3mm}
    \label{fig_hardware_quad}
\end{figure}

We evaluate \Cref{alg:MPC} in extensive real-world scenarios of control under uncertainty. 
For the experiments, we use a quadrotor as shown in \Cref{fig_hardware_quad} to track a circular trajectory despite ground effects and wind disturbances~(\Cref{fig_hardware}).
Additional results under aerodynamic drag effects and turbulent effects are given in \Cref{app:hardware}.

\begin{figure*}[t]
    \centering
    \subfigure[Average RMSE under Ground Effects.]{\includegraphics[width=0.3\textwidth]{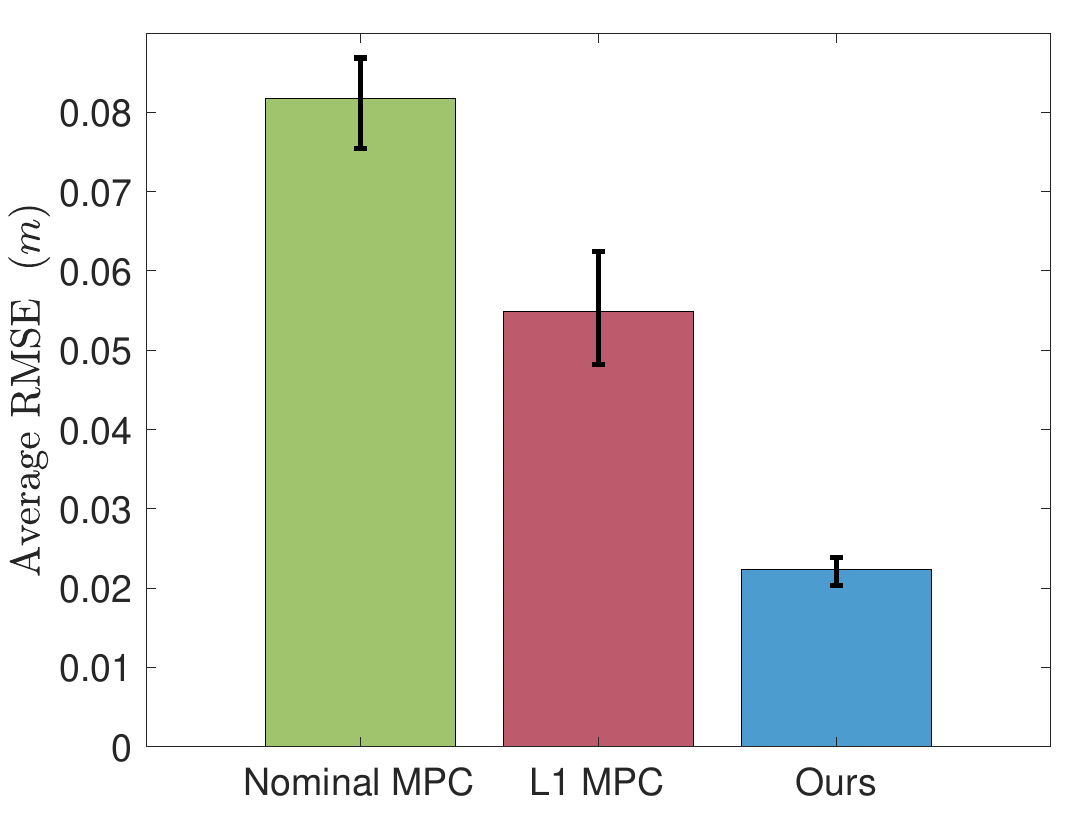}\label{fig_hardware_ground_error}}	
    \subfigure[Average RMSE under Wind Disturbances.]{\includegraphics[width=0.3\textwidth]{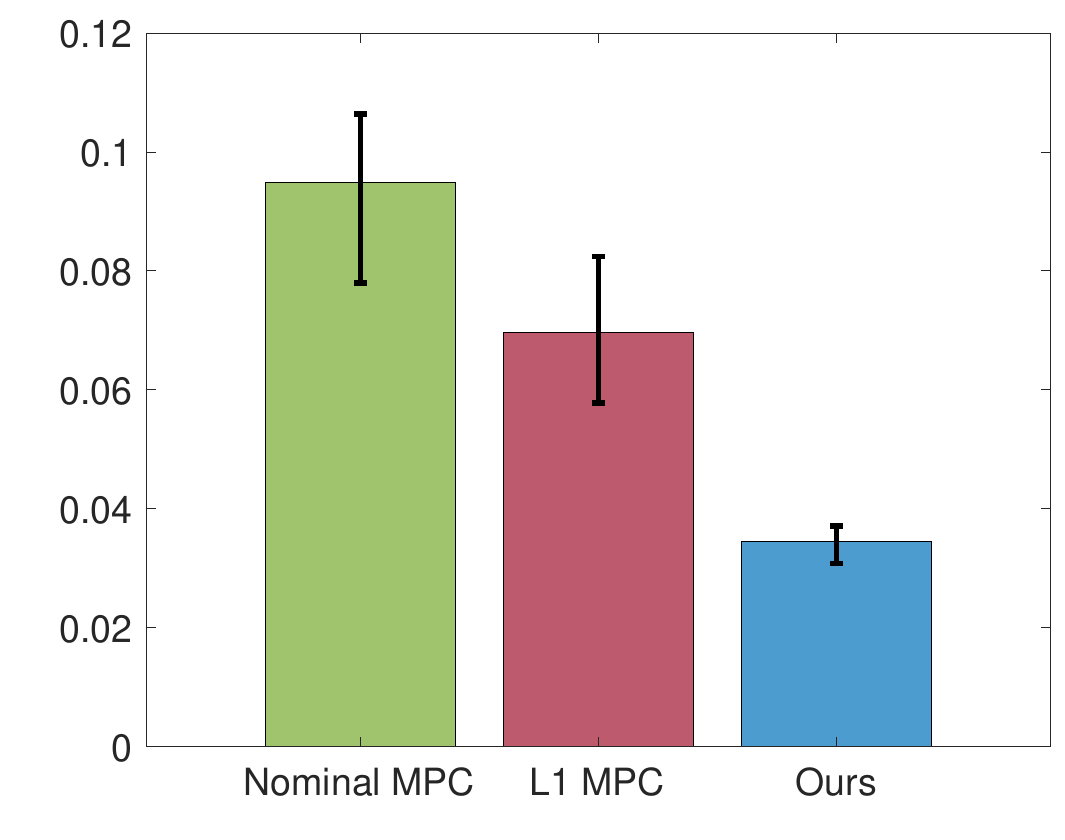}\label{fig_hardware_wind_error}}	
    \subfigure[Average RMSE under Ground Effects \& Wind Disturbances. \scenariof{Nominal MPC} is not presented due to crash.]{\includegraphics[width=0.3\textwidth]{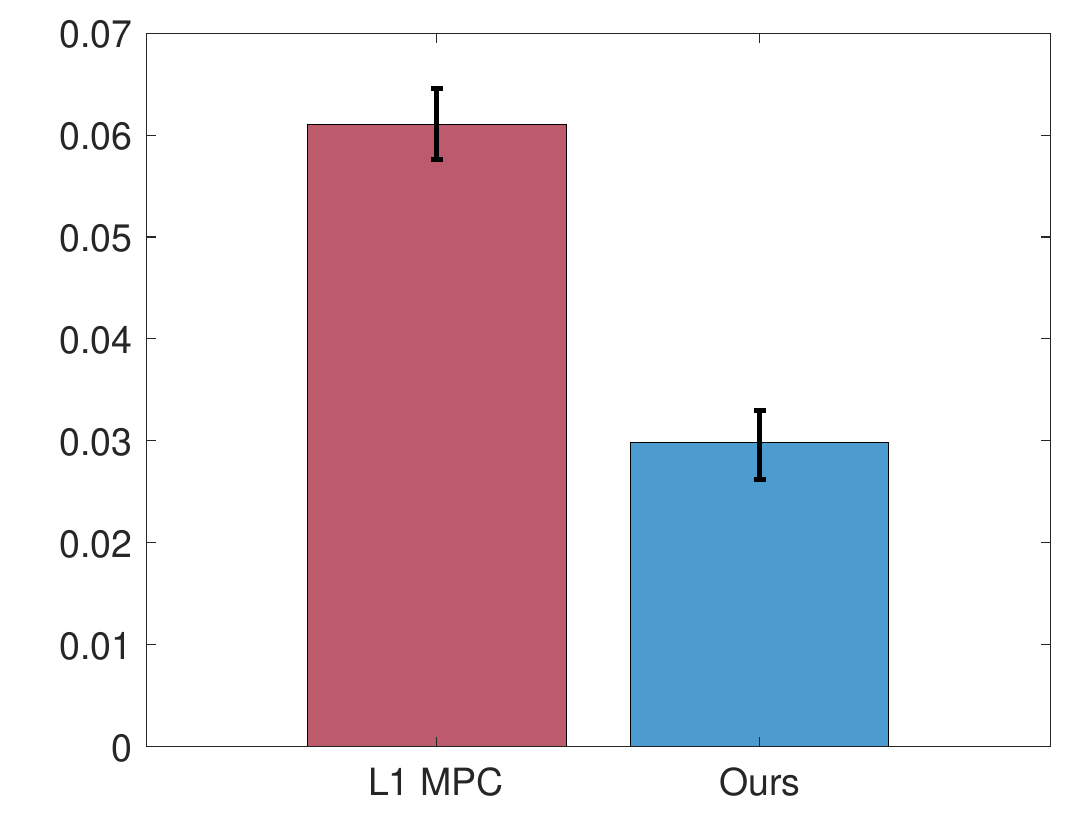}\label{fig_hardware_wind_ground_error}}	
    \caption{\textbf{Tracking Performance Comparison for the Hardware Quadrotor Experiments in \Cref{sec:exp-hardware}.} The error bar represents the minimum and maximum RMSE. \Cref{alg:MPC} (ours) demonstrates improved performance compared to \NMPC and \LMPC in terms of tracking error over all tested scenarios.}
    \label{fig_hardware_rmse}
\end{figure*}

\begin{figure*}[t]
    \centering
    \includegraphics[width=\textwidth]{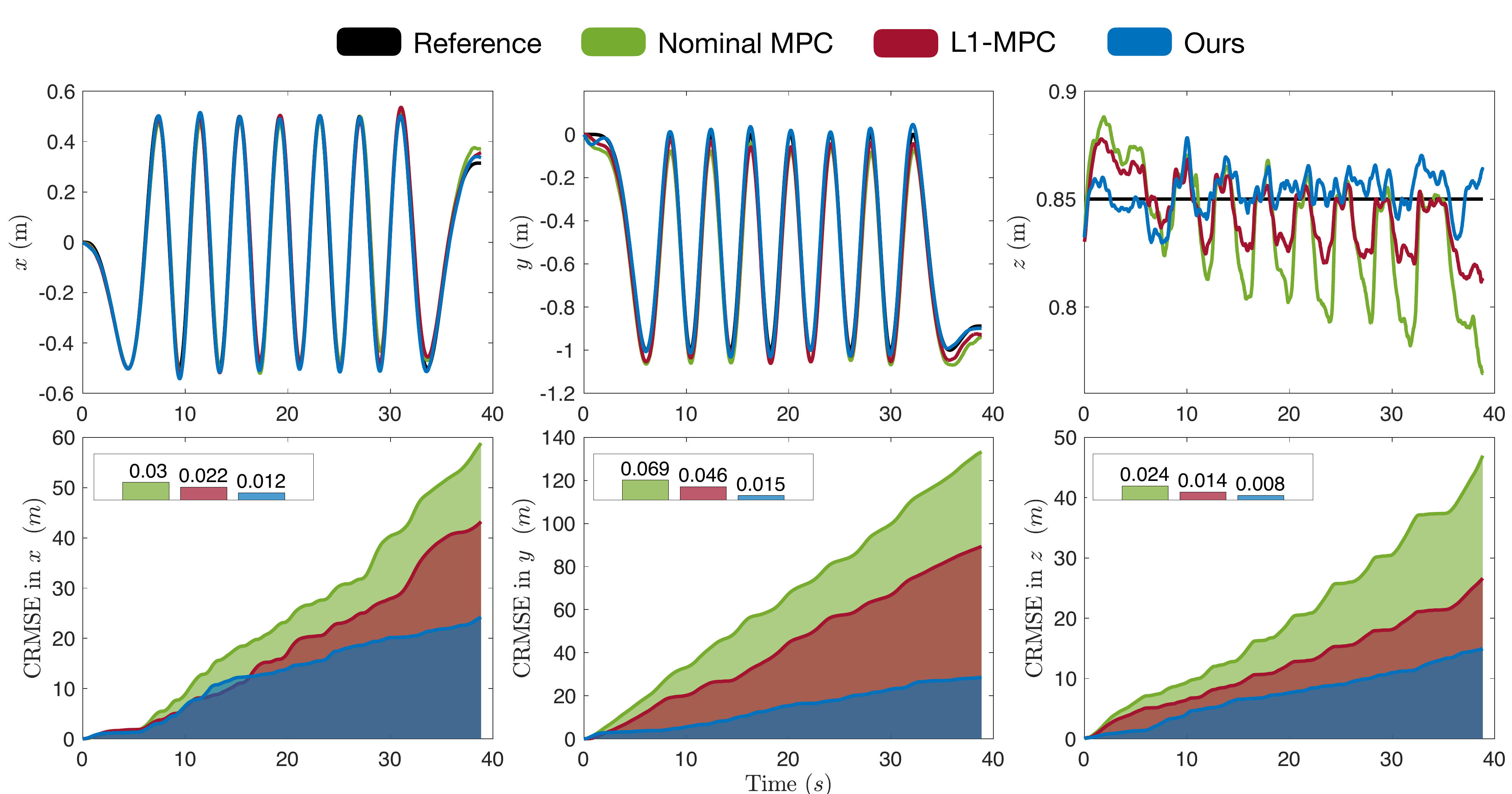}
    \caption{\textbf{Tracking Performance Comparison for the Hardware Quadrotor Experiments Under Ground Effects in \Cref{sec:exp-hardware}.} The plots present sample trajectories of \NMPC, \LMPC, and \Cref{alg:MPC} in $x$-, $y$-, and $z$- position~(top) and the corresponding the cumulative RMSE~(CRMSE)~(bottom). The inserted plots present the average RMSE~($m$).}
    \label{fig_hardware_traj_ground}
\end{figure*} 

\begin{figure*}[]
    \centering
    \includegraphics[width=\textwidth]{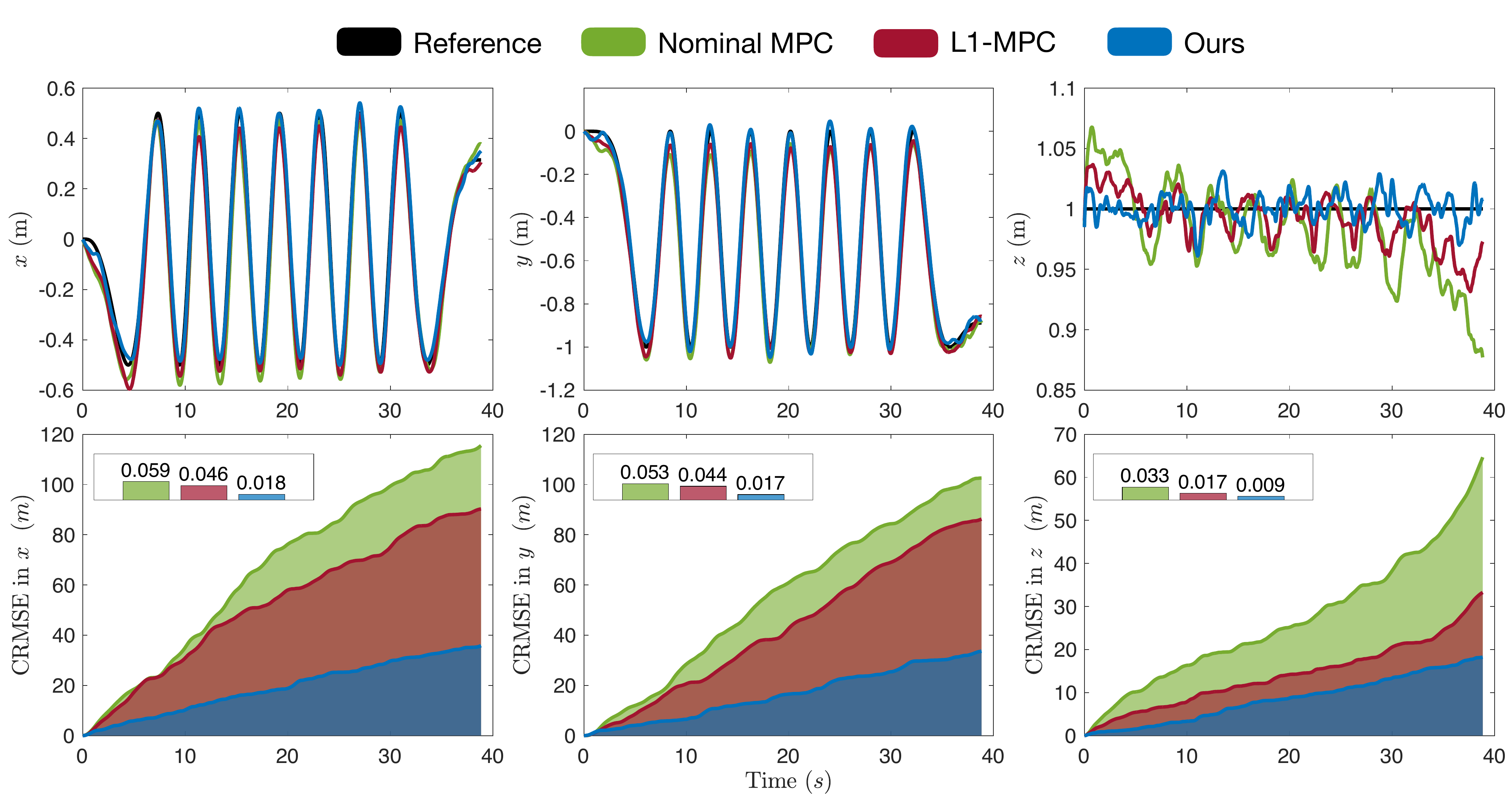}
    \caption{\textbf{Tracking Performance Comparison for the Hardware Quadrotor Experiments Under Wind Disturbances in \Cref{sec:exp-hardware}.} The plots present the sample trajectories of \NMPC, \LMPC, and \Cref{alg:MPC} in $x$-, $y$-, and $z$- position~(top) and the corresponding CRMSE~(bottom). The inserted plots present the average RMSE~($m$).}
    \label{fig_hardware_traj_wind}
\end{figure*} 

\begin{figure*}[]
    \centering
    \includegraphics[width=\textwidth]{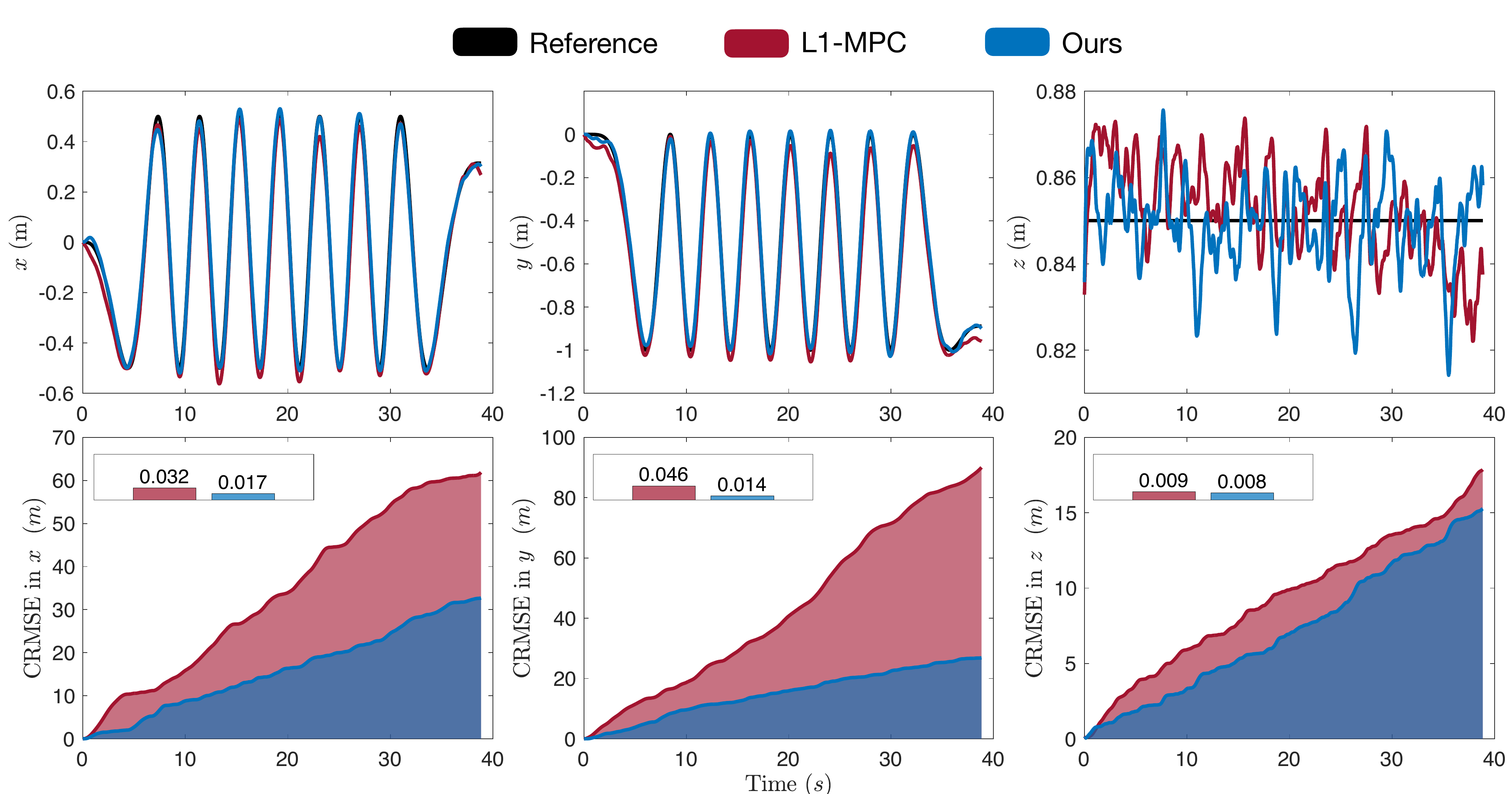}
    \caption{\textbf{Tracking Performance Comparison for the Hardware Quadrotor Experiments under both Ground Effects and Wind Disturbances in \Cref{sec:exp-hardware}.} The plots present sample trajectories of \LMPC and \Cref{alg:MPC} in $x$-, $y$-, and $z$- position~(top) and the corresponding CRMSE~(bottom). The inserted plots present the average RMSE~($m$). \NMPC is not presented since it causes crashes under both ground effects and wind disturbances.}
    \label{fig_hardware_traj_wind_ground}
\end{figure*}

\myParagraph{Quadrotor}
The quadrotor dynamics are modeled as in \cref{eq:uav_trans,eq:uav_attitude} with weight $0.68kg$ and $0.22m$ diagonal motor-to-motor distance.  
The quadrotor is equipped with an Nvidia Jetson Orin NX 16GB~\cite{youvan2024developing} to handle on-board computation.
We use the low-level controller in Holybro Pix32 v6 running PX4-Autopilot firmware~\cite{PX4}.
The communication between Jetson Orin NX and PX4 is through MAVROS~\cite{mavros}.
A Vicon motion capture system provides the pose of the quadrotor at $30Hz$, and we use PX4's Extended Kalman filter (EKF) to fuse the Vicon measurements with IMU readings onboard to estimate the quadrotor's odometry. The odometry is sent to the controller at $100Hz$ and \MPC runs at $50Hz$.

\myParagraph{Control Design} 
We use the same \MPC parameters and feature $z_t$ as in \Cref{subsec:sim-3}. 
We use $M=25$ random Fourier features and $\eta=0.05$, and initialize $\hat{\alpha}$ as a zero vector. 
Those parameters are fixed for all hardware experiments.
We use CasADi~\cite{andersson2019casadi} and acados~\cite{verschueren2022acados} to solve \cref{eq:mpc_ada_def}.

\myParagraph{Compared Algorithms} 
We compare \Cref{alg:MPC} with a nominal \MPC that assumes no aerodynamic forces (\NMPC), and the L1-adaptive \MPC  (\LMPC)~\cite{wu2023mathcal}. 
The \LMPC uses the L1 adaptive control to estimate the aerodynamic force and moment in \cref{eq:uav_trans,eq:uav_attitude}. Then, the \MPC calculates the control input based on the estimated aerodynamic force and moment.  We adopt the default parameters of estimation using L1 adaptive control per~\cite{wu2023mathcal}'s open-sourced code.

\myParagraph{Benchmark Experiment Setup} 
We consider that the quadrotor is tasked to track a circular trajectory with diameter $1m$ at maximal speed $v_{m} = 0.8~m/s$.
We consider the following disturbance scenarios:
\begin{itemize}
    \item \textit{Ground Effects:} We use an elevated foam mat to create the ground effects to the quadrotor~(\Cref{fig_hardware}, left). The surface of the foam mat is $0.78m$ high and the circular trajectory is set to be $0.85m$ in height. Half of the circular trajectory is above the foam mat and the other half is outside. This setting creates significant ground effects when the quadrotor is above the foam mat and a sharp transition between w/ and w/o the ground effects.
    \item \textit{Wind Disturbances:} We use a fan to create wind disturbaces~(\Cref{fig_hardware}, center). The circular trajectory is at $1m$ height and its center is $3m$ away from the fan. The quadrotor experiences wind speed from $2.0~m/s$ to $5.4~m/s$ along the circular trajectory, depending on its distance to the fan. The trajectory ends at a position with wind speed from $3.5~m/s$ to $5.0~m/s$.
    \item \textit{Ground Effects + Wind Disturbances:} We combine the previous two scenarios and now the quadrotor suffers from both ground effects and wind disturbances~(\Cref{fig_hardware}, right). The circular trajectory is at $0.85m$ and the wind speed varies from $1.5~m/s$ to $5.0~m/s$ along the circular trajectory. The trajectory ends at a position with wind speed from $3.3~m/s$ to $4.7~m/s$.
\end{itemize}

Additional unknown disturbances come from the (i) EKF estimation errors, (iii) battery's voltage drop during the flight, which causes a decrease in motors' thrust, (iii) communication latency between Jetson Orin NX and PX4, and (iv) modeling error of quadrotor dynamics. The resulting unknown $h$ is a combination of multi-source heterogeneous disturbances.
We conduct each scenario 5 times. We use as the performance metric the RMSE in position.

\myParagraph{Results} 
The results are given in \Cref{fig_hardware_rmse}, \Cref{fig_hardware_traj_ground}, \Cref{fig_hardware_traj_wind}, and \Cref{fig_hardware_traj_wind_ground}. 
\Cref{fig_hardware_rmse} shows that our algorithm achieves the best performance in average RMSE over all tested scenarios. 
\NMPC is the worst as it does not account for unknown disturbances, and it has increased $z$-direction tracking error during the flight due to insufficient thrust caused by voltage drop~(\Cref{fig_hardware_traj_ground} and \Cref{fig_hardware_traj_wind}, top right). \NMPC fails under both ground effects and wind disturbances~(\Cref{fig_hardware_traj_wind_ground}) due to strong wind and voltage drop.
\LMPC has improved performance over \NMPC by compensating for the estimated disturbances. However, it does not account for the future evolution of the unknown disturbances and has large tracking error when the unknown disturbances change abruptly, \eg a sharp transition between the ground and free space~(\Cref{fig_hardware_traj_ground}, top right).
In contrast, \Cref{alg:MPC} learns the model of unknown disturbances for predictive control, and therefore, achieves the best tracking performance even when the unknown disturbances change abruptly. \Cref{alg:MPC} achieves up to $75\%$ and $60\%$ improvement in average RMSE over \NMPC and \LMPC, respectively. 
In addition, \Cref{alg:MPC} using the same set of parameters can learn different unknown disturbances, \ie ground effects~(\Cref{fig_hardware_traj_ground}), wind disturbances~(\Cref{fig_hardware_traj_wind}), a combination of ground effects and wind disturbances~(\Cref{fig_hardware_traj_wind_ground}), and voltage drop. This is enabled by the self-supervised learning framework using the data collected online.

\section{Related Work}\label{sec:lit_review}
 
We discuss work on system identification and control across the research topics of adaptive control; robust control; control based on offline learning; control based on offline learning with online adaptation; and control based on online learning.\footnote{The related work also includes papers on \textit{sequential online model learning and predictive control}~\cite{lale2021model,muthirayan2022online}. These works assume no knowledge of system dynamics and instead learn the whole system dynamics online. Therefore, their methods first need to use random control inputs to excite the systems and estimate the unknown system dynamics; then they design control policy using the estimated system dynamics. In contrast, in this paper, we consider only partially unknown dynamics, and we simultaneously estimate unknown dynamics/disturbances and control the system. Finally,~\cite{lale2021model,muthirayan2022online} require persistent excitation to obtain regret guarantees, while we do not.} 

\paragraph*{Adaptive control}
Adaptive control methods often assume parametric uncertainty additive to the known system dynamics, \eg parametric uncertainty in the form of unknown coefficients multiplying known basis functions~\cite{slotine1991applied,krstic1995nonlinear,ioannou1996robust}. These coefficients are updated online and generate an adaptive control input to compensate for the estimated disturbances. 
These methods often require persistent excitation to guarantee the exponential stability of the closed-loop system~\cite{slotine1991applied,krstic1995nonlinear,ioannou1996robust}. 
In contrast, inspired by~\cite{boffi2022nonparametric}, our method handles unknown dynamics and disturbances that can be expressed in reproductive kernel Hilbert space (\RKHS), and does not require persistent excitation to enjoy near-optimality guarantees.
\cite{tal2020accurate,wu2023mathcal,das2024robust} bypass the assumption of parametric uncertainty and persistent excitation, and directly estimate the value of unknown disturbances. However, the methods therein, as well as, the relevant method in~\cite{boffi2021regret,boffi2022nonparametric,jia2023evolver}, focus on adaptive control to compensate for the estimated disturbances, instead of learning a model of the disturbances to enable model predictive control.

\paragraph*{Robust control}
Robust control algorithms select control inputs upon simulating the future system dynamics across a lookahead horizon~\cite{mayne2005robust,goel2020regret,sabag2021regret,martin2022safe,didier2022system,zhou2023safe,martin2024guarantees}. To this end, they assume a worst-case realization of disturbances, given an upper bound to the magnitude of the disturbances~\cite{mayne2005robust,goel2020regret,sabag2021regret,martin2022safe,didier2022system,zhou2023safe,martin2024guarantees}.
However, assuming the worst-case disturbances can be conservative. In addition, \cite{goel2020regret,sabag2021regret,martin2022safe,didier2022system,zhou2023safe,martin2024guarantees} focus on linear dynamical systems, and \cite{martin2022safe,didier2022system,zhou2023safe,martin2024guarantees} may be computationally expensive.
Hence, the application of these approaches is often limited to low-dimensional linear systems.

\paragraph*{Offline learning for control}
This line of work trains neural-network, Gaussian-process, or Koopman models using training data collected offline~\cite{sanchez2018real,carron2019data,torrente2021data,hewing2019cautious,tobin2017domain,ramos2019bayessim,lee2020learning,koopman1931hamiltonian,brunton2016sparse,abraham2017model,bruder2020data}. 
\cite{tobin2017domain,ramos2019bayessim,lee2020learning} train the models with data collected from random environments such that the controller can exhibit robustness to unseen environments~\cite{mohri2018foundations,abu2012learning},
In this paper, instead, we require no offline data collection and training, employ one-shot online learning in a self-supervised manner based on data collected online.

\paragraph*{Offline learning and online adaption for control}
Such control methods have two components: offline learning and online adaptation~\cite{finn2017model,williams2017information,nagabandi2018learning,belkhale2021model,shi2019neural,o2022neural,saviolo2023active}. The offline learning aims to train a neural network model using data collected from different environments, \eg the disturbances of a quadrotor under different wind speeds, so that the learned neural network model captures the underlying features across different environments. The online adaptation aims to update online either all the parameters of the neural network or only the parameters of the last layer to better predict unknown disturbances when encountering new environments. Similar to data-driven control with only offline learning, the methods need big data for effective training.
Also, updating all parameters online can be computationally heavy for real-time control~\cite{finn2017model,williams2017information,nagabandi2018learning,belkhale2021model}.
\cite{shi2019neural,o2022neural,saviolo2023active} only update the last layer parameters, but either directly compensate the estimated disturbances without predicting their future evolution to optimize the controller~\cite{shi2019neural,o2022neural}
or it is computationally expensive to use a neural network in on-board model predictive control~\cite{saviolo2023active}. 
In contrast, we propose to use random Fourier features to approximate online a model the unknown dynamics and disturbances.  Our approach allows us to maintain the computational efficiency of classical finite-dimensional parametric approximations that can be used in model predictive control while retaining the expressiveness of the \RKHS with high probability.

\paragraph*{Online learning for control} 
We discuss two lines of online learning algorithms: non-stochastic control algorithms~\cite{hazan2022introduction,agarwal2019online,zhao2022non,gradu2023adaptive,zhou2023efficient,zhou2023safecdc,zhou2023saferal} and contextual optimization-based control algorithms~\cite{krause2011contextual,berkenkamp2016safe,chowdhury2017kernelized,djeumou2022fly,vinod2022fly,valko2013finite}. 
These methods often quantify the performance of the algorithms through \textit{regret}, \ie the suboptimality against an optimal clairvoyant controller that knows the unknown disturbances and dynamics.
Non-stochastic control algorithms select control inputs based on past information only since they assume no model that can be used to simulate the future evolution of the disturbances~\cite{hazan2022introduction,agarwal2019online,zhao2022non,gradu2023adaptive,zhou2023efficient,zhou2023safecdc,zhou2023saferal}. 
Instead, they provide controllers with bounded regret guarantees, upon employing the \OCO framework to capture the non-stochastic control problem as a sequential game between a controller and an adversary~\cite{hazan2016introduction}. They rely on the knowledge of a known upper bound to the magnitude of the disturbances, and a pre-stabilizing controller.
The proposed approaches have been observed to be sensitive to the choice of the pre-stabilizing controller and the tuning parameters of \OGD, \eg in \cite{agarwal2019online,gradu2023adaptive,zhou2023safecdc}. 
Contextual optimization-based control algorithms ignore the system identification step, and optimize control by minimizing a one-step surrogate cost function, also providing regret guarantees~\cite{krause2011contextual,berkenkamp2016safe,chowdhury2017kernelized,djeumou2022fly,vinod2022fly,valko2013finite}. Specifically, such surrogate function is model based on past information, and depends on the current control action to be optimized and the known current state. Then the current control is selected by minimizing this one-step surrogate cost function due to the lack of a predictive model.
In contrast, our method uses \OGD to learn the model of the disturbances instead of optimizing the online controller, and uses model predictive control to optimize online the control input based on the learned online disturbance model.\footnote{In \cite{tsiamis2024predictive}, a predictive control method with online learning is provided. It focuses on learning unknown trajectories of a target, where the target trajectory does not adapt to the pursuer's motion. In contrast, we focus on learning unknown dynamics/disturbances that are adaptive to the controller's state and control input. In addition, \cite{tsiamis2024predictive} focuses on linear systems and linear \MPCf, while we utilize nonlinear \MPCf for nonlinear control-affine systems. }
In addition, we consider a stronger definition of regret than \cite{agarwal2019online,zhao2022non,gradu2023adaptive,zhou2023efficient,zhou2023safecdc,zhou2023saferal,djeumou2022fly,vinod2022fly} by (i) allowing the unknown dynamics and disturbances adaptive to the state and control, instead of the same disturbances for the designed controller and the optimal controller~\cite{agarwal2019online,zhao2022non,gradu2023adaptive,zhou2023efficient,zhou2023safecdc,zhou2023saferal},  (ii) allowing the optimal controller to select control input with multi-step prediction into the future, instead of only one-step prediction~\cite{djeumou2022fly,vinod2022fly}.

\section{Conclusion} \label{sec:con}

We provided \Cref{alg:MPC} for the problem of \textit{Simultaneous System Identification and Model
Predictive Control} (\Cref{prob:control}). 
\Cref{alg:MPC} guarantees no dynamic regret against an optimal clairvoyant (non-causal) policy that knows the disturbance function $h$ a priori~(\Cref{theorem:regret_OLMPC}). 
The algorithm uses random Fourier features to approximate the unknown dynamics or disturbances in reproducing kernel Hilbert spaces. Then, it employs model predictive control based on the current learned model of the unknown dynamics (or disturbances).
The model of the unknown dynamics is updated online in a self-supervised manner using least squares based on the data collected while controlling the system.

We validate \Cref{alg:MPC} in simulated and hardware experiments.  Our physics-based simulations include (i) a cart-pole aiming to maintain the pole upright despite inaccurate model parameters~(\Cref{subsec:sim-1,subsec:sim-2}), 
and (ii) a quadrotor aiming to track reference trajectories despite unmodeled aerodynamic drag effects~(\Cref{subsec:sim-3}).
{
Our hardware experiments involve a quadrotor aiming to track a circular trajectory despite unmodeled aerodynamic drag effects, ground effects, and wind disturbances~(\Cref{sec:exp-hardware} and \Cref{app:hardware}).}
We observed that our method demonstrated better tracking performance and computational efficiency than the state-of-the-art methods \GPMPC~\cite{hewing2019cautious,torrente2021data}, \NSMPC~\cite{zhou2023saferal}, and \LMPC~\cite{wu2023mathcal}.


\section*{Acknowledgements}
We thank Torbjørn Cunis, Ilya Kolmanovsky, and Miguel Castroviejo-Fernandez from the University of Michigan, for their invaluable discussion on the Lipschitzness of value function in nonlinear \MPC. We thank Devansh Agrawal, Hardik Parwana, and Dimitra Panagou from the University of Michigan, for their invaluable help in hardware experiments.

\appendices
\vspace{-1mm}
\appendix

\subsection{Extension to Systems with Unmodeled Process Noise}\label{app:extension}
In this section, we provide the extension of \Cref{alg:MPC} to systems corrupted with also unmodeled (potentially non-stochastic) noise that is additive to the system dynamics, \ie
\begin{equation}
	x_{t+1} = f\left(x_{t}\right) + g\left(x_{t}\right) u_{t} + h\left(z_{t}\right) + w_{t}, 
    \label{eq:noisy_affine_sys}
\end{equation}
where $w_{t}$ represents the unmodeled process noise.


Then, \Cref{alg:MPC} guarantees the following.

\begin{corollary}[Dynamic Regret in Systems with Process Noise]\label{corollary:regret_OLMPC_extension}
Assume \Cref{alg:MPC}'s learning rate is $\eta=\calO\left({1}/{\sqrt{T}}\right)$.  Then, for the system in \cref{eq:noisy_affine_sys}, \Cref{alg:MPC} achieves 
\begin{equation}
    \DReg \leq \calO\left(T^{\frac{3}{4}}\right) + L \sqrt{ T\sum_{t=1}^{T}\|w_{t}\|^2 }.
    \label{eq:corollary_regret_OLMPC_extension}
\end{equation}
\end{corollary}

Compared to \Cref{theorem:regret_OLMPC}, the regret bound in \Cref{corollary:regret_OLMPC_extension} has an additional term that depends on the energy of the noise sequence $(w_1, \dots, w_T)$, \ie $\sum_{t=1}^{T}\|w_{t}\|^2$. Specifically, when the energy is less than $\calO(T)$, we achieve sublinear regret.

\subsection{Proof of \Cref{theorem:regret_OLMPC}}\label{app:theorem_regret_OLMPC}
We use $x_{t_1:t_2}$ to denote the sequence $\left( x_{t_1}, \; \dots, \; x_{t_2} \right)$, and $\Phi\left(\cdot\right) \triangleq \frac{1}{M}[\Phi\left(\cdot, \theta_1 \right), \; \dots, \; \Phi\left(\cdot, \theta_M \right)]$.
We use $\hat{x}$ to denote the one-step-ahead prediction given an estimate $\hat{\alpha}$, \ie $\hat{x}_{t+1} = f\left(x_{k}\right) + g\left(x_{k}\right) u_{k} + \Phi\left(z_k\right) \hat{\alpha}_t=x_{t+1}+ \Phi\left(z_k\right) \hat{\alpha}-h(z_k)$.

Then, the dynamic regret in \Cref{def:DyReg_control} can be upper bounded as follows,
\begin{equation}
    \begin{aligned}
         \DReg &= \sum_{t=1}^{T} c_{t}\left(x_{t}, u_{t}\right)-\sum_{t=1}^{T} c_{t}\left(x_{t}^{\star}, u_{t}^{\star}\right) \\
         &\leq \sum_{t=1}^{T} V_{t}\left(x_{t}; \hat{\alpha}_{t}\right)-\sum_{t=1}^{T} c_{t}\left(x_{t}^{\star}, u_{t}^{\star}\right) \\
         &\leq \sum_{t=1}^{T} V_{t}\left(x_{t}; \hat{\alpha}_{t}\right),
    \end{aligned}
\end{equation}
where the first inequality holds by definition of $V_{t}\left(x_{t}; \hat{\alpha}_{t}\right)$, and the second inequality holds by $\sum_{t=1}^{T} c_{t}\left(x_{t}^{\star}, u_{t}^{\star}\right) \geq 0$. 

Hence, we aim to bound $\sum_{t=1}^{T} V_{t}\left(x_{t}; \hat{\alpha}_{t}\right)$. To this end, we first establish the relationship between $V_{t}\left(x_{t}; \hat{\alpha}_{t}\right)$ and $V_{t+1}\left(x_{t+1}; \hat{\alpha}_{t+1}\right)$ using the following lemmas (the proofs of them are given in \Cref{app:lemma_A} and \Cref{app:lemma_B}). 

\begin{lemma}[]\label{lemma:A}
    Suppose \Cref{assumption:stability}, \Cref{assumption:func_space}, and \Cref{assumption:feature_map}, hold, then for all $N > \left(\overline{\lambda}/\underline{\lambda}\right)^2 +1$, $V_{t+1}\left(\hat{x}_{t+1}; \hat{\alpha}_{t}\right)$ and $V_{t}\left(x_{t}; \hat{\alpha}_{t}\right)$ satisfy 
    \begin{equation}
         V_{t+1}\left(\hat{x}_{t+1}; \hat{\alpha}_{t}\right) \leq \epsilon V_{t}\left(x_{t}; \hat{\alpha}_{t}\right),
    \end{equation}
    where $\epsilon = 1 - \left(1 - \frac{\left(\overline{\lambda}/\underline{\lambda}\right)^2}{N-1} \right) \left({\underline{\lambda}}/{\overline{\lambda}} \right) \in (0,1)$.
\end{lemma}

\Cref{lemma:A} indicates that the system in \cref{eq:mpc_ada_dyn} is globally asymptotic stable \cite{grimm2005model} when the \MPC policy is applied: the value function keeps decreasing and converges to zero, thus, the state must converge to zero.

\begin{lemma}[]\label{lemma:B}
    Suppose \Cref{assumption:lipschitz}, \Cref{assumption:func_space}, \Cref{assumption:feature_map}, and \Cref{assump:small_approx_error}, hold, then $V_{t+1}\left({x}_{t+1}; \hat{\alpha}_{t+1}\right)$ and $V_{t+1}\left(\hat{x}_{t+1}; \hat{\alpha}_{t}\right)$ for \cref{eq:affine_sys} satisfy 
    \begin{equation}
        V_{t+1}\left({x}_{t+1}; \hat{\alpha}_{t+1}\right) - V_{t+1}\left(\hat{x}_{t+1}; \hat{\alpha}_{t}\right)  \leq L \left( \sqrt{ l_t\left(\hat{\alpha}_t\right) } + \| \eta \nabla_t \| \right).
    \end{equation}
\end{lemma}
\Cref{lemma:B} is achieved upon using \Cref{assumption:lipschitz} to upper bound $| V_{t+1}\left({x}_{t+1}; \hat{\alpha}_{t+1}\right) - V_{t+1}\left(\hat{x}_{t+1}; \hat{\alpha}_{t}\right) |$ with $\|{x}_{t+1}- \hat{x}_{t+1}\|$ and $\| \hat{\alpha}_{t}-  \hat{\alpha}_{t+1}\|$.

Using \Cref{lemma:A} and \Cref{lemma:B}, we have 
\begin{equation}
    V_{t+1}\left({x}_{t+1}; \hat{\alpha}_{t+1}\right) \leq \epsilon V_{t}\left(x_{t}; \hat{\alpha}_{t}\right) + L \left( \sqrt{ l_t\left(\hat{\alpha}_t\right) } + \| \eta \nabla_t \| \right).
    \label{eq:app_V_expansion}
\end{equation}

Then, we can bound $V_{t+1}\left({x}_{t+1}; \hat{\alpha}_{t+1}\right)$ by
\begin{equation}
    \begin{aligned}
        & \; V_{t+1}\left({x}_{t+1}; \hat{\alpha}_{t+1}\right)  \\
        \leq & \; \epsilon V_{t}\left(x_{t}; \hat{\alpha}_{t}\right) + L \left( \sqrt{ l_t\left(\hat{\alpha}_t\right) } + \| \eta \nabla_t \| \right) \\
        \leq & \; \epsilon^2 V_{t-1}\left(x_{t-1}; \hat{\alpha}_{t-1}\right) + \epsilon L \left( \sqrt{ l_{t-1}\left(\hat{\alpha}_{t-1}\right) }+ \| \eta \nabla_{t-1} \| \right) \\ 
        & \qquad  + L \left( \sqrt{ l_t\left(\hat{\alpha}_t\right) } + \| \eta \nabla_t \| \right) \\
        \leq & \; \epsilon^t V_{1}\left(x_{1}; \hat{\alpha}_{1}\right) + \sum_{k=1}^{t} \epsilon^{t-k} L \left( \sqrt{ l_k\left(\hat{\alpha}_k\right) } + \| \eta \nabla_k \| \right) \\
        \leq & \; \epsilon^t \overline{\lambda}\sigma\left(x_1\right) + \sum_{k=1}^{t} \epsilon^{t-k} L \left( \sqrt{ l_k\left(\hat{\alpha}_k\right) } + \| \eta \nabla_k \| \right),
    \end{aligned}
    \label{eq:app_mpc_stability_value}
\end{equation}
where the last inequality uses \Cref{assumption:stability}~(ii).

Therefore, we have 
\begin{equation}
    \begin{aligned}
        & \; \sum_{t=1}^{T} V_{t}\left(x_{t}; \hat{\alpha}_{t}\right) \\
        \leq & \; \sum_{t=1}^{T} \epsilon^{t-1} \overline{\lambda}\sigma\left(x_1\right)  + \sum_{t=1}^{T}\sum_{k=1}^{t-1} \epsilon^{t-1-k} L \left( \sqrt{ l_k\left(\hat{\alpha}_k\right) } + \| \eta \nabla_k \| \right) \\
        \leq & \; \sum_{t=1}^{T} \epsilon^{t-1} \overline{\lambda}\sigma\left(x_1\right)  +  \left(\sum_{t=0}^{T} \epsilon^{t} \right) \left(\sum_{t=1}^{T} L \left( \sqrt{ l_t\left(\hat{\alpha}_t\right) } + \| \eta \nabla_t \| \right)\right) \\
        \leq & \; \frac{\overline{\lambda}}{1-\epsilon}\sigma\left(x_1\right) + \frac{1}{1-\epsilon} \left(\sum_{t=1}^{T} L \left( \sqrt{ l_t\left(\hat{\alpha}_t\right) } + \| \eta \nabla_t \| \right)\right),
    \end{aligned}
\end{equation}
where the second inequality holds by adding positive terms to $\sum_{t=1}^{T}\sum_{k=1}^{t-1} \epsilon^{t-1-k} L \left( \sqrt{ l_k\left(\hat{\alpha}_k\right) } + \| \eta \nabla_k \| \right) $ to complete $\left(\sum_{t=0}^{T} \epsilon^{t} \right) \left(\sum_{t=1}^{T} L \left( \sqrt{ l_t\left(\hat{\alpha}_t\right) } + \| \eta \nabla_t \| \right)\right) $.

Since $x_1$ is bounded, we have 
\begin{equation}
    \frac{\overline{\lambda}}{1-\epsilon}\sigma\left(x_1\right) = \calO\left(1\right).
    \label{eq:app_O1}
\end{equation}

We now upper bound the term $\sum_{t=1}^{T} L \sqrt{ l_t\left(\hat{\alpha}_t\right) }$. By the Cauchy-Schwarz inequality, we have 
\begin{equation}
    \sum_{t=1}^{T} L \sqrt{ l_t\left(\hat{\alpha}_t\right) } \leq L \sqrt{T} \sqrt{ \sum_{t=1}^{T} l_t\left(\hat{\alpha}_t\right) }.
\end{equation}

From \Cref{theorem:OGD}, we have $\sum_{t=1}^{T} l_t\left(\hat{\alpha}_t\right)  \leq \calO\left(\sqrt{T}\right)$. Thereby, we obtain
\begin{equation}
    \sum_{t=1}^{T} L \sqrt{ l_t\left(\hat{\alpha}_t\right) } \leq \calO\left({T}^{\frac{3}{4}}\right).
    \label{eq:app_OLS_regret}
\end{equation}

Next, we consider the term $\sum_{t=1}^{T} L \| \eta \nabla_t \|$. Since $\Phi\left(\cdot\right)$, $\alpha$, $\hat{\alpha}_t$ are uniformly bounded for all $t$, then $\nabla_t$ is bounded by a constant $G$. Then
\begin{equation}
    \sum_{t=1}^{T} L \| \eta \nabla_t \|\, \leq \sum_{t=1}^{T} L G \| \eta \| \,\leq \calO\left(\sqrt{T}\right), 
    \label{eq:app_eta_regret}
\end{equation}
where the last inequality holds by $\eta = \calO\left(\frac{1}{\sqrt{T}}\right)$.

Combining \cref{eq:app_O1,eq:app_OLS_regret,eq:app_eta_regret} gives the result in \Cref{theorem:regret_OLMPC}. \qed

\subsection{Proof of \Cref{corollary:regret_OLMPC_extension}}\label{app:corollary_regret_OLMPC_extension}

For \cref{eq:noisy_affine_sys} where $w_t$ is the unmodeled process noise, we follow similar steps of the proof of \Cref{theorem:regret_OLMPC}, with the difference of \cref{eq:app_OLS_regret}:
\begin{equation}
    \sum_{t=1}^{T} L \sqrt{ l_t\left(\hat{\alpha}_t\right) } \leq \calO\left({T}^{\frac{3}{4}}\right) + L  \sqrt{ T\sum_{t=1}^{T}\|w_{t}\|^2 } ,
    \label{eq:app_OLS_regret_noise}
\end{equation}
due to $\sum_{t=1}^{T} l_t\left(\hat{\alpha}_t\right) - \sum_{t=1}^{T} l_t\left({\alpha}\right) = \sum_{t=1}^{T} l_t\left(\hat{\alpha}_t\right) - \sum_{t=1}^{T}\|w_{t}\|^2\,\leq \calO\left(\sqrt{T}\right)$ per \Cref{theorem:OGD}.

Hence, we obtain 
\begin{equation}
    \DReg \leq \calO\left(T^{\frac{3}{4}}\right) + L  \sqrt{ T\sum_{t=1}^{T}\|w_{t}\|^2 }.
\end{equation}
\qed



\subsection{Proof of \Cref{lemma:A}}\label{app:lemma_A}
Consider that at time step $t$, the system in \cref{eq:affine_sys} is at state $x_t$ and the \MPC problem becomes
\begin{equation}
    \label{eq:app_mpc_ada}
    \begin{aligned}
        & \underset{{u}_{t}, \ \ldots, \ {u}_{t+N-1}}{\operatorname{\textit{argmin}}} \sum_{k=t}^{t+N-1} c_{k}\left(x_{k},u_{k}\right)  \\
        & \ \ \operatorname{\textit{subject~to}} \;\quad x_{k+1} = f\left(x_{k}\right) + g\left(x_{k}\right) u_{k} + \Phi\left(z_{k}\right)\hat{\alpha}_t,\\
        & \qquad \qquad \qquad \ u_{k}\in \calU,  \;  k\in\{t,\ldots, t+N-1\}.
    \end{aligned}
\end{equation}
Let $\hat{u}_{0:N-1}^{t}$ and $\hat{x}_{t:t+N}^{t}$ be the optimal control input and state sequences to the above problem, where the superscript denotes the \MPC problem is solved at time step $t$.
We use $\hat{x}_{t+k}^{t} \triangleq \psi_t\left(k, x_t, \hat{u}_{0:N-1}^{t}; \hat{\alpha}_t\right)$ as the state reached from $x_t$ by applying $\hat{u}_{0:k-1}^{t}$ to the system dynamics with parameter $\hat{\alpha}_t$. By definition, $\psi_t\left(k, x_t, \hat{u}_{0:N-1}^{t}; \hat{\alpha}_t\right)$ satisfies
\begin{align}
     & \psi_t\left(k, x_t, \hat{u}_{0:N-1}^{t}; \hat{\alpha}_t\right)= \psi_t\left(k, x_t, \hat{u}_{0:k-1}^{t}; \hat{\alpha}_t\right), \\
     & \psi_t\left(k, x_t, \hat{u}_{0:k-1}^{t}; \hat{\alpha}_t\right)= \psi_{t+1}\left(k-1, \hat{x}_{t+1}^{t}, \hat{u}_{1:k-1}^{t}; \hat{\alpha}_t\right),
\end{align}
where the first equality holds since $\hat{x}_{t+k}^{t}$ does not depend on $\hat{u}_{k:N-1}^{t}$, and the second equality holds by the evolution of system dynamics $x_{k+1} = f\left(x_{k}\right) + g\left(x_{k}\right) u_{k} + \Phi\left(z_{k}\right)\hat{\alpha}_t$.
 
Then, we have
\begingroup
\allowdisplaybreaks
    \begin{align}
        & \; V_{t}\left({x}_{t}; \hat{\alpha}_{t}\right) \notag \\
        = & \;  \sum_{k=0}^{N-1} c_{t+k}\left(\psi_t\left(k, x_t, \hat{u}_{0:k-1}^{t}; \hat{\alpha}_t\right),\hat{u}_{k}^{t}\right)  \notag \\
        = & \; c_{t}\left(x_{k},\hat{u}_{0}^{t}\right) + \sum_{k=0}^{N-2} c_{t+k+1}\left(\psi_t\left(k+1, {x}_{t}, \hat{u}_{0:k}^{t}; \hat{\alpha}_t\right),\hat{u}_{k+1}^{t}\right) \notag \\ 
        = & \; c_{t}\left(x_{k},\hat{u}_{0}^{t}\right) + \sum_{k=0}^{N-2} c_{t+k+1}\left(\psi_{t+1}\left(k, \hat{x}_{t+1}^{t}, \hat{u}_{1:k}^{t}; \hat{\alpha}_t\right),\hat{u}_{k+1}^{t}\right) \notag \\ 
        = & \; c_{t}\left(x_{k},\hat{u}_{0}^{t}\right) + \sum_{k=0}^{j-2} c_{t+k+1}\left(\psi_{t+1}\left(k, \hat{x}_{t+1}^{t}, \hat{u}_{1:k}^{t}; \hat{\alpha}_t\right),\hat{u}_{k+1}^{t}\right) \notag \\ 
        &   + \sum_{k=j-1}^{N-2} c_{t+k+1}\left(\psi_{t+1}\left(k, \hat{x}_{t+1}^{t}, \hat{u}_{1:k}^{t}; \hat{\alpha}_t\right),\hat{u}_{k+1}^{t}\right) \notag \\
        = & \; c_{t}\left(x_{k},\hat{u}_{0}^{t}\right) + \sum_{k=0}^{j-2} c_{t+k+1}\left(\psi_{t+1}\left(k, \hat{x}_{t+1}^{t}, \hat{u}_{1:k}^{t}; \hat{\alpha}_t\right),\hat{u}_{k+1}^{t}\right) \notag \\ 
        &   + \sum_{k=0}^{N-j-1} c_{t+k+j}\left(\psi_{t+1}\left(k+j-1, \hat{x}_{t+1}^{t}, \hat{u}_{1:k+j-1}^{t}; \hat{\alpha}_t\right),\hat{u}_{k+j}^{t}\right) \notag \\
        = & \; c_{t}\left(x_{k},\hat{u}_{0}^{t}\right) + \sum_{k=0}^{j-2} c_{t+k+1}\left(\psi_{t+1}\left(k, \hat{x}_{t+1}^{t}, \hat{u}_{1:k}^{t}; \hat{\alpha}_t\right),\hat{u}_{k+1}^{t}\right) \notag \\ 
        &   + \sum_{k=0}^{N-j-1} c_{t+k+1}\left(\psi_{t+j}\left(k, \hat{x}_{t+j}^{t}, \hat{u}_{j:k+j-1}^{t}; \hat{\alpha}_t\right),\hat{u}_{k+j}^{t}\right) . \label{eq:app_V_x_hat_alpha}
    \end{align}
\endgroup

Similarly,
\begingroup
\allowdisplaybreaks
    \begin{align}
        & \; V_{t+1}\left(\hat{x}_{t+1}; \hat{\alpha}_{t}\right) =  V_{t+1}\left(\hat{x}_{t+1}^{t}; \hat{\alpha}_{t}\right) \notag \\
        = & \;  \sum_{k=0}^{N-1} c_{t+k+1}\left(\psi_{t+1}\left(k, \hat{x}_{t+1}^{t}, \hat{u}_{0:k-1}^{t+1}; \hat{\alpha}_t\right),\hat{u}_{k}^{t+1}\right) \notag  \\
        = & \;  \sum_{k=0}^{j-2} c_{t+k+1}\left(\psi_{t+1}\left(k, \hat{x}_{t+1}^{t}, \hat{u}_{0:k-1}^{t+1}; \hat{\alpha}_t\right),\hat{u}_{k}^{t+1}\right)  \notag \\ 
        &   + \sum_{k=j-1}^{N-2} c_{t+k+1}\left(\psi_{t+1}\left(k, \hat{x}_{t+1}^{t}, \hat{u}_{0:k-1}^{t+1}; \hat{\alpha}_t\right),\hat{u}_{k}^{t+1}\right) \notag \\
        = & \;  \sum_{k=0}^{j-2} c_{t+k+1}\left(\psi_{t+1}\left(k, \hat{x}_{t+1}^{t}, \hat{u}_{0:k-1}^{t+1}; \hat{\alpha}_t\right),\hat{u}_{k}^{t+1}\right) \notag \\ 
        & \;  + \sum_{k=0}^{N-j} c_{t+k+j}\left(\psi_{t+1}\left(k+j-1, \hat{x}_{t+1}^{t}, \hat{u}_{0:k+j-2}^{t+1}; \hat{\alpha}_t\right),\hat{u}_{k+j-1}^{t+1}\right) \notag \\
        \leq & \;  \sum_{k=0}^{j-2} c_{t+k+1}\left(\psi_{t+1}\left(k, \hat{x}_{t+1}^{t}, \hat{u}_{1:k}^{t}; \hat{\alpha}_t\right),\hat{u}_{k+1}^{t}\right)  \notag \\ 
        &   + \min_{{v}_{0:N-j}} \sum_{k=0}^{N-j} c_{t+k+j}\left(\psi_{t+j}\left(k, \hat{x}_{t+j}^{t}, {v}_{0:k-1}; \hat{\alpha}_t\right),v_{k}\right) \notag \\
        \leq & \;  \sum_{k=0}^{j-2} c_{t+k+1}\left(\psi_{t+1}\left(k, \hat{x}_{t+1}^{t}, \hat{u}_{1:k}^{t}; \hat{\alpha}_t\right),\hat{u}_{k+1}^{t}\right) \notag \\ 
        & \; + V_{t+j}\left(\hat{x}_{t+j}^{t}; \hat{\alpha}_{t}\right)  \notag \\ 
        \leq & \;  \sum_{k=0}^{j-2} c_{t+k+1}\left(\psi_{t+1}\left(k, \hat{x}_{t+1}^{t}, \hat{u}_{1:k}^{t}; \hat{\alpha}_t\right),\hat{u}_{k+1}^{t}\right) +  \overline{\lambda}\sigma\left(\hat{x}_{t+j}^{t}\right), \label{eq:app_V_hat_x_hat_alpha}
    \end{align}
\endgroup
where the first inequality is due to sub-optimal control sequence $\hat{u}_{1:k}^{t}$ and the definition of $\hat{u}_{j-1:N-1}^{t+1}$, the second inequality holds by definition of $V_{t+j}\left(\hat{x}_{t+j}^{t}; \hat{\alpha}_{t}\right)$, and the last inequality is due to \Cref{assumption:stability}~(ii).

Combining \cref{eq:app_V_x_hat_alpha} and \cref{eq:app_V_hat_x_hat_alpha} gives 
\begin{equation}
    \begin{aligned}
        & \;  V_{t+1}\left(\hat{x}_{t+1}; \hat{\alpha}_{t}\right) - V_{t}\left({x}_{t}; \hat{\alpha}_{t}\right) \\
        \leq & \;  \overline{\lambda}\sigma\left(\hat{x}_{t+j}^{t}\right) - c_{t}\left(x_{t},\hat{u}_{0}^{t}\right) \\
        \leq & \;  \overline{\lambda}\sigma\left(\hat{x}_{t+j}^{t}\right) -  \underline{\lambda}\sigma\left(x_{t}\right),
    \end{aligned}
    \label{eq:app_V_diff}
\end{equation}
where the last inequality uses \Cref{assumption:stability}~(i).

Using \Cref{assumption:stability} and $\hat{x}_{t+j}^{t} = \psi_{t}\left(j, \hat{x}_{t}^{t}, \hat{u}_{0:j-1}^{t}; \hat{\alpha}_t\right)$, for any $j\in[1, \; N-1]$, we have
\begin{equation}
    \begin{aligned}
        & \;  \underline{\lambda} \sum_{j=0}^{N-1} \sigma\left(\hat{x}_{t+j}^{t}\right) \\
        \leq & \;  \sum_{j=0}^{N-1} c_{t+j}\left(\psi_{t}\left(j, \hat{x}_{t}^{t}, \hat{u}_{0:j-1}^{t}; \hat{\alpha}_t\right),\hat{u}_{j}^{t}\right) \\
        = & \;  V_{t}\left({x}_{t}; \hat{\alpha}_{t}\right) \\
        \leq & \; \overline{\lambda}\sigma\left({x}_{t}\right).
    \end{aligned}
\end{equation}
Hence, there exists $j\in[1, \; N-1]$, such that $\sigma\left(\hat{x}_{t+j}^{t}\right) \leq \frac{\overline{\lambda}/\underline{\lambda}}{N-1}\sigma\left({x}_{t}\right)$. 
Then, \cref{eq:app_V_diff} becomes 
\begin{equation}
    \begin{aligned}
        & \;  V_{t+1}\left(\hat{x}_{t+1}; \hat{\alpha}_{t}\right) - V_{t}\left({x}_{t}; \hat{\alpha}_{t}\right) \\
        \leq & \;  \left( \frac{\left(\overline{\lambda}/\underline{\lambda}\right)^2}{N-1} -1 \right) \underline{\lambda} \sigma\left({x}_{t}\right) \\
        \leq & \;  \left( \frac{\left(\overline{\lambda}/\underline{\lambda}\right)^2}{N-1} -1 \right) \left({\underline{\lambda}}/{\overline{\lambda}} \right)V_{t}\left({x}_{t}; \hat{\alpha}_{t}\right) \\
    \end{aligned}
\end{equation}

Then, for all $N > \left(\overline{\lambda}/\underline{\lambda}\right)^2 +1$, we have
\begin{equation}
    V_{t+1}\left(\hat{x}_{t+1}; \hat{\alpha}_{t}\right) \leq \epsilon V_{t}\left(x_{t}; \hat{\alpha}_{t}\right),
\end{equation}
where $\epsilon \triangleq 1 - \left(1 - \frac{\left(\overline{\lambda}/\underline{\lambda}\right)^2}{N-1} \right) \left({\underline{\lambda}}/{\overline{\lambda}} \right) < 1$.
\qed



\begin{figure*}[t]
    \centering
    \subfigure[Average RMSE with maximal speed $0.8m/s$.]{\includegraphics[width=0.36\textwidth]{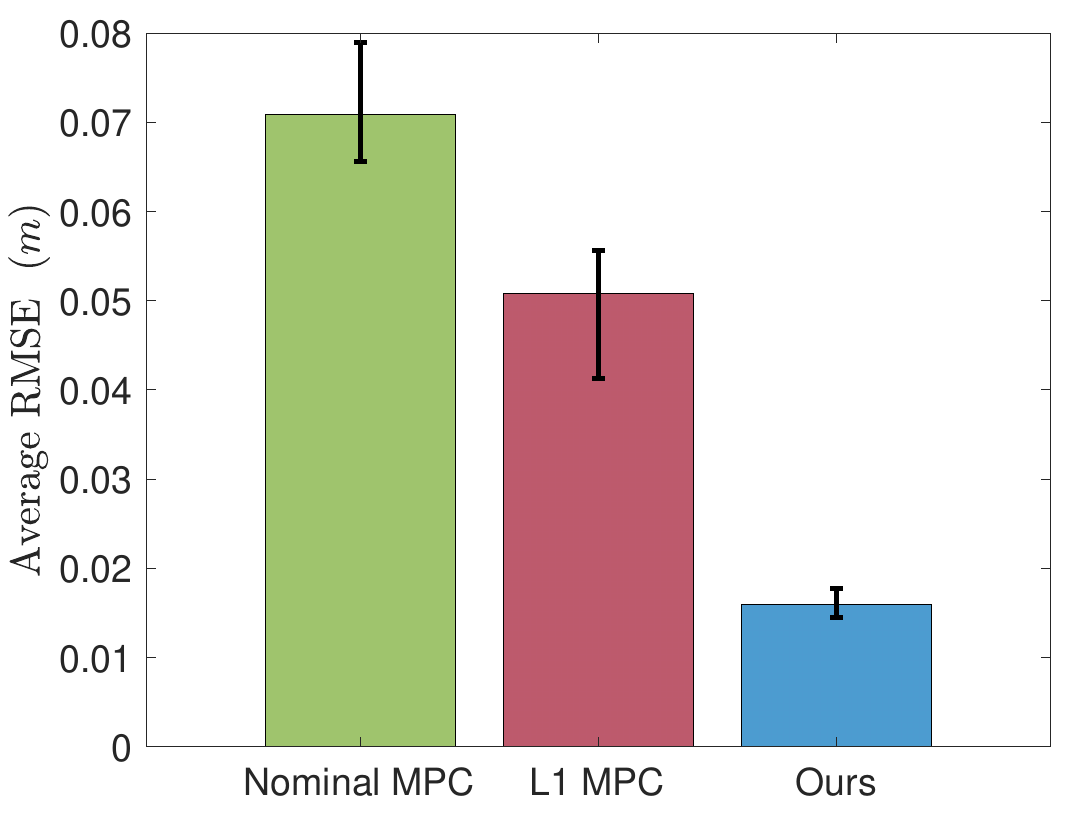}\label{fig_hardware_track_05_error}}	
    \subfigure[Average RMSE with maximal speed $1.3m/s$.]{\includegraphics[width=0.36\textwidth]{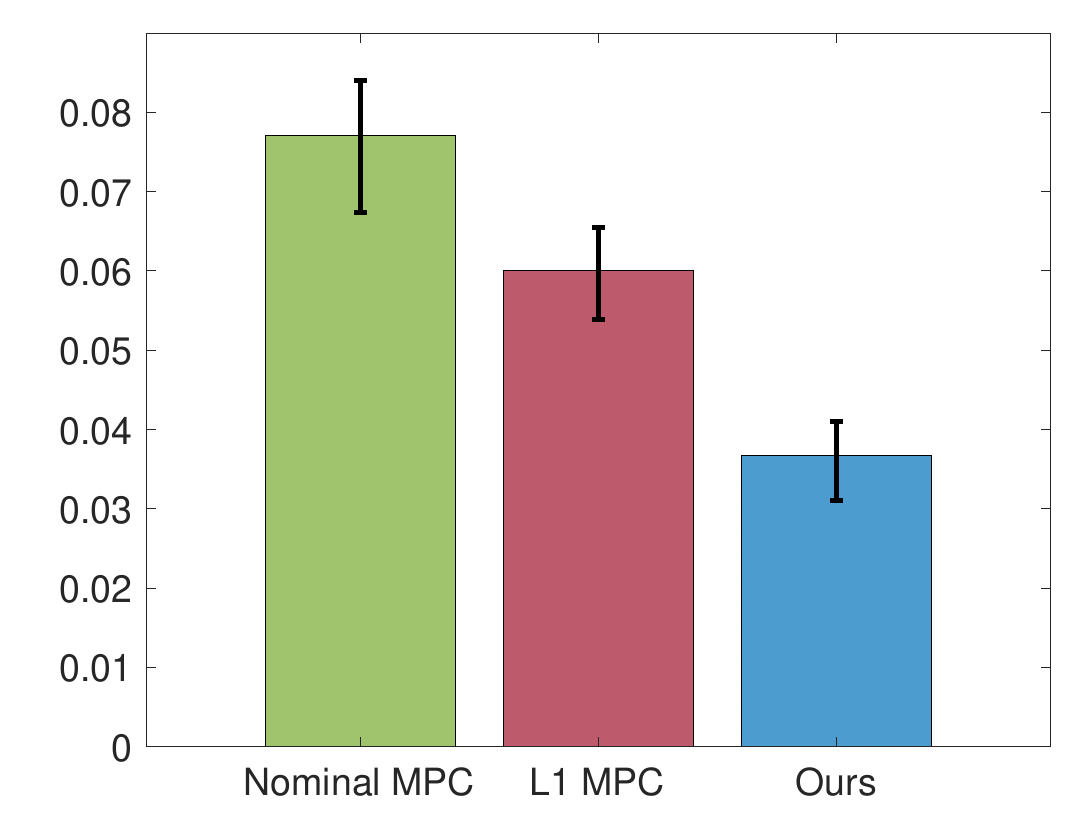}\label{fig_hardware_track_10_error}}	
    \caption{\textbf{Tracking Performance Comparison for the Quadrotor Experiments in \Cref{sec:exp-hardware}.} the quadrotor tracking the circular trajectory at maximal speeds $0.8m/s$ and $1.3m/s$ w/o adding ground effects and wind disturbances. The error bar represents the minimum and maximum RMSE. \Cref{alg:MPC} demonstrates improved performance compared to \NMPC and \LMPC in terms of tracking error over all tested scenarios.}
    \label{fig_hardware_track_rmse}
\end{figure*}

\subsection{Proof of \Cref{lemma:B}}\label{app:lemma_B}

We first establish the Lipschitzness of the value function and boundedness of state by the following lemmas.

\begin{lemma}[Lipschitzness of Value Function]\label{lemma:lipschitz}
    Under \Cref{assumption:lipschitz}, the value function is Lipschitz in the initial condition $x$ and the parameter $\hat{\alpha}$ over bounded domain sets, \ie there exists a constant $L$ such that $|V_t\left(x;\hat{\alpha}\right)-V_t\left(x^\prime;\hat{\alpha}^\prime\right)| \,\leq L \left( \|x - x^\prime \| + \| \hat{\alpha} - \hat{\alpha}^\prime \| \right)$, $\forall x, \; x^\prime, \; \hat{\alpha}, \; \hat{\alpha}^\prime$ such that $\|x - x^\prime \|\, \leq R_1$ and $\| \hat{\alpha} - \hat{\alpha}^\prime \|\, \leq R_2$ for some constants $R_1\geq 0$ and $R_2\geq 0$.
\end{lemma}

\begin{lemma}[Boundedness of State]\label{lemma:boundedness}
    Under \Cref{assumption:stability} and \Cref{assump:small_approx_error}, the state $x_t$ is bounded for all $t$.
\end{lemma}

Then we have 
\begin{equation}
    \begin{aligned}
        &\;  V_{t+1}\left({x}_{t+1}; \hat{\alpha}_{t+1}\right) - V_{t+1}\left(\hat{x}_{t+1}; \hat{\alpha}_{t}\right)  \\
        \leq &\; L \left( \| \hat{x}_{t+1} - {x}_{t+1} \| + \| \hat{\alpha}_{t+1} - \hat{\alpha}_{t} \| \right) \\
        = &\; L \left( \| \Phi\left(z_{t}\right)\hat{\alpha}_{t} - \Phi\left(z_{t}\right)\alpha \| + \| \hat{\alpha}_{t+1} - \hat{\alpha}_{t} \| \right) \\
        = &\; L \left( \sqrt{ l_t\left(\hat{\alpha}_t\right) } + \| \hat{\alpha}_{t+1} - \hat{\alpha}_{t} \| \right) \\
        \leq &\; L \left( \sqrt{ l_t\left(\hat{\alpha}_t\right) } + \| \eta \nabla_t \| \right),
    \end{aligned}
\end{equation}
where the first inequality holds per \Cref{assumption:lipschitz}, the first equality holds per \Cref{assump:small_approx_error}, the second inequality holds by definitions of $l_t\left(\hat{\alpha}_t\right)$, and the second inequality holds by the Pythagorean theorem \cite[Theorem~2.1]{hazan2016introduction}.
\qed

\subsection{Proof of \Cref{lemma:lipschitz}}\label{app:lemma_lipschitz}

Given that ${c}_t\left(x_t, u_t\right)$ and $x_{k+1} = f\left(x_{k}\right) + g\left(x_{k}\right) u_{k} + \hat{h}\left(z_{k}\right)$ are locally Lipschitz in $x_t$, $u_t$, and $\hat{\alpha}$, we have that the objective function in \cref{eq:ada_mpc_value} is locally Lipschitz in the initial condition $x$, $u_k$ and $\hat{\alpha}$ by substituting \cref{eq:ada_mpc_value_dyn} into \cref{eq:ada_mpc_value_obj}.
Also, since $\calU$ is compact, the objective function is Lipschitz in $\calU \times \dots \times \calU$ and the value function is Lipschitz in $x$ and $\hat{\alpha}$.
Then the proof follows the same steps in the proof of \cite[Theorem~C.~29]{rawlings2017model} 
\qed





\subsection{Proof of \Cref{lemma:boundedness}}\label{app:lemma_boundedness}

Under \Cref{assump:small_approx_error}, since $\Phi\left(\cdot\right)$, $\alpha$, $\hat{\alpha}_t$ are uniformly bounded for all $t$, then $l_t\left(\hat{\alpha}_t\right)$ and $\nabla_t$ are bounded for all $t$. From \cref{eq:app_mpc_stability_value}, we have:
\begin{equation}
    \begin{aligned}
        & \; V_{t+1}\left({x}_{t+1}; \hat{\alpha}_{t+1}\right)  \\
        \leq & \; \epsilon^t \overline{\lambda}\sigma\left(x_1\right) + \sum_{k=1}^{t} \epsilon^{t-k} L \left( \sqrt{ l_k\left(\hat{\alpha}_k\right) } + \| \eta \nabla_k \| \right) \\
        \leq & \; \epsilon^t \overline{\lambda}\sigma\left(x_1\right) +  \frac{\mu}{1-\epsilon} \\
        \leq & \; \overline{\lambda}\sigma\left(x_1\right) +  \frac{\mu}{1-\epsilon},
    \end{aligned}
\end{equation}
where the second inequality holds with $\mu$ as the uniform upper bound of $L \left( \sqrt{ l_t\left(\hat{\alpha}_t\right) } + \| \eta \nabla_t \| \right)$ for all $t$, and the third inequality holds by  $\epsilon \in (0,1)$.

Under \Cref{assumption:stability}, we have
\begin{equation}
    \begin{aligned}
        \sigma\left(x_t\right) \leq & \frac{1}{\underline{\lambda}} c_{t}\left(x_t,u_t\right) \leq  \frac{1}{\underline{\lambda}} V_{t}\left(x_{t}; \hat{\alpha}_{t}\right) \\
        \leq & \; \frac{\overline{\lambda}}{\underline{\lambda}}\sigma\left(x_1\right) +  \frac{\mu}{\underline{\lambda}(1-\epsilon)},
    \end{aligned}
\end{equation}
where we obtain that $\sigma\left(x_t\right)$ is bounded by constant $\frac{\overline{\lambda}}{\underline{\lambda}}\sigma\left(x_1\right) +  \frac{\mu}{\underline{\lambda}(1-\epsilon)}$.
Hence, $x_t$ is bounded for all $t$.
\qed

\subsection{Discussion on State Constraints}\label{app:safety}
In the \MPC problem in \cref{eq:mpc_ada_def}, we can combine several approaches with the current framework to handle state constraints. For example:
\begin{itemize}
    \item Safety filter with robust control barrier function~\cite{jankovic2018robust}. Since this can arbitrarily modify the control and the resulting state, the no-regret guarantee~(\Cref{theorem:regret_OLMPC}) can be difficult to derive.
    \item Robust safety in robust \MPC with an assumption of bounded $\| h\left(\cdot\right)\|$~\cite{mayne2005robust,bouffard2012learning,aswani2012extensions}. {If the Lipschitzness of the value function can be obtained~\cite[Theorem~4.14]{dempe2002foundations}}, we expect that \Cref{theorem:regret_OLMPC}~(no-regret guarantee) would still hold when the $u_t$ in dynamic regret definition is obtained by solving \MPC~(\cref{eq:mpc_ada_def}) with tightened state constraints. 
    \item Uncertainty quantification using the set-membership algorithm~\cite{milanese1991optimal} for the case of time-invariant optimal parameter $\alpha^\star$. Under the assumption of bounded $\|h\left(\cdot\right)\|$, this method essentially prunes the uncertainty set $\calD\left(B_h\right)$ at every time step using the observed $h\left(z_t\right)$, and the pruned uncertainty set $\calD_t$ is non-expanding. Then in line~9 of \Cref{alg:MPC}, we need to project $\hat{\alpha}_{i,t+1}^\prime$ onto $\calD_t$, and the \Cref{theorem:OGD} remains the same.
    We can apply an approach similar to the robust safety method for deriving tightened state constraints and regret guarantees.
\end{itemize}

\subsection{Additional Results of Hardware Experiments}\label{app:hardware}
In this section, we provide additional hardware experiments results to \Cref{sec:exp-hardware} of the quadrotor tracking the circular trajectory at speeds $0.8m/s$ and $1.3m/s$ w/o ground effects and wind disturbances. In this case, the quadrotor suffers from unknown body drag, rotor drag, and turbulent effects caused by the propellers.
The results are given in \Cref{fig_hardware_track_rmse}. \Cref{alg:MPC} demonstrates improved performance compared to \NMPC and \LMPC in terms of tracking error in both cases of maximal speeds.


\subsection{Stability Analysis}\label{app:stability}
In this section, we analyze the stability of \Cref{alg:MPC} under the assumption of the stability of the nominal system. Such nominal stability is common in the literature~\cite{lale2021model,boffi2021regret,boffi2022nonparametric,karapetyan2024regret,nonhoff2024online,agarwal2019online,zhao2022non,gradu2023adaptive,li2021online,zhou2023efficient,zhou2023safecdc,zhou2023saferal,muthirayan2022online,mayne2011tube,kohler2020computationally,berberich2020data}. 

\begin{theorem}[Stability Analysis]
Assume that \Cref{assumption:stability} only holds for the nominal system~(\cref{eq:mpc_def_dyn}, \ie $\alpha = 0$), $\|g(\cdot)\|\leq\bar{g}$, domain set $\calD$ is compact with radius of $D$, control input is Lipschitz in terms of $\hat{\alpha}$~($\|\bar{u}_{k} - u_{k}\| \leq L_u \|\hat{\alpha}_{t} - 0 \| \leq L_u D$), and  unknown disturbances is bounded with $\|h(\cdot)\|\leq\bar{h}$. Then the state $x_t$ remain bounded under $\Cref{alg:MPC}$.
\end{theorem}

\begin{proof}
    
Since \Cref{assumption:stability} holds for the nominal system, then \Cref{lemma:A} implies that there exists a nominal control input $\bar{u}_{t}$ such that 
\begin{equation}
    V_{t+1}\left(\bar{x}_{t+1}; 0\right) \leq \epsilon V_{t}\left(\bar{x}_{t}; 0\right),
\end{equation}
where $ \bar{x}_{t+1} = f\left(\bar{x}_{t}\right) + g\left(\bar{x}_{t}\right) \bar{u}_{t}$.

By \Cref{lemma:lipschitz}, we have
\begin{equation}
    \begin{aligned}
        & |V_{t+1}\left(\bar{x}_{t+1};0\right)-V_{t+1}\left(x_{t+1};0\right)| \; \\
        \leq  & L  \|\bar{x}_{t+1} - x_{t+1} \| \\
        =& L  \|g(\bar{x}_t) (\bar{u}_{t} - u_{t}) + h(z_t) \|
    \end{aligned}
\end{equation}

Similar to \Cref{lemma:boundedness}, we have 
\begingroup
\allowdisplaybreaks
    \begin{align}
        & V_{t+1}\left(x_{t+1};0\right) \\
        \leq &\epsilon V_{t+1}\left(\bar{x}_{t+1};0\right) +  L   \|g\left(\bar{x}_t\right) (\bar{u}_{t} - u_{t}) + h\left(z_k\right) \| \\
        \leq &\epsilon^t V_{1}\left(\bar{x}_{1};0\right) +  L  \sum_{k=1}^{t} \epsilon^{t-k}\|\left(\bar{x}_t\right) (\bar{u}_{k} - u_{k}) + h\left(z_k\right) \| \\
        \leq &\epsilon^t \sigma\left(\bar{x}_{1}\right) +  L  \sum_{k=1}^{t} \epsilon^{t-k}\|\left(\bar{x}_t\right) (\bar{u}_{k} - u_{k}) + h\left(z_k\right) \| \\
        \leq &\epsilon^t \sigma\left(\bar{x}_{1}\right) +  \frac{L}{1-\epsilon} (\bar{g}L_u D + \bar{h}),
\end{align}
\endgroup
where the last inequality follows by assuming $\|g(\cdot)\|\leq\bar{g}$, compactness of $\calD$ with radius of $D$, Lipschitzness of control input $\|\bar{u}_{k} - u_{k}\| \leq L_u \|\hat{\alpha}_{t} - 0 \| \leq L_u D$, and boundedness of unknown disturbances $\|h(\cdot)\|\leq\bar{h}$.

The boundedness of $x_t$ follows by \Cref{assumption:stability}, \ie $\underline{\lambda} \sigma\left(x_t\right) \leq c_{t}\left(x_t,u_t\right) \leq V_{t}\left(x_{t};0\right)$.
\end{proof}

\bibliographystyle{IEEEtran}
\bibliography{References}

\begin{IEEEbiography}[{\includegraphics[width=1in,height=1.25in,clip,keepaspectratio]{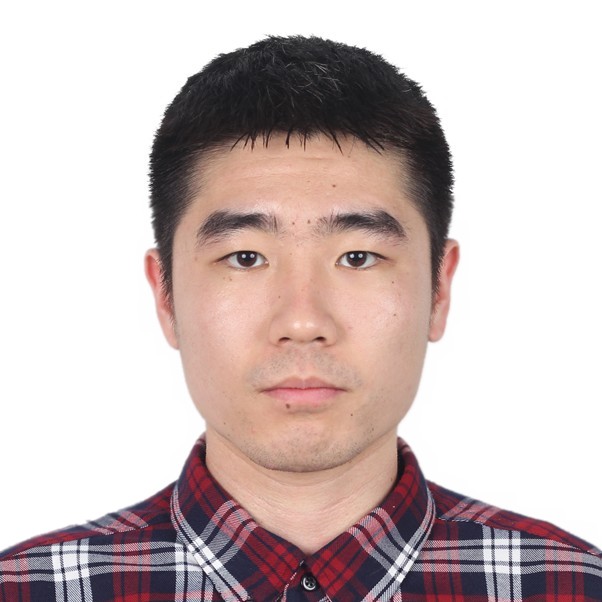}}]{Hongyu Zhou} received the B.Eng. degree in Naval Architecture and Ocean Engineering from the Huazhong University of Science and Technology in 2018, and the M.Sc. degree in Marine Technology from the Norwegian University of Science and Technology in 2020. He received M.Sc. degree in Robotics from the University of Michigan in 2025. He is currently pursuing a Ph.D. degree in Aerospace Engineering at the University of Michigan. Hongyu works on algorithms to achieve agile, efficient, and safe robot motion control in uncertain and dynamic environments, with formal performance guarantees. His research interests include optimal control, adaptive control, model learning for control, and online learning.
\end{IEEEbiography}

\begin{IEEEbiography}[{\includegraphics[width=1in,height=1.25in,clip,keepaspectratio]{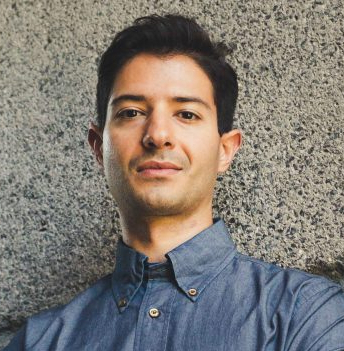}}]{Vasileios Tzoumas} (IEEE Senior Member) received his Ph.D. in Electrical and Systems Engineering at the University of Pennsylvania (2018). He holds a Master of Arts in Statistics from the Wharton School of Business at the University of Pennsylvania (2016), a Master of Science in Electrical Engineering from the University of Pennsylvania (2016), and a diploma in Electrical and Computer Engineering from the National Technical University of Athens (2012). Vasileios is as an Assistant Professor in the Department of Aerospace Engineering, University of Michigan, Ann Arbor. Previously, he was at the Massachusetts Institute of Technology (MIT), in the Department of Aeronautics and Astronautics, and in the Laboratory for Information and Decision Systems (LIDS), where he was a research scientist (2019-2020) and a post-doctoral associate (2018-2019). Vasileios works on algorithms and innovative hardware for scalable and reliable cyber-physical systems in resource-constrained, uncertain, and contested environments via resource-aware decision-making, online learning, and resilient adaptation. Vasileios is a recipient of an NSF CAREER Award, the Best Paper Award in Robot Vision at the 2020 IEEE International Conference on Robotics and Automation (ICRA), an Honorable Mention from the 2020 IEEE Robotics and Automation Letters (RA-L), and was a Best Student Paper Award finalist at the 2017 IEEE Conference in Decision and Control (CDC).
\end{IEEEbiography}


\end{document}